\documentclass[10pt,twocolumn,letterpaper]{article}

\usepackage{cvpr}              

\usepackage{bm}

\def\rvepsilon{{\mathbf{\epsilon}}}

\def\vzero{{\bm{0}}}

\def\vc{{\bm{c}}}

\def\vf{{\bm{f}}}

\def\vn{{\bm{n}}}

\def\vp{{\bm{p}}}

\def\mI{{\bm{I}}}

\DeclareMathAlphabet{\mathsfit}{\encodingdefault}{\sfdefault}{m}{sl}
\SetMathAlphabet{\mathsfit}{bold}{\encodingdefault}{\sfdefault}{bx}{n}
\newcommand{\tens}[1]{\bm{\mathsfit{#1}}}

\def\tG{{\tens{G}}}

\newcommand{\normltwo}{L^2}

\newcommand{\mytilde}{\raisebox{0.5ex}{\texttildelow}}

\usepackage{multirow} 
\usepackage[ruled,vlined]{algorithm2e}
\providetoggle{beginFlag}
\settoggle{beginFlag}{true}
\SetAlgoNlRelativeSize{0} 

\SetAlgoSkip{AlgoSkipBeforeAndAfter}

\newcommand{\ourmethod}{ShapeShifter\xspace}
\newcommand{\silenced}[1]{}

\def\feat{\vf}

\def\mask{m}
\def\grid{\tG}

\def\lvl{l}
\def\maxLvl{L}
\def\model{\mathcal{M}}
\def\upsampler{\mathcal{U}}

\def\blurredGrid{\tilde{\grid}}

\usepackage{xurl}

\definecolor{cvprblue}{rgb}{0.21,0.49,0.74}
\usepackage[pagebackref,breaklinks,colorlinks,allcolors=cvprblue]{hyperref}

\def\paperID{15165} 
\def\confName{CVPR}
\def\confYear{2025}

\title{\ourmethod:  3D Variations Using Multiscale and Sparse Point-Voxel Diffusion}

\author{
Nissim Maruani\\[-0.5mm]
\small{Inria, Universit\'e C\^ote d'Azur}\\[-1mm]
{\tt\small nissim.maruani@inria.fr}
\and
Wang Yifan\\[-.5mm]
\small{Adobe Research}\\[-1mm]
{\tt\small yifwang@adobe.com}
\and
Matthew Fisher\\[-.5mm]
\small{Adobe Research}\\[-1mm]
{\tt\small matfishe@adobe.com}
\and
Pierre Alliez\\[-.5mm]
\small{Inria, Universit\'e C\^ote d'Azur}\\[-1mm]
{\tt\small pierre.alliez@inria.fr}
\and
Mathieu Desbrun\\[-.5mm]
\small{Inria/X, IP Paris}\\[-1mm]
{\tt\small mathieu.desbrun@inria.fr}
}

\begin{document}
\maketitle
\begin{abstract}
This paper proposes \ourmethod, a new 3D generative model that learns to synthesize \emph{shape variations} based on a single reference model. 
While generative methods for 3D objects have recently attracted much attention, current techniques often lack geometric details and/or require long training times and large resources. Our approach remedies these issues by combining sparse voxel grids and point, normal, and color sampling within a multiscale neural architecture that can be trained efficiently and in parallel. We show that our resulting variations better capture the fine details of their original input and can handle more general types of surfaces than previous SDF-based methods. Moreover, we offer interactive generation of 3D shape variants, allowing more human control in the design loop if needed.
\end{abstract}\vspace*{-8mm}
\section{Introduction}
\label{sec:intro}

\silenced{
First paragraph (what's our task):
\begin{enumerate}
\item  The problem we are addressing is 3D Generation. 
\item  This space is dominated by very few players, because of resource-intensive nature of this task. At the same time, the geometric quality is a weakpoint because the quality variance in existing 3D dataset is enormous, and among all available training data, there are few with detailed geometric features. 
\item We take a different approach: instead of training from large dataset, we learn to generate variations from a single input shape. 
\end{enumerate}}

Creating 3D content through generative models is currently attracting significant attention. Traditional 3D modeling demands both time and specialized skills to create complex shapes, whereas advancements in generative AI promise a broader exploration of design possibilities, free from the usual constraints of time or technical expertise. However, current 3D generative models have numerous shortcomings that limit their usefulness in applications such as movies, gaming, and product design. First, state-of-the-art methods often struggle to produce the fine geometric details and sharp features necessary for digital shapes in geometric modeling. Additionally, these models require large, high-quality 3D datasets, which are significantly more challenging to curate compared to image datasets, and involve long training times and substantial computational resources.

\silenced{
Second paragraph (what do existing work):
\begin{enumerate}
    \item Existing method could generate visually OK 3D asset, but the geometry struggles to reach the comparable level of quality. 
    \item The main cause is that their representation is not geometry-centric. They use occupancy field~\cite{wu2022learning, li2023patch, son2023singraf, zhao2024michelangelo} or signed distance functions~\cite{wu2024sindm, zhang20233dshape2vecset} as the geometric representation, smoothing out sharp features. \nissimrmk{This means they use VOLUME supervision whereas we use SURFACE supervision which is much faster} At the same time, supervised with through volumetric rendering ~\cite{son2023singraf, li2023patch}. Often times, texture hides the geometry artifacts.
    \item They are also very slow, requiring hours of training and (or) long inference time.
\end{enumerate}}

\begin{figure}[!ht] 
    \centering
     \includegraphics[width=\linewidth]{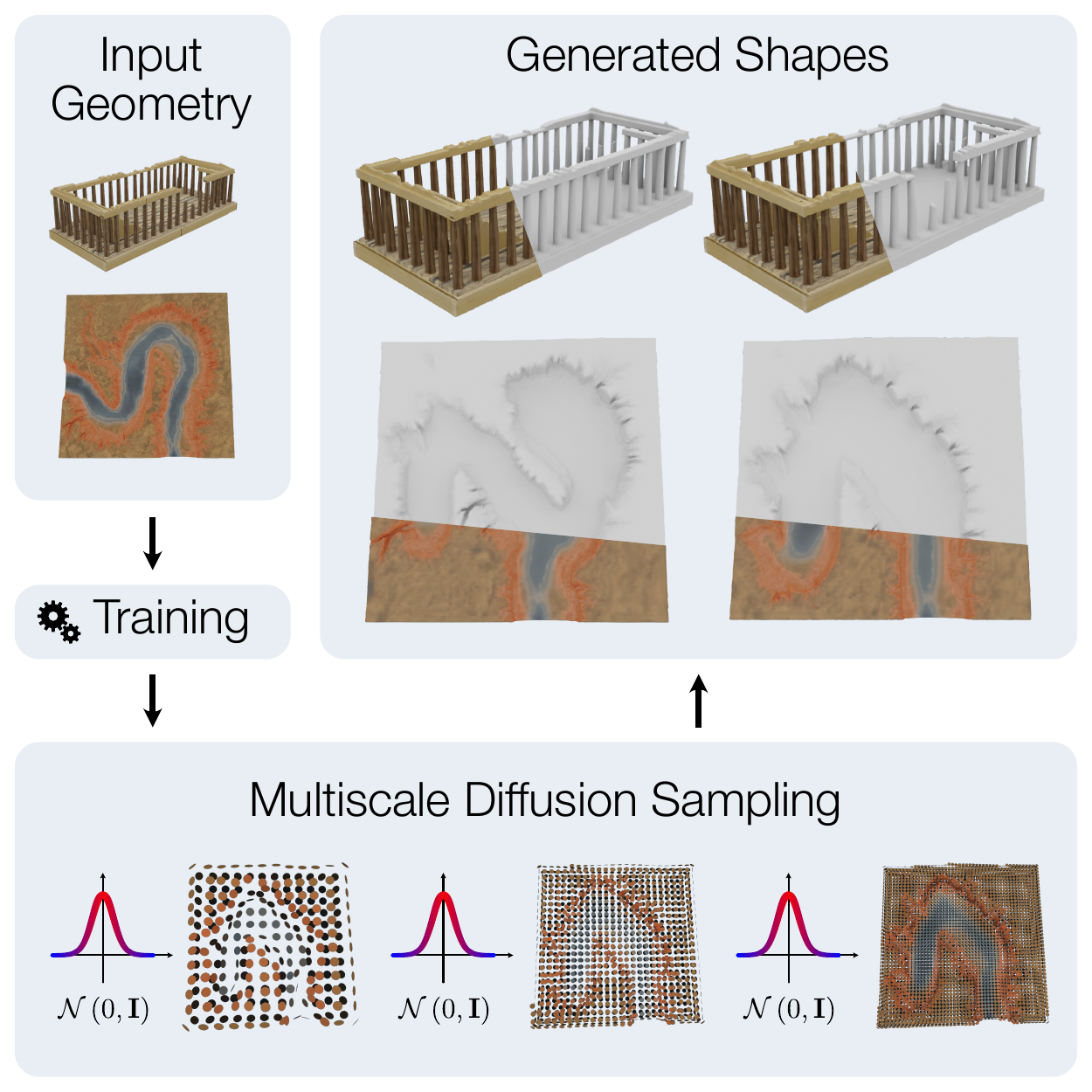}\vspace*{-1mm}
    \caption{\emph{\ourmethod.} Given a 3D exemplar, we propose to train a hierarchical diffusion model to create variations preserving the \emph{geometric details and styles} of the exemplar. By combining compact yet explicit 3D features (colored, oriented points) with a sparse voxel grid, we shorten training times from hours to minutes, while yielding significantly better geometric quality than prior work. The hierarchical point representation and fast inference times further enable intuitive interactive editing.\vspace*{-5mm}}
    \label{fig:enter-label}
\end{figure}

\begin{figure*}[t]
    \centering
      \includegraphics[width=0.99\linewidth]{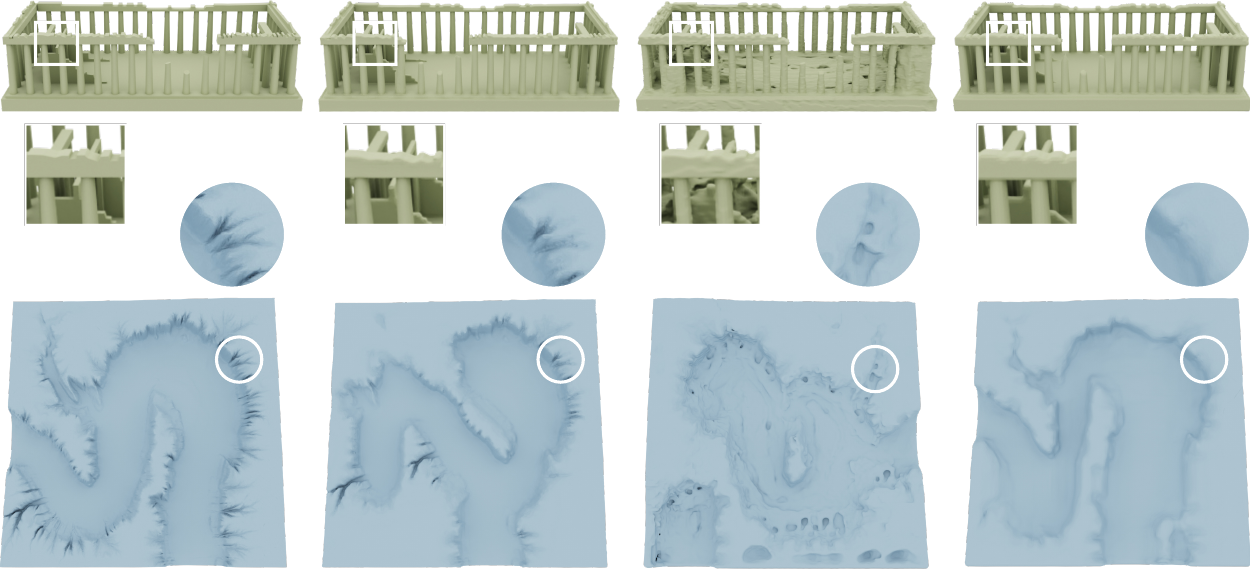}\vspace*{-5mm}
      \subfloat[Input Geometry]{\hspace{.25\linewidth}}
      \subfloat[Ours]{\hspace{.25\linewidth}}
        \subfloat[Sin3DGen]{\hspace{.25\linewidth}}
              \subfloat[Sin3DM]{\hspace{.25\linewidth}}
    \vspace*{-2.5mm}
    \caption{\emph{Geometric details.} Our generation captures significantly more geometric details present in the exemplar mesh (leftmost). Prior work, Sin3DGen~\cite{li2023patch} and Sin3DM~\cite{wu2024sindm}, operates with plenoxels and neural radiance fields encoded in single-resolution triplane features, respectively, which lack the capability to sufficiently represent and supervise high-resolution geometric details. In contrast, our method employs a colored and oriented point set, providing precise geometric information.\vspace*{-3mm}}
    \label{fig:results}
\end{figure*}

In this paper, 
we tackle a task-specific, yet effective approach to synthesizing geometry: we propose generating shape variations from a single high-quality example. This lesser-explored generative method offers several benefits beyond avoiding the curation of large training datasets: it has the potential to provide a resource-efficient way to generate shape variants for retargeting or editing, automatically inheriting the style, symmetries, semantics, and geometric details from the exemplar.
Although existing generative methods from exemplars are able to create varied 3D assets, they struggle to produce \emph{clean and detailed geometry} due to their reliance on occupancy fields~\cite{wu2022learning, li2023patch, son2023singraf} or signed distance functions~\cite{wu2024sindm} (which smooth out geometric features), or because they are supervised through volumetric rendering~\cite{li2023patch,son2023singraf} (which often leads to large geometric artifacts) --- and without a clean geometric output model, the use of 2D textures to further enhance visual complexity is severely hindered. Consequently, existing exemplar-based methods are relative slow in generating variations of the input as they rely on volumetric sampling within the surface's neighborhood~\cite{wu2022learning,son2023singraf,wu2024sindm}. 

We propose a novel technique, that we call \ourmethod, to synthesize high-quality shape variations of an input 3D model, with training and inference times well suited for practical real-world applications. We use points (with their normals and optionally colors for additional semantic information) as our lightweight and efficient base geometric representation~\cite{prokudin_dynamic_2023}, which we pair with a multiscale 3D diffusion network. While these explicit surface features already streamline the generative process and help preserve geometric details, we propose to significantly reduce training times and achieve interactive inference rates by adopting sparse convolutions based on fVDB~\cite{williams2024fvdb}, a recent spatial learning framework based on sparse voxel grids.
Mixing point sampling and sparse convolutions, a novel combination in generative modeling, results in a multiscale generative approach capable of producing 3D variants of shapes of different styles and topologies. Furthermore, its fast inference allows for interactive editing control.
\silenced{
Fourth paragraph (contribution):
\begin{enumerate}
    \item Much higher geometry quality 
    \item Much shorter training time 
    \item Multi-level editability
    \item Open surfaces
    \item Can be combined with off-the-shelf 2D image suepr-resolution to generate HD texture, exceeding prior work in visual details. Flexible to be coupled with various texturing sythesis methods such as ~\cite{richardson2023texture} for more stylized texture creating.
\end{enumerate}}

\vspace*{-5mm}
\paragraph{Contributions.} This paper proposes a neural network approach to generating high-quality shapes from a single 3D reference example. Compared to previous exemplar-based generative methods, we demonstrate significantly improved geometric quality of our outputs, as shown in \cref{fig:results}. Moreover, the simplicity of our geometric representation (using point sampling in a sparse voxel grid) and its hierarchical refinement (learned per level in parallel) to control and generate variations of an arbitrary closed or open input shape results in significantly reduced training times (minutes instead of hours) and interactive inference. While our results can be easily converted into textured meshes, 
direct visualization of our point-based representation in realtime enables iterative co-creation guided by an artist. Finally, our high-quality output geometric models can be assigned a fine texture if needed, using off-the-shelf image super-resolution or more advanced texturing synthesis methods such as~\cite{richardson2023texture}. \vspace*{-1.5mm}

\section{Related Work}
\label{sec:related}

\paragraph{3D Generation.} 
The field of 3D generation has seen rapid development in recent years. Advances in generative models and large-scale 3D datasets have underpinned this progress. Generative adversarial networks (GANs)~\cite{goodfellow2020generative} have been widely used in works like~\cite{achlioptas2018learning, gao2022get3d, chan2022efficient}, while normalizing flows~\cite{rezende2015variational} were utilized in~\cite{yang2019pointflow}. Other approaches include variational autoencoders (VAEs)~\cite{kingma2019introduction} and autoregressive (AR) models~\cite{bengio1994learning, graves2013generating, van2016pixel}, as shown in~\cite{park2019deepsdf, zhang20223dilg, siddiqui2024meshgpt, yin2023shapegpt, nash2020polygen,chen2024meshanything}.
The recent introduction of diffusion models~\cite{sohl2015deep, ho2020denoising} has enabled training on larger datasets such as Objaverse~\cite{deitke2024objaverse}. A survey by Po \etal~\cite{po2024state} provides a comprehensive taxonomy of 3D diffusion approaches. A primary line of work builds on 2D diffusion models, generating multiview-consistent images through Score Distillation Sampling (SDS)~\cite{poole2023dreamfusion, wang2023score}. However, SDS faces practical challenges such as high optimization times~\cite{metzer2023latent, liang2024luciddreamer}, color artifacts~\cite{lukoianov2024score, chen2023fantasia3d, wang2024prolificdreamer, li2024sweetdreamer}, and 3D inconsistencies~\cite{liu2024syncdreamer, wang2024taming}.
Fine-tuning diffusion models on 3D assets for direct multiview output~\cite{shi2024mvdream, liu2023zero, qiu2024richdreamer, long2024wonder3d, lin2023magic3d} can address these issues, with further speedups through reconstructor networks for radiance fields~\cite{li2024instantd, chen2023single, liu2024one, wei2024meshlrm, wang2024pflrm} or Gaussian splats~\cite{zhang2025gs, xu2024grm, zou2024triplane}. However, photometric losses often lead to geometric artifacts.
A separate direction directly trains 3D diffusion models on 3D data~\cite{luo2021diffusion, nichol2022point} or encodes 3D data through autoencoders~\cite{cheng2023sdfusion, gupta20233dgen, zhang20233dshape2vecset,vahdat2022lion, jun2023shap, zhao2024michelangelo, ren2024xcube, ren2024scube, zhang2024clay}. These methods demand extensive, high-quality data and substantial computational resources, and the generated geometry, while improved, still lacks the geometric details required in real-world 3D applications. Although Xcube~\cite{ren2024xcube} initially proposed using an efficient multiscale diffusion pipeline, it relies on pre-trained VAEs and verbose SDF representation which are not adapted to the problem of single-shape variations.
We adopt an alternative approach which generates 3D assets with high-quality geometry from a single exemplar, trainable on a single GPU in minutes, while enabling user control over output shapes.

\begin{figure}[!t]
    \includegraphics[width=\linewidth]{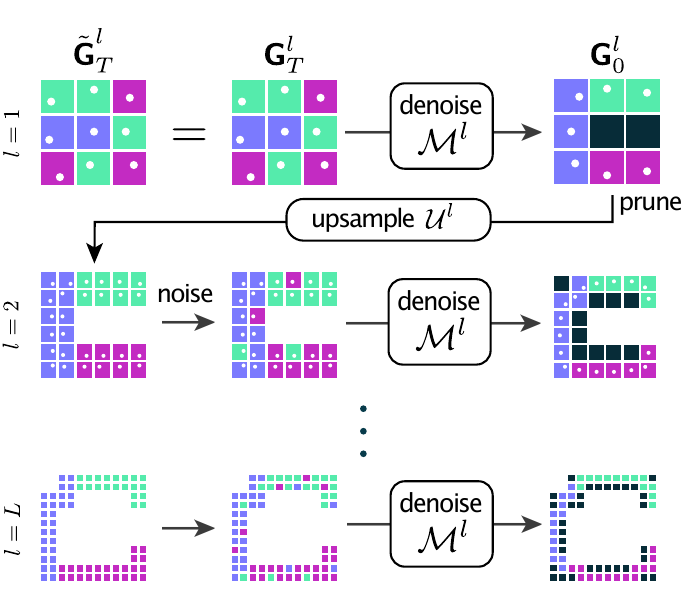}\vspace*{-4mm}
    \caption{\emph{Multiscale diffusion on sparse voxel grid.} We start from noise \(\rvepsilon\!\sim\!\mathcal{C}\left( \vzero,\mI \right)\) at the coarsest level \(\lvl\!=\!1\), and obtain the 3D feature grid \(\grid^{\lvl}\) through reverse diffusion. Each subsequent level uses the output of the previous level. Inactive voxels are first pruned, then upsampled with a level-specific upsampler \(\upsampler^{\lvl}\). The upsampled grid \(\blurredGrid{\vphantom{\grid}}^{\lvl}\) is subsequently noised and passed through the diffusion model to obtain a clean version of the sparse feature grid \(\grid^{\lvl}\). All levels are independent and can thus be trained in parallel. \vspace*{-5mm}} \label{fig:overview}
\end{figure}

\vspace*{-5mm}
\paragraph{Generation from a single instance.}
Despite advancements in large-scale 3D generation, high-quality 3D data remains scarce, and computational costs for training and inference are significant. Generating 3D content from a single high-quality example offers an appealing alternative, giving users control through concrete exemplar inputs. Non-learning, patch-based methods such as PatchMatch~\cite{han2008multiscale,barnes2009patchmatch} and more recent works~\cite{granot2022drop, elnekave2022generating} produce variations by finding and blending similar patches within the exemplar.
In contrast, SinGAN~\cite{shaham2019singan} and its successors~\cite{shocher2019ingan, hinz2021improved} have shown that generative models can be trained on a single example. More recently, SinFusion~\cite{nikankin2023sinfusion, kulikov2023sinddm} demonstrated that diffusion models, known for their stability, can also be adapted to this approach. Applications of these single-instance models include texture synthesis~\cite{zhou2018non, niklasson2021self, rodriguez2022seamlessgan, Mitchel_2024_CVPR}, video synthesis~\cite{haim2022diverse}, and more.

In 3D generation, similar approaches have emerged. Herz \etal~\cite{10.1145/3386569.3392471} applied SinGAN with mesh convolutions~\cite{hanocka2019meshcnn} to enhance surface details without altering structure, while works like Son \etal~\cite{son2023singraf} and Karnewar \etal~\cite{karnewar20223ingan} generated radiance fields from a single instance. Wu \etal~\cite{wu2022learning} used 3D convolutions to generate significant variations, later extending their work to include texture synthesis~\cite{wu2024sindm}, though details are often degraded due to 3D convolution bottlenecks. These methods, however, are typically slow, with training times ranging from 2 to 4 hours.
To bypass training, Li \etal~\cite{li2023patch} adopted a PatchMatch-based method for 3D scenes represented by plenoxels~\cite{fridovich2022plenoxels}. Although this removes the training step, it still requires approximately 10 minutes per variation and struggles to maintain geometric quality due to challenges in converting the plenoxels’ occupancy fields to signed distance functions.
By combining the strengths of a recent framework for learning sparse spatial information~\cite{williams2024fvdb} with compact, geometry-centric features, 
our method achieves higher geometric fidelity while requiring less than 12 minutes for the entire generation process (including training and inference).

\vspace*{-4mm}
\paragraph{3D Representation.}

Earlier 3D representations focused solely on geometry, using formats like voxels~\cite{wu20153d}, 3D points~\cite{qi2017pointnet}, meshes~\cite{hanocka2019meshcnn, maruani_voromesh_2023, maruani_ponq_2024}, SDFs~\cite{park2019deepsdf}, and occupancy fields~\cite{mescheder2019occupancy,Chen_2019_CVPR}.
Differentiable rendering later enabled joint modeling of geometry and appearance, spanning from neural mesh rasterization~\cite{kato2018neural,liu2019soft} and splatting methods~\cite{yifan2019differentiable,wiles2020synsin,kerbl20233d} to neural radiance fields~\cite{lombardi2019neural,mildenhall2021nerf}.
This integration allowed the use of 2D assets~\cite{chan2022efficient,schwarz2020graf} and facilitated large-scale generation and reconstruction~\cite{xu2024dmvd,xu2024grm,wei2024meshlrm,tang2025lgm,hong2024lrm}, enhancing semantic context for geometry generation. However, appearance signals often exhibit higher frequencies than geometry, leading existing methods to separate texture and geometry branches~\cite{mildenhall2021nerf}. Tying texture generation to geometry imposes unnecessarily high resolution requirement on geometry, slowing single-exemplar generation~\cite{wu2024sindm}.
Other works generate geometry first, before refining the appearance in 2D texture space at higher resolutions~\cite{huo2025texgen,yang2024dreammesh,zeng2024paint3d}. Recently, Clay~\cite{zhang2024clay} fully separates geometry and appearance aspects, achieving state-of-the-art quality in both. Following this idea, we focus on high-fidelity geometry generation, which is the weakest aspect of current 3D generation approaches. Unlike Clay’s strict separation, however, we use RGB features to add contextual information-—crucial for single-exemplar generation where data priors are weaker.

Our results show that this approach produces superior geometry which, when combined with state-of-the-art image super-resolution~\cite{kang2023scaling,li2022srdiff} and texture synthesis~\cite{richardson2023texture,Cao_2023_ICCV}, allows for highly detailed shapes with sharp geometry and HD textures. We argue that this design, for a fixed computational budget, optimally balances quality and efficiency.


\section{Method}
\label{sec:method}

Our goal is to train a generative model from a single 3D input mesh to generate new variations efficiently.
We use a multiscale diffusion model with limited receptive fields to learn the internal structures of the given shape, adapting an approach that has been used for training a generative model on a single image~\cite{kulikov2023sinddm}.
We use compact, explicit 3D features directly extracted from the exemplar shape for diffusion.
These features are encoded in a sparse voxel grid, and processed efficiently using a specialized 3D convolution framework (fVDB~\cite{williams2024fvdb}) to capture fine-scale geometric details without incurring high memory cost.
We introduce our 3D features in \cref{sec:sparse-representation}, the hierarchical diffusion model in \cref{sec:model}, and the final meshing process of a generated output in \cref{sec:meshing}.
\vspace*{2mm}

\subsection{3D Representation}
\label{sec:sparse-representation}

\paragraph{Explicit Multiscale 3D Features.}
Our method employs \emph{explicit} 3D information to encode the geometry of the input exemplar mesh at multiple scales. They are composed of merely 10 values per voxel of a sparse voxel grid,
\begin{equation}
\feat = (\vp_x, \vp_y, \vp_z, \vn_x, \vn_y, \vn_z, \vc_r, \vc_g, \vc_b, \mask),
\end{equation}
where \((\vp_x, \vp_y, \vp_z)\!\in\! [-0.5, 0.5]^3\) represents a point sample encoded as an offset from the voxel center, \((\vn_x, \vn_y, \vn_z)\!\in\! [-1, 1]^3\) represents the local normal of the underlying geometry associated to the point sample,
\((\vc_r, \vc_g, \vc_b)\!\in\! [-2, 2]^3\) represents the color scaled up to the \([-2, 2]\) range,
and \(\mask\!\in\! [-1, 1]\) is a scalar indicating if the voxel contains the mesh surface, which we will use as a mask to prune voxels after refinement (see \cref{sec:model}).
The value ranges of the feature components were empirically chosen since feature scale can be important in diffusion models~\cite{chen_importance_2023}. 

These features are extracted from the different scales of the input mesh.
Specifically, starting from the finest scale \(\maxLvl\), for each voxel that intersects the surface, we find the nearest surface point to the voxel center, whose position, normal, and color are used to form the feature. The mask is set to 1 for all selected voxels as they contain the surface.
At each subsequent scale \(\lvl\!<\!\maxLvl\), we obtain coarser points, normals, and colors features through a \(2^{3}\) average pooling, and the mask value through max pooling. To better preserve sharp features, we average the point positions in the Quadric Error Metric sense~\cite{maruani_ponq_2024} (see supplemental material for details).

This 3D representation yields three advantages:
\begin{itemize}
    \item it captures surface details in a compact form and carries contextual information from the texture;
    \item it encodes the 3D shape explicitly at each level, which enables a generated shape to be easily visualized or even edited at every level in a coarse-to-fine fashion;
    \item it is extracted from an input exemplar efficiently and deterministically, and will allow us to train each level of our model in parallel. 
\end{itemize}

\begin{algorithm}[t]
    \caption{Training at level \(\lvl\)}
    \label{alg:forward-diffusion}
    \KwIn{Extracted sparse features $\{\grid^{1}, \cdots, \grid^{\maxLvl}\}$}
    // Upsampler Training for Levels 2 to $\maxLvl$\\
    \Repeat{convergence}{
        Update model $\upsampler^{\lvl}$ with the loss \cref{eq:upsampler-loss}
    }
    // Diffusion Model Training\\
    \Repeat{convergence}{
        $t \sim \text{Uniform}(0, T)$\\
        $\rvepsilon \sim \mathcal{N}(0, \mathbf{I})$\\
        \eIf{$\lvl = 1$}{
            $\grid^{\lvl, \text{mix}} = \grid^{1} $ \\
        }{
            $\blurredGrid^{\lvl} = \upsampler^{\lvl}\left( \grid^{\lvl-1} \right)$\\
            $\grid^{\lvl, \text{mix}} = \gamma(t) \grid^{\lvl} + (1-\gamma(t)) \blurredGrid^{\lvl}$\\
        }
        $\grid_{t}^{\lvl} = \sqrt{\bar{\alpha}(t)} \grid^{\lvl, \text{mix}}  + \sqrt{1-\bar{\alpha}(t)} \rvepsilon$\\
        Update model with $\lVert\model^{\lvl}(\grid^{\lvl}_t | t) - \grid^{\lvl}\rVert^2.$
    }
\end{algorithm}

\paragraph{Sparse Voxel Grid.}
The inductive biases of convolutional neural networks exploit the shared information across internal crops within the input data, which is essential to learning from a single example and prevents overfitting~\cite{shaham2019singan,nikankin2023sinfusion,kulikov2023sinddm}.
However, 3D convolutional operations are notoriously expensive computationally and memory intensive.
To address this issue, we leverage fVDB~\cite{williams2024fvdb}, a recently proposed framework that supports efficient operations on sparse voxels. 
As a result, only active features are stored and processed, which significantly reduces the memory footprint and computational complexity.
We denote the sparse feature grid storing active 3D features at level \(\lvl\) as \(\grid^{\lvl} = \left\{ \feat^{\lvl} \right\}\).

\vspace*{2mm}
\subsection{Multiscale Diffusion}\label{sec:model}
Our multiscale diffusion pipeline generalizes SinDDM~\cite{kulikov2023sinddm} to 3D and adapts it to properly work with sparse voxel grid. 
As shown in \cref{fig:overview}, the pipeline consists of multiple diffusion models \(\{\model^\lvl\}_{1\leq \lvl \leq L}\).
During training, these diffusion models can be trained in parallel; at inference time, new variations are generated by sequentially running the reverse diffusion in a coarse-to-fine manner.
Below, we explain the hierarchical multi-scale diffusion and highlight our design differences compared to SinDDM.

\begin{algorithm}
    \caption{Sampling}
    \label{alg:reverse-diffusion}
    \KwIn{Choice of sampler $\mathcal{S} \in \{\text{DDPM}, \text{DDIM}\}$}
    \KwOut{Generated sparse grid $\grid_{\maxLvl}$}

     \emph{// Upsampler training for levels 2 to $\maxLvl$}\\
    $\grid_T^{1} \sim \mathcal{N}(0, \mathbf{I})$\\

    \For{$\lvl \leftarrow 1$ \KwTo $\maxLvl$}{
        \If{$\lvl > 1$}{
            $\rvepsilon \sim \mathcal{N}(0, \mathbf{I})$\\
            $\blurredGrid^{\lvl} = \upsampler^{\lvl}(\grid^{\lvl-1})$\\
            $\grid_{T[\lvl]}^{\lvl} =  \sqrt{\bar{\alpha}(T[\lvl])} \blurredGrid^{\lvl}  + \sqrt{1-\bar{\alpha}(T[\lvl])} \;\rvepsilon$\\
        }
        \For{$t \leftarrow T[\lvl]$ \KwTo $1$}{
            $\rvepsilon \sim \mathcal{N}(0, \mathbf{I})$\\
            $\grid_{t-1}^{\lvl} = \mathcal{S}(\model^l, \grid_{t}^{\lvl},\bar{\alpha}, \rvepsilon, t)$\\
        }
        $\grid^{\lvl} = \textbf{Prune}(\grid_{0}^{\lvl})$\\
    }
\end{algorithm}

\vspace*{-4mm}
\paragraph{Forward Multiscale Diffusion.}
Except at its coarsest level \(\model^1\), our diffusion model $\model^{\lvl>1}$ generates the signal of the current level based on the output of the previous \((\lvl-\!1)\) level. 
This initial guess is obtained by upsampling the output from the previous level \(\smash{\blurredGrid{\vphantom{\grid}}^{l} = \upsampler\bigl( \grid^{l-1} \bigr)}\), which can be seen as a ``blurred'' version of \(\grid^{\lvl}\). 
This means that, for $l\!>\!1$, the diffusion model not only needs to denoise but also ``deblur" during sampling.
As a result, SinDDM modifies the forward diffusion process to be 
\vspace*{-2mm}
\begin{equation}
    \grid^{\lvl}_t = \!\sqrt{\bar{\alpha}(t)} \left( \gamma(t) \grid^{\lvl} \!+\! (1\!-\!\gamma(t)) \blurredGrid^{\lvl} \right) \!+\! \sqrt{1\!-\!\bar{\alpha}(t)}\,\rvepsilon, \vspace*{-2mm}
\end{equation}
where \(\rvepsilon \!\sim\! \mathcal{N}(\vzero, \mI)\), while \(\bar{\alpha}\left( t \right)\) and \(\gamma\left( t \right)\) are monotocally decreasing functions from 1 to 0 as \(t\) grows from 0 to \(T\). 
The model learns to denoise the corrupted feature \(\feat^{\lvl}_t\) at time step \(t\) by minimizing the reconstruction loss 
\begin{equation}
    \lVert\model^{\lvl}(\grid^{\lvl}_t | t) - \grid^{\lvl}\rVert^2. \label{eq:diffusion-loss}
\end{equation}


\noindent Contrasting with SinDDM which employs a bilinear upsampler as \(\upsampler\), we use a level-specific upsampler \(\upsampler^{\lvl}\) motivated by the fact our spatial features (points and normals) are extracted by projecting the voxel centers onto the mesh surface --- thus potentially exhibiting abrupt local changes.
This results in improved preservation of sharp geometric features as we show in \cref{sec:experiments}.
The upsampler \(\upsampler^{\lvl}\) is trained to minimize the \(\normltwo\)-loss between upsampled and ground-truth features, \ie, 
\begin{equation}
\lVert \upsampler^{\lvl}(\grid^{\lvl-1}) - \grid^{\lvl} \rVert^2. \label{eq:upsampler-loss}
\end{equation}


Crucially, the training of different levels can be parallelized. For each level \(l>1\), we first train the upsampler and the diffusion model as summarized in \cref{alg:forward-diffusion}.

\noindent Unlike SinDDM, training needs to accommodate our use of sparse grids.
When comparing the denoised sparse feature grid and the ground-truth sparse feature grid (\cref{eq:diffusion-loss}), the denoised grid can contain more active voxels (see dark voxels in \cref{fig:overview}, even though their mask could be -1 --- yet
fVDB operations on two sparse feature grids assume that they have the same active voxels. 
To solve this problem, we flood those inactive voxels in the ground-truth \(\grid^{\lvl}\) with feature values of the nearby active cells using a blurring kernel. All features except the mask value are flooded in this way, whereas the mask value is set to -1. Empirically, we observe that soft blending the feature values this way (instead of hard setting the values to an arbitrary number or applying an additional mask for loss) achieves the best result. 

\paragraph{Reverse Multiscale Diffusion.}
Once trained, we can apply standard DDPM~\cite{ho2020denoising} or DDIM~\cite{song2021denoising} sampling sequentially from levels 1 to \(\maxLvl\).
As outlined in \cref{alg:reverse-diffusion}, we start from a noise \(\rvepsilon\!\sim\! \mathcal{N}\left(\vzero, \mI\right)\)  and run the reverse sampling to obtain an initial prediction at the coarsest level.
Then, for each level, we first prune the predicted inactive voxels from the previous level by removing any feature entries with mask value $m \!<\! 0$.
The resulting feature grid is then upsampled with \(\upsampler^{l}\), and subsequently corrupted with noise, before being given to the diffusion model for reverse sampling.
Similar to SinDDM, we only add noise up to timestep $T[l] \!<\! T$ to prevent destroying the prediction from the previous level. 
A schematic overview of the sampling process is illustrated in \cref{fig:overview}. 





\subsection{Meshing}\label{sec:meshing}
Once a new geometric variant has been created, we can directly visualize the generated shape using the points (one per finest voxel in the fVDB data structure) along with their associated normal and color. We can also trivially generate a mesh of the geometry through Poisson reconstruction~\cite{kazhdan_screened_2013} (or APSS~\cite{guennebaud_algebraic_2007} if we are dealing with open surfaces). One can assign colors to the mesh nodes based on the output colors, bake texture maps (as used sporadically in figures), or further refine and stylize the texture with off-the-shelf image enhancement models (see \cref{sec:exp-texture}).


\section{Implementation Details}\label{sec:implementation}
\paragraph{Implementation Details.}
We implemented our method in Python with PyTorch~\cite{paszke_pytorch_2019}, libigl~\cite{jacobson_libigl_nodate}, and Open3D~\cite{zhou2018open3d}, and our code is available on \href{https://nissmar.github.io/projects/shapeshifter/}{our project page}. All reported timings were obtained on a desktop with an NVIDIA GeForce RTX 3080 GPU (10 GB) to underscore the efficiency of our approach even on consumer-grade hardware.

\begin{table*}[t]
    \centering
    \resizebox{1.\linewidth}{!}{\begin{tabular}{ l |c|  c  c  c  c  c  c  c  c | c }\toprule
Metric & Method & acropolis & canyon & fighting-pillar & house & ruined-tower & small-town & stone-cliff & wood & average\\\midrule

\multirow{3}{*}{G-Qual. $\downarrow$} & Sin3DGen  &  4.83   &    6.16   &    8.45   &    17.95   &    6.98   &    4.02   &    13.02   &    10.32 &  8.97\\
&Sin3DM  &  0.92   &    2.23   &    \textbf{0.26}   &    2.01   &    0.49   &    \textbf{0.84}   &    \textbf{0.02}   &    0.09 &  0.86\\
&\ourmethod{}  &  \textbf{0.01}   &    \textbf{0.21}   &    \textbf{0.26}   &    \textbf{0.01}   &    \textbf{0.11}   &    1.00   &    0.10   &    \textbf{0.02} &  \textbf{0.21}\\

\midrule
\multirow{3}{*}{G-Div. $\uparrow$} & Sin3DGen  &  \textbf{0.25}   &    \textbf{0.50}   &    \textbf{0.59}   &    \textbf{0.41}   &    \textbf{0.86}   &    \textbf{0.70}   &    \textbf{0.65}   &    \textbf{0.44} &  \textbf{0.55}\\
&Sin3DM  &  0.12   &    0.17   &    0.15   &    0.01   &    0.21   &    0.60   &    0.32   &    0.10 &  0.21\\
&\ourmethod{}  &  0.04   &    0.19   &    0.24   &    0.01   &    0.32   &    0.60   &    0.23   &    0.08 &  0.21\\

\bottomrule

\end{tabular}

}
    \vspace*{-3mm}
    \caption{\emph{Evaluating geometric quality and diversity using SSFID and pairwise IoU scores.} Our model shows clear advantage in quality, and performs similar to Sin3DM in diversity. As we discussed in \cref{sec:comparison}, both metrics have their blindspots, SSFID overlooks geometric details and pairwise IoU rewards artifacts. Finding a more holistic metric to evaluate shape variation remains an open problem.  \vspace*{-2mm}} 
    \label{tab:ssfid}
\end{table*}

\paragraph{Model Parameters.}
By default we use 5 levels, the lowest and highest grid resolutions being 16 and 256 respectively.
The upsamplers \(\upsampler\) consist of 4 layers of 64 channels, containing {\mytilde}55k parameters that are trained for 10k iterations with a learning rate of $5 \!\cdot\! 10^{-4}$ and a 5\% dropout rate.
The diffusion models have 128 feature channels and 7 layers, for a total of {\mytilde}565k parameters for the coarsest model $\model^{1}$ and 1.2M parameters As in prior work, the receptive fields of each model \(\model^{1}\) are kept small to prevent overfitting to the fixed global structure: \(\model^{1}\) thus uses a receptive field of $5^3$, while the rest use $9^3$.
We train our diffusion models with $T\!=\!1000$ diffusion steps.
For sampling, we set $T[1]\!=\!1000$ and $T[l\!>\!1]\!=\!300$. 
$\model^{1}$ is trained for 20,000 iterations, and the rest for 40,000 iterations. 
All levels are trained with random crops of the same resolution to help ensure that each scale is trained in roughly the same time, and we use a learning rate of $10^{-4}$ with a 1\% dropout rate.

\paragraph{Feature Extraction.}
In terms of shape processing, we normalize each mesh to fit within $[-1, 1]^3$.
3D features are sampled at a resolution of $1024^3$ and downsampled to a coarsest resolution of $16^3$ voxels, as described in \cref{sec:sparse-representation}.

\section{Experiments}\label{sec:experiments}

\noindent\textbf{Data.} We demonstrate our approach on 3D textured exemplars provided by Sin3DM (from \cite{impjive2021pillar,
kurd2021akropolis,
ashoori2020smalltown,
ustal2020canyon,
carnota2015industrial,
djmaesen2021cliff,
allaboutblender2020wood}), and also used an open surface example that we created.
Note that \ourmethod can operate as-is on untextured inputs; but colors can help distinguish geometrically similar, yet semantically different parts of the geometry. 

\subsection{Comparison.}\label{sec:comparison}
\paragraph{Baselines.}
We compare with two state-of-the-art papers on 3D generation from single examples: Sin3DM~\cite{wu2024sindm} and Sin3DGen~\cite{li2023patch}.
Sin3DM uses a single-scale triplane diffusion model with a small receptive field to learn internal feature distribution within the exemplar shape. Features are learned in a separate autoencoder that parameterizes the input shape as an implicit neural surface~\cite{wang2021neus}. We use their publicly available generated results for comparison.
Sin3DGen operates instead on radiance field represented by plenoxels: it learns a hierarchical deformation field to transform the input plenoxels based on patch similarity. Following their data preparation guideline, we first rendered 200 images with Blender~\cite{blender2018}, then trained a \(512^3\) plenoxel to obtain the input exemplar which we provide to Sin3DGen to generate results to which we compare ourselves. 

\vspace*{-4mm}
\paragraph{Quantitative Evaluations.}
Following prior work, we use Single Shape Fréchet Inception Distance (SSFID~\cite{wu2022learning}) to evaluate the output quality, and pairwise \((1\!-\!\text{IoU})\)-distances among 10 generated variations to evaluate the output diversity.
SSFID is extended from Single-Image Fréchet Inception Distance (SIFID)~\cite{shaham2019singan}, which compares the statistics of the input and the generation feature extracted at different levels of a pretrained multiscale 3D encoder~\cite{chen2021decor} trained on voxelized shapes from ShapeNet;
thus we voxelized input and generations at a $256^3$ resolution for evaluation purposes. We show in \cref{tab:ssfid} that \ourmethod{} outperforms competing methods in SSFID. (Note that our SSFID scores for Sin3DM differ from their reported scores as they used post-processed smoothed shapes as reference; please refer to the supplementary material for details.)  
While SSFIDs usually match the perceptual quality of the results, we note that the voxelization step used for scoring removes sharp features and high-frequency details, which our paper captures particularly well. 
The qualitative examples shown in \cref{fig:results} and in our supplementary material better demonstrate our strengths. Overall, the underlying geometry from Sin3DGen results is heavily distorted as it relies on plenoxel matching and blending, and while the SDF supervision from Sin3DM makes for better outputs, it does not capture sharp features well and the features extracted from their single-resolution triplane representation struggle to encode high-frequency details.
Finally, the diversity scores based on IoU must be handled with care as it rewards artifacts. Our results show low diversity scores for very structured exemplars like the acropolis or the house, but good scores for more organic and varied shapes like the canyon or the small town, which is in line with our goal of generating variants of the input shapes. 

\begin{table}[h] \vspace*{-1mm}
    \centering
    \begin{tabular}{lcccc}\toprule
       Method  & encoding and training & inference \\\midrule
       SSG &  4 hours & 0.1 sec \\
       Sin3DM  & 2.5 hrs & 15.8 sec \\
       Sin3DGen$\ast$ & 15 min & 139 sec \\
       \ourmethod{} & 12 min & 10.7 sec\\\bottomrule
    \end{tabular} \vspace*{-2mm}
    \caption{\emph{Timings}. Sin3DGen~\cite{li2023patch} was tested on a more performant GPU (Nvidia A100 40GB), as it could not fit in our regular GPUs (Nvidia 3080 10GB).\vspace*{-5mm}}
    \label{tab:timing}
\end{table}

\vspace*{-4mm}
\paragraph{Training and inference speed.}
We show timing comparisons in ~\cref{tab:timing}, where we also included GAN-based SSG~\cite{wu2022learning} as its architecture is faster than its successor at inference time, albeit with lower quality and diversity~\cite{wu2024sindm}. 
Despite its impressive inference time, the training time of SSG is the longest.
Sin3DGen does not require training as it is based on patch-matching; but each variation generation takes around 2 minutes, limiting interactive use cases.
Sin3DM takes in total of 2.5 hours to train, of which {\mytilde}30 minutes are used to learn triplane features.
Our method takes merely 6 seconds to encode the shape, and trains each level of the hierarchy of diffusion models in 12 minutes, significantly outperforming all other trained models. The inference time takes from 0.15 seconds (at level 2) to 7.5 seconds (at level 5) totaling 10.7 seconds.
It is worth mentioning that since our method outputs colored oriented pointset at \emph{every level}, requiring no additional conversion \eg marching cubes in other methods, we can flexibly choose the working level depending on the application. For example, in editing, we can operate on level 3, which takes only 1 s. per generation, thus enabling interactive modeling as shown next. More details are provided in the supplementary material.

\begin{figure}[!h] \vspace*{-2mm} 
    \centering
  \includegraphics[width=0.99\linewidth]{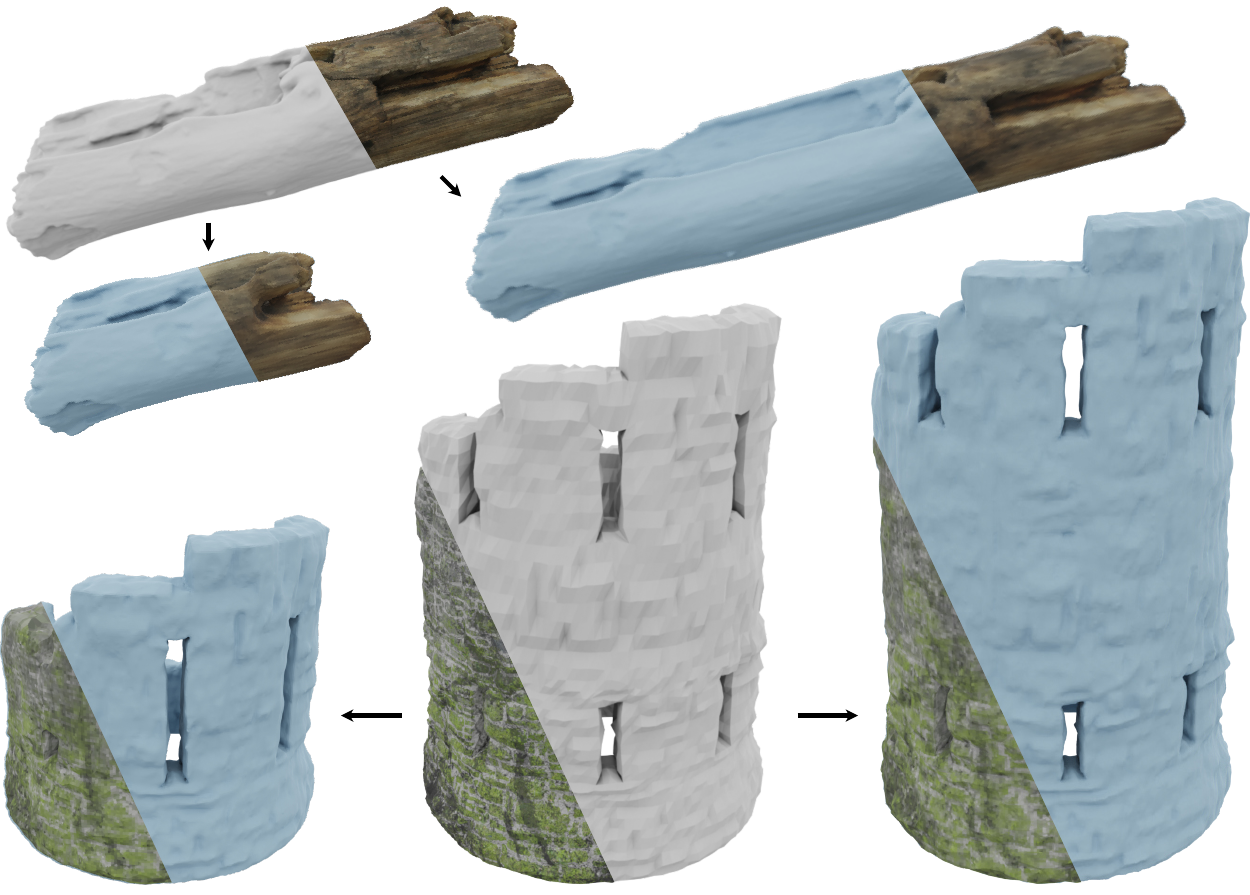}
  \vspace*{-3mm}
    \caption{\emph{Controlled generation.} The span of the output can be trivially controlled by resizing the initial grid anisotropically. \vspace*{-4mm}}
    \label{fig:bbox_control}
\end{figure}

\subsection{Control and editing.}
\begin{figure}[!h] \vspace*{-2mm}
    \centering
  \includegraphics[width=0.99\linewidth]{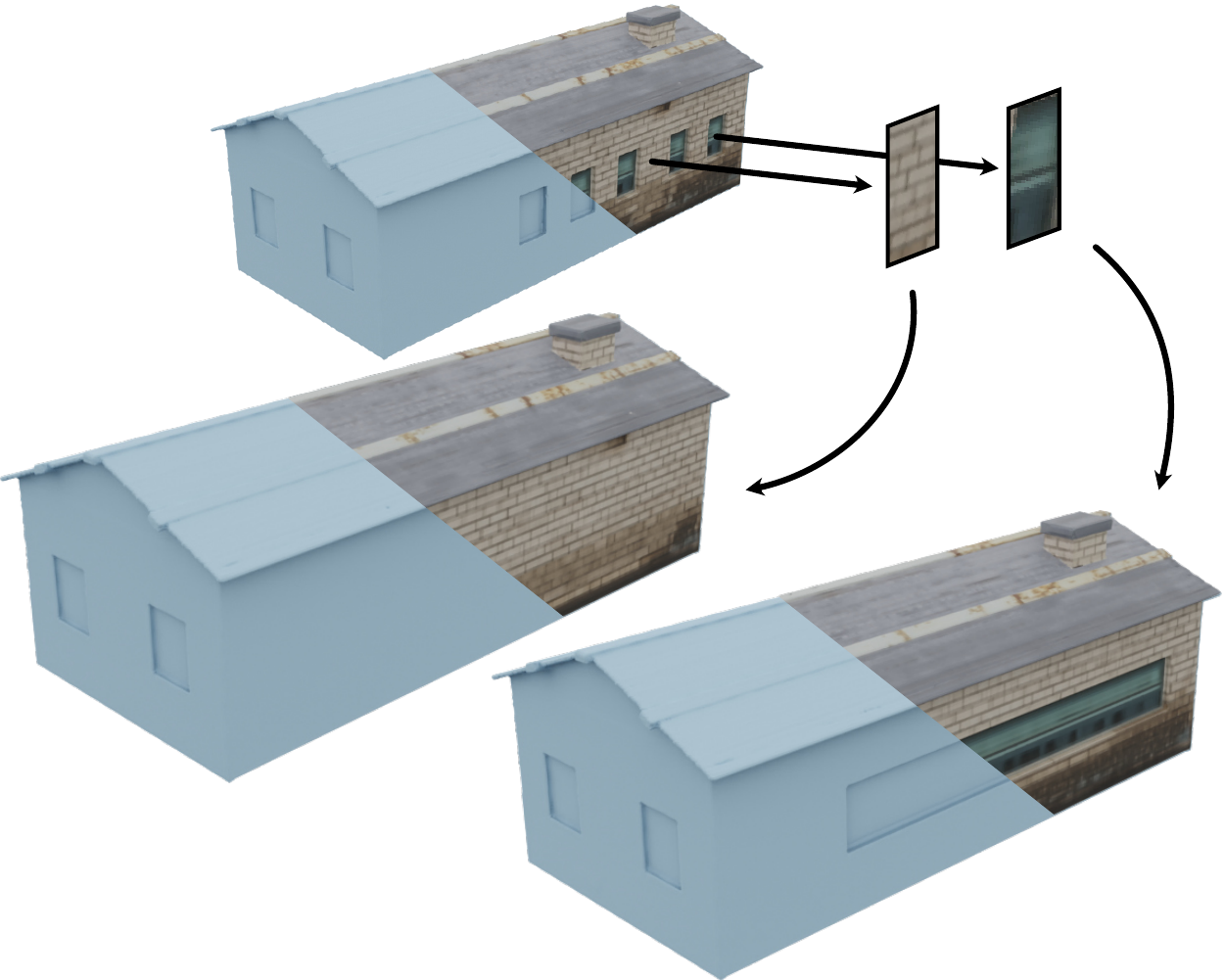}
   \vspace*{-2mm}
    \caption{\emph{Editing.} Using sparse voxel grid allows users to intuitively apply more precise edits. Here, a user can copy and paste a selected part of a generated variation at an intermediate level to manually alter the variation.
     \vspace*{-4mm}
     }
    \label{fig:duplication_control}
\end{figure}

Our multiscale explicit representation makes it easy to control and edit the output.
We demonstrate two examples: the user can trivially  change the span of the model by resizing the initial grid \(\grid^{0}_{T}\) anisotropically, see \cref{fig:bbox_control}; moreover, \cref{fig:duplication_control} demonstrates that a generated output can be further edit by copy-and-pasting  parts of the output within one of the levels of the multiscale description of the shape, here to remove windows or adding a bay window.
While existing works can offer similar capabilities, their use of triplane features or deformation fields to drive the generation renders editing less intuitive.
For example, a patch from the input shape can be duplicated in Sin3DM to appear in the generated variations; however, the duplication must be done on three interdependent triplanes features, which do not directly correspond to a feature in 3D space.

\subsection{Open Surfaces}
\begin{figure}[!h] \vspace*{-2.5mm} \centering
  \includegraphics[width=0.95\linewidth]{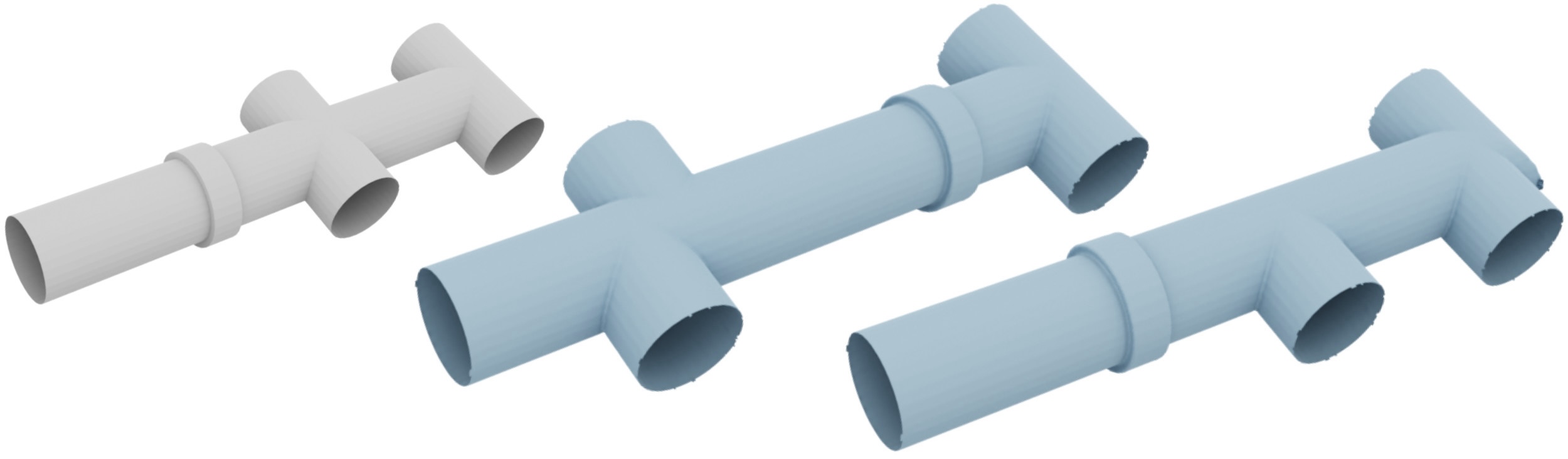}
  \vspace*{-3mm}
    \caption{\emph{Open surfaces.} Our oriented points representation also handles open surfaces by simply using APSS~\cite{guennebaud_algebraic_2007} to mesh the generated point set, while it is challenging for existing methods.\vspace*{-3mm}}
    \label{fig:opensurfaces}
\end{figure}

Our use of points and normals to represent the geometry makes the treatment of open surfaces not only possible but just as simple as the case of closed surfaces: only the surface extraction method needs to be altered, \ie with APSS~\cite{guennebaud_algebraic_2007} instead of \cite{kazhdan_screened_2013}. 
An example is shown in \cref{fig:opensurfaces}.

\subsection{Ablation studies}\label{sec:ablation}
\begin{figure}[!h] 
    \centering
    \centering
    \begin{subfigure}{.48\linewidth}
      \centering
      \includegraphics[width=\linewidth]{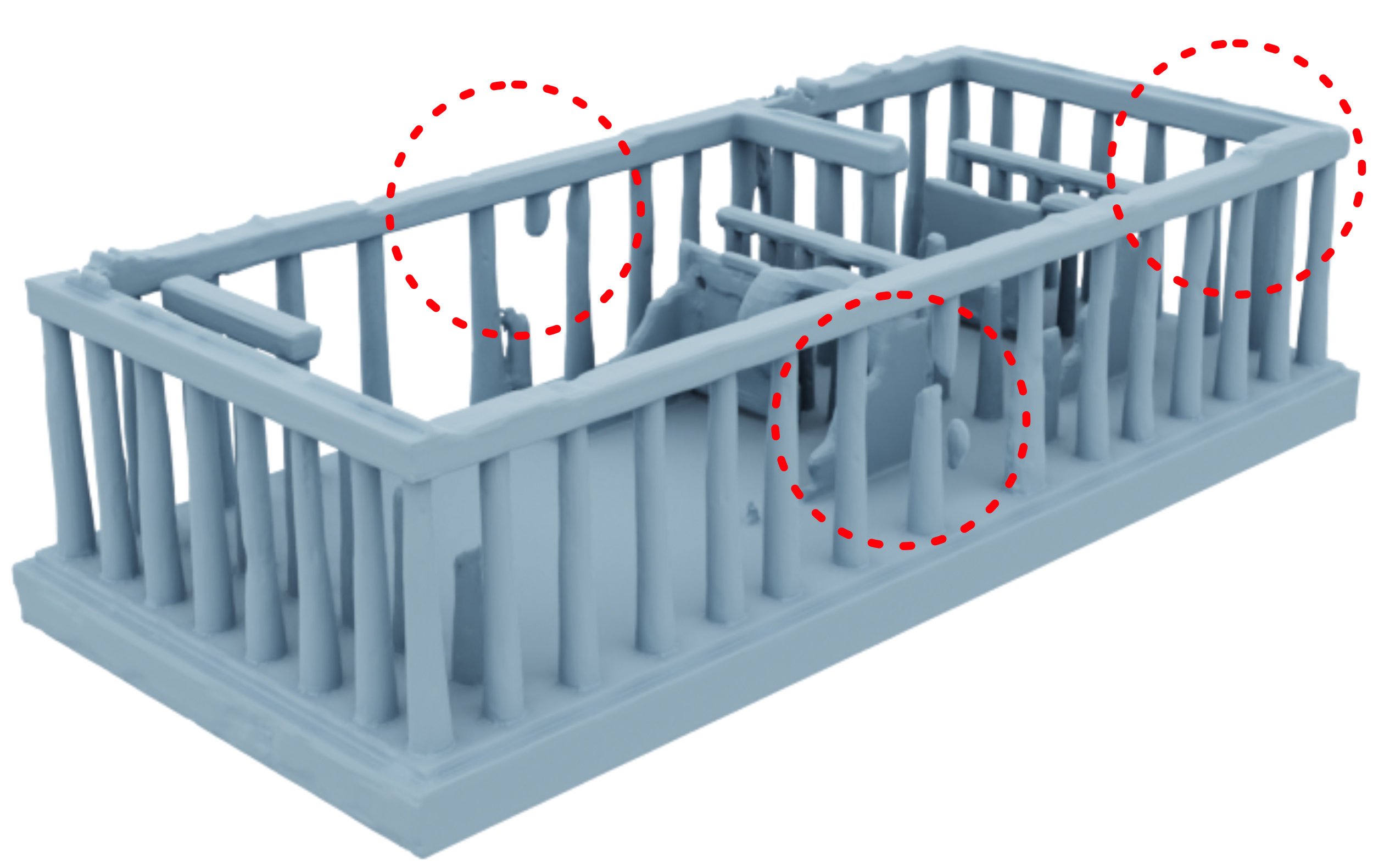}
    \end{subfigure}
     \begin{subfigure}{.48\linewidth}
      \centering
      \includegraphics[width=\linewidth]{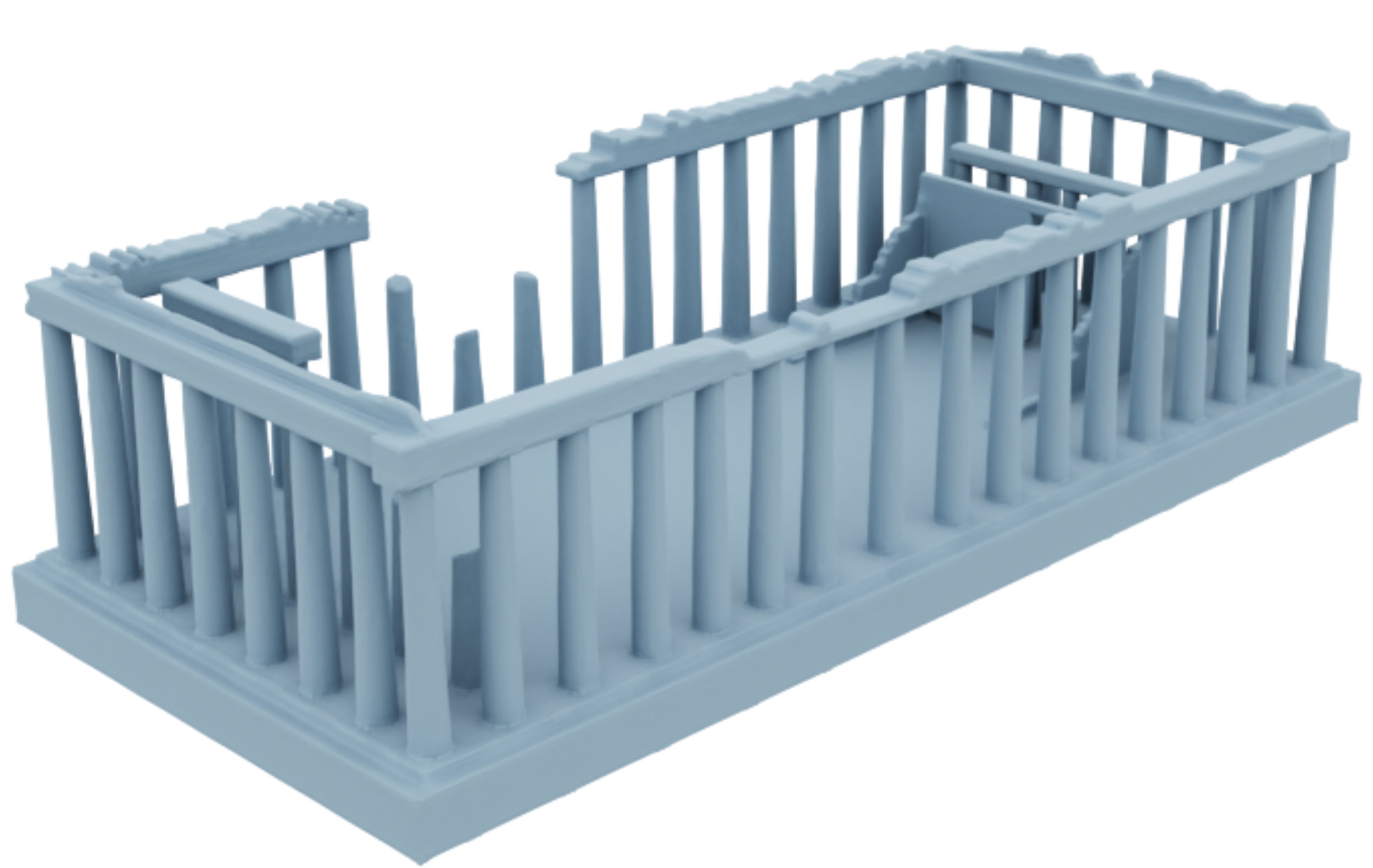}
    \end{subfigure}

    \centering
    \begin{subfigure}{.48\linewidth}
      \centering
      \includegraphics[width=\linewidth]{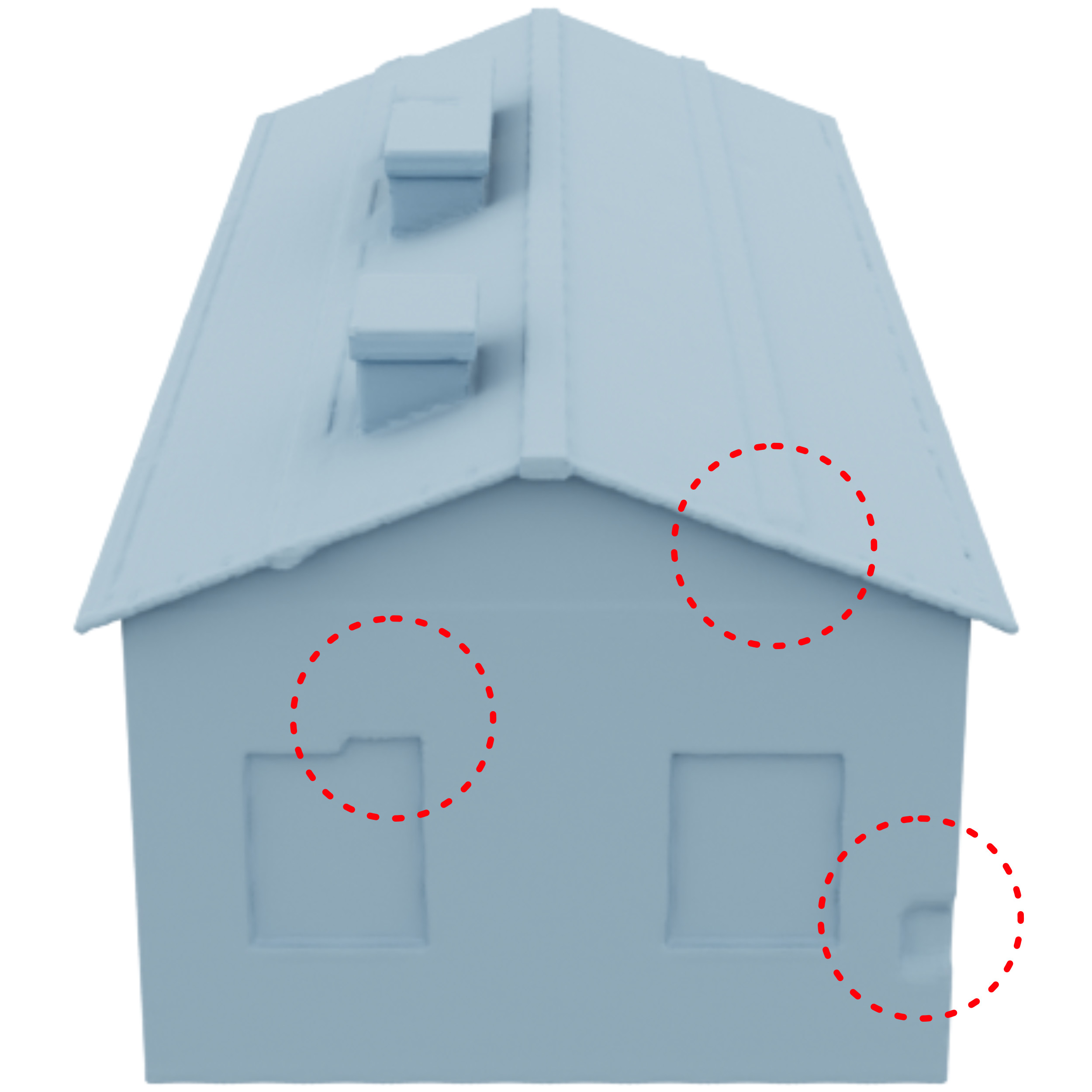}
      \caption{Trilinear Interpolation}
    \end{subfigure}
     \begin{subfigure}{.48\linewidth}
      \centering
      \includegraphics[width=\linewidth]{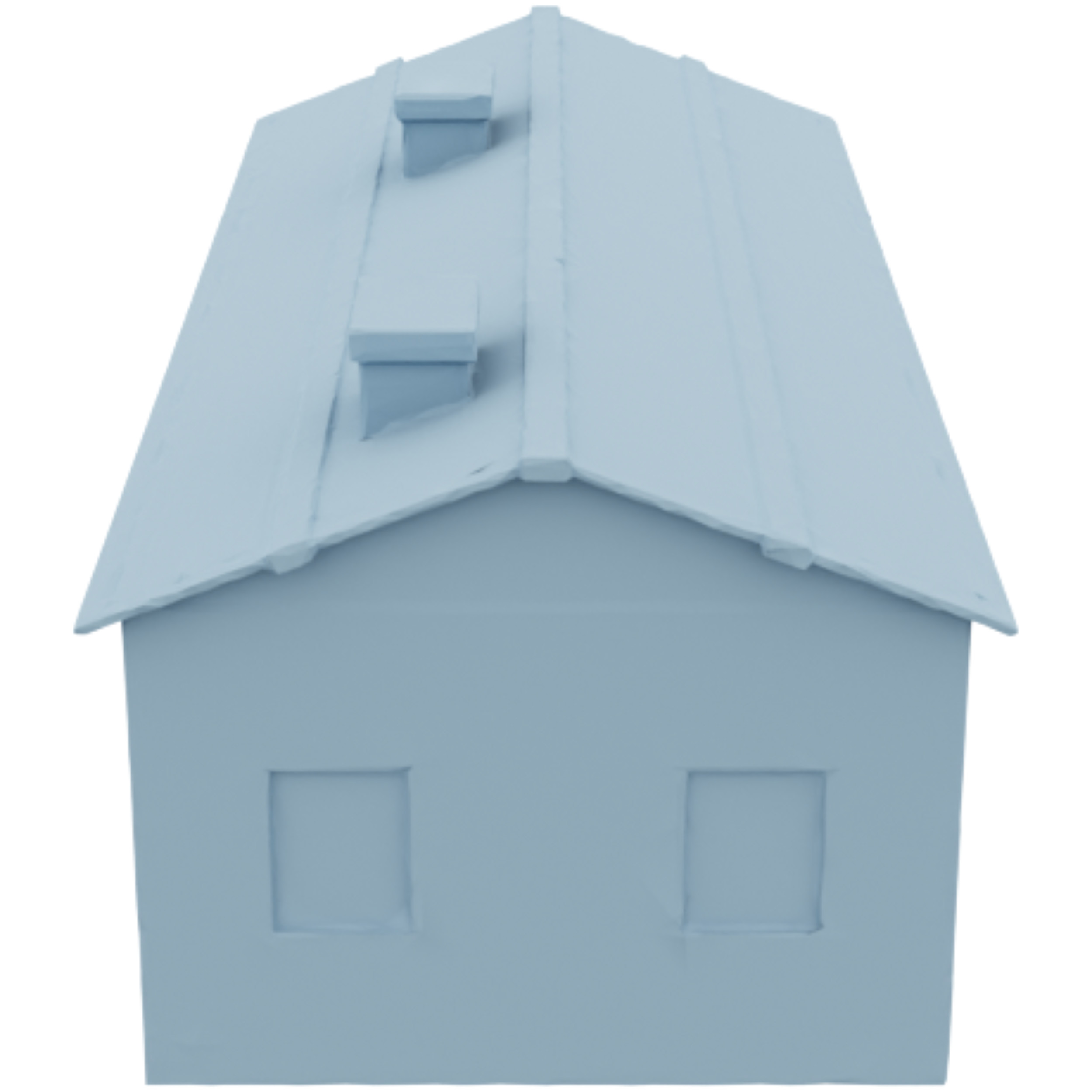}
      \caption{Learned Upsampler}
    \end{subfigure}
    \vspace*{-2mm}
    \caption{\emph{Ablation test of upsampling.} Comparing trilinear interpolation (left) with learned upsampler (right), the interpolation causes artifacts (see circled areas), whereas the learned upsampler provides a more detailed and structurally coherent output.\vspace*{-4mm}}
    \label{fig:ablation_upsampler}
\end{figure}

\paragraph{Learned Upsampler.}
We demonstrate the benefit of our learned upsampler by replacing it (both in training and inference) by a trilinear interpolation as used in SinDDM in \cref{fig:ablation_upsampler}: artifacts appear as our point features are mispositioned because of the trilinear upsampling.

\begin{figure}[h]

    \begin{subfigure}{.48\linewidth}
      \centering
      \includegraphics[width=\linewidth]{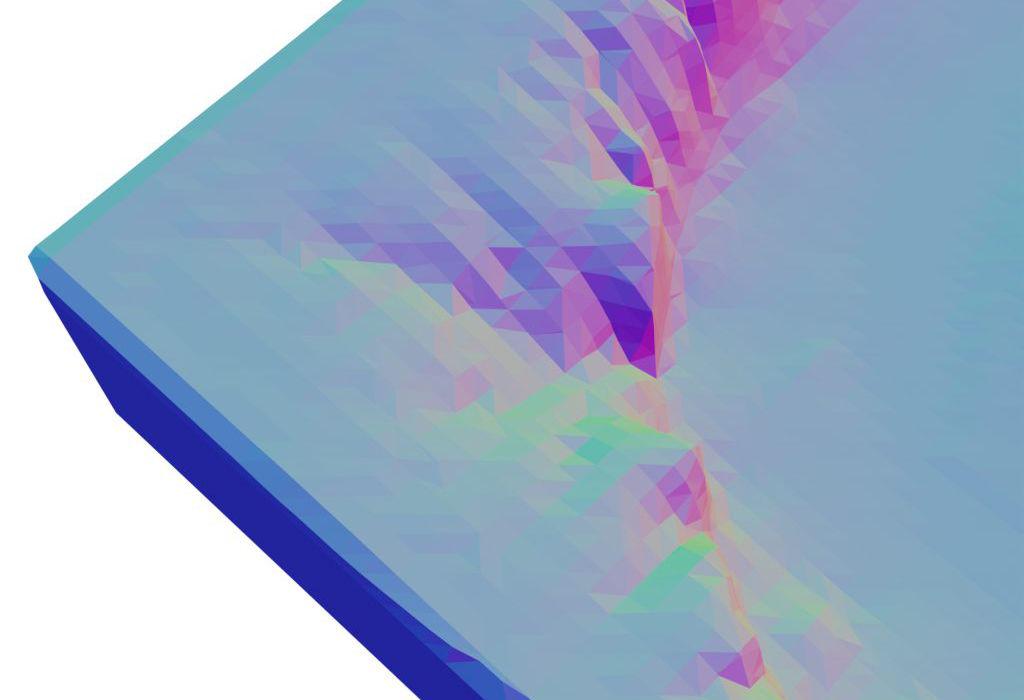}
      \caption{SDF + Marching Cube~\cite{lorensen1998marching}}
    \end{subfigure}
     \begin{subfigure}{.48\linewidth}
      \centering
      \includegraphics[width=\linewidth]{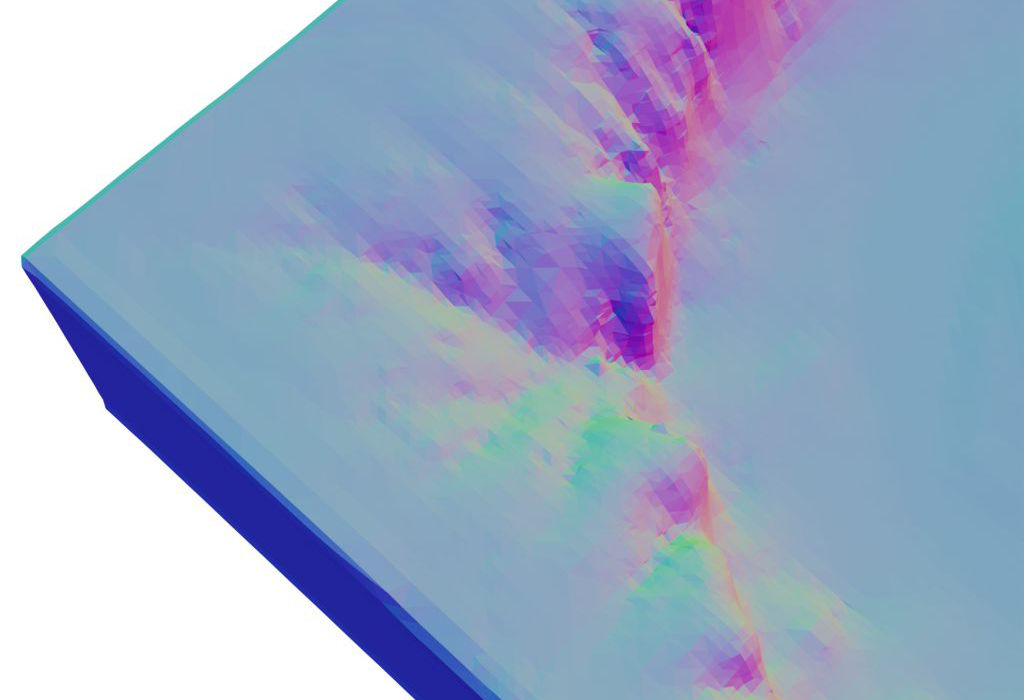}
       \caption{Ours + Poisson~\cite{kazhdan_screened_2013}}
    \end{subfigure}
    \vspace*{-2mm}
    \caption{\emph{Ablation test of features.} Comparing SDF (left) with our proposed point and normal features (right) at the same resolution ($128^3$) demonstrates that our proposed features produce richer geometric details. The mesh color encodes the normal direction to reflect the difference in geometry details.\vspace*{-4mm}} 
    \label{fig:ablation_feature}
\end{figure}

\vspace*{-2mm}
\paragraph{3D Features.} We also replace our geometric features with an SDF. For fairness of comparison, we use two layers of active cells (instead of one) close to the surface to compensate for the reduced feature dimensionality.
For the same input resolution, our features have more details (\cref{fig:ablation_feature}).

\begin{figure}[b] \vspace*{-3mm}
    \centering
    \includegraphics[width=0.96\linewidth]{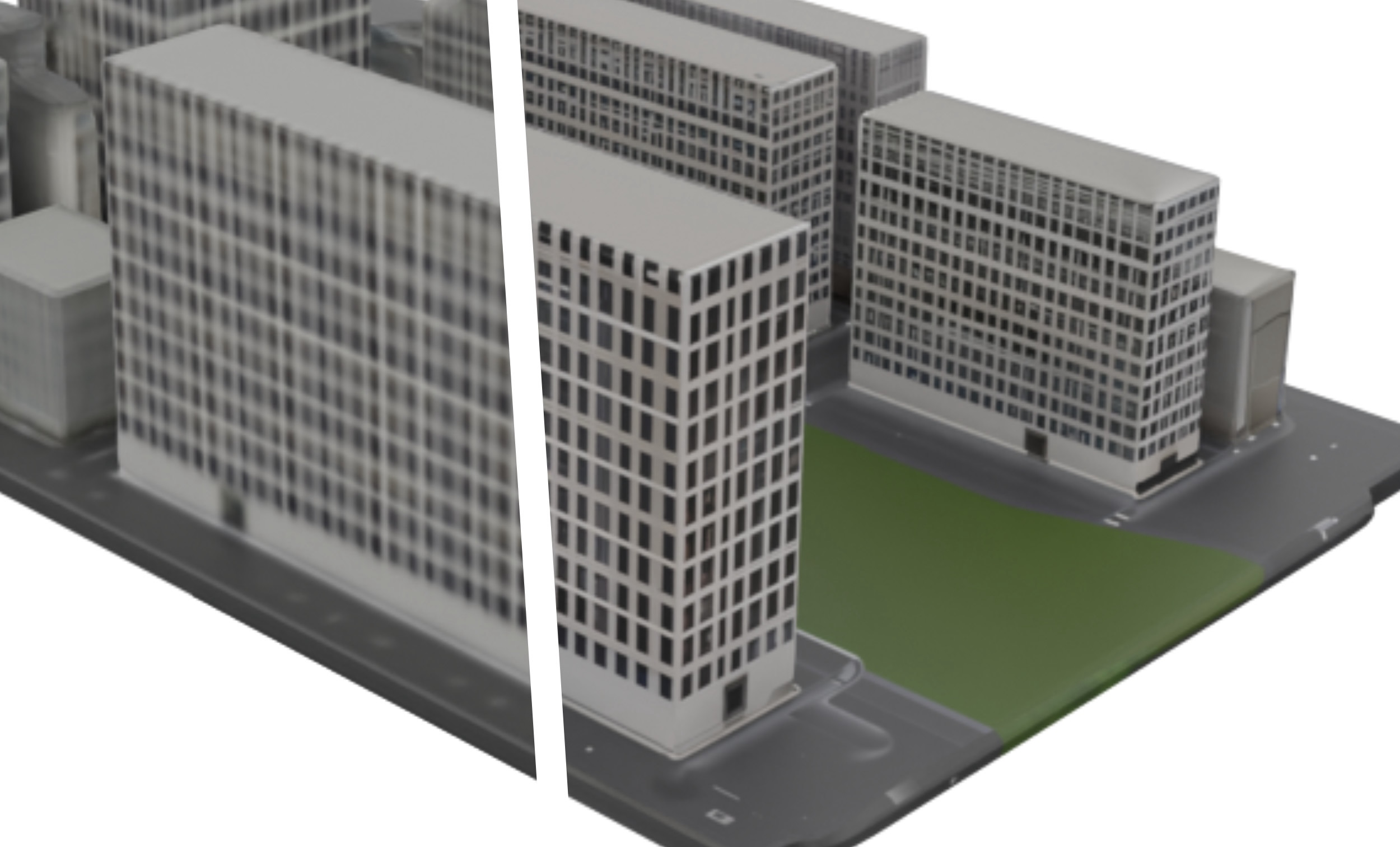}
       \vspace*{-2mm}
    \caption{\emph{Texturing.} As explained \cref{sec:meshing}, ultra-high resolution texture can be obtained by applying state-of-the-art AI image-enhancing tools on the texture maps created by our method from the colored point set outputs.\vspace*{-6mm}}
    \label{fig:textured-example}
\end{figure}

\subsection{Texture Augmentation}\label{sec:exp-texture}
Finally, we show that one can texture our generated models by applying contemporary image super-resolution on the baked texture maps in \cref{fig:textured-example}: using Magnific AI~\cite{magnific_ai} for example can efficiently generate a fine texture improving the visual impact of our results. While this is only a proof-of-concept example, exploring the texturing of our geometric models is an exciting, albeit orthogonal, research direction.

\section{Limitations and Future Work}
\label{sec:limitations}

Just like previous exemplar-based generative methods, \ourmethod is limited in the shape variations it can generate: while our approach is not strictly patch-based, it is similarly restricted in its ability to consider widely different variants. 
Extending its range of alterations through data augmentation or more involved (equivariant) features remains an intriguing possibility that would broaden the applicability of our method. Moreover, we focused our approach on generating high-quality, detailed geometry and did not consider fine texture generation. While existing exemplar-based methods have proposed approaches to generate textures for meshes that we could apply as-is, we believe there may be other exciting possibilities to explore, such as fitting 2D Gaussian splats~\cite{Huang2DGS2024} within our finest voxels to enrich our geometry with radiance field reconstruction.

Now that we have proven the efficacy of explicit geometry encoding through colored points and normals for creating shapes in our exemplar context, it would be interesting to study its adequacy in the more general case of generative modeling from large datasets: its lightweight, surface-based nature may circumvent a number of issues plaguing current state-of-the-art approaches. 

\section{Conclusion}
\label{sec:conclusion}
We proposed a novel generative approach for generating high-quality and detailed 3D models from a single exemplar. Our approach stands out as the first 3D generative method based on an explicit encoding of geometry through points, normals, and optionally colors. Combined with sparse voxel grid, we demonstrated that both training and inference times are (at times drastically) reduced compared to previous methods, despite a significantly improved quality of our geometric outputs and an ability to deal seamlessly with closed or open surfaces alike. We thus believe that \ourmethod sets a new standard for the quality of geometric outputs in generative modeling. 

\section{Acknowledgments}
This work was supported by the French government through the 3IA Cote d’Azur Investments in the project managed by the National Research Agency (ANR-23-IACL-0001), Ansys, Adobe Research, and a Choose France Inria chair. \vspace*{-2mm}

{
    \small
    \bibliographystyle{ieeenat_fullname}
    \bibliography{main}
}

\clearpage
\maketitlesupplementary

This supplementary material provides additional details, results, and comparisons.

\begin{figure}[h] \vspace*{-2mm}
\centering
        \begin{subfigure}{.32\linewidth}
        \centering
        \includegraphics[width=\linewidth]{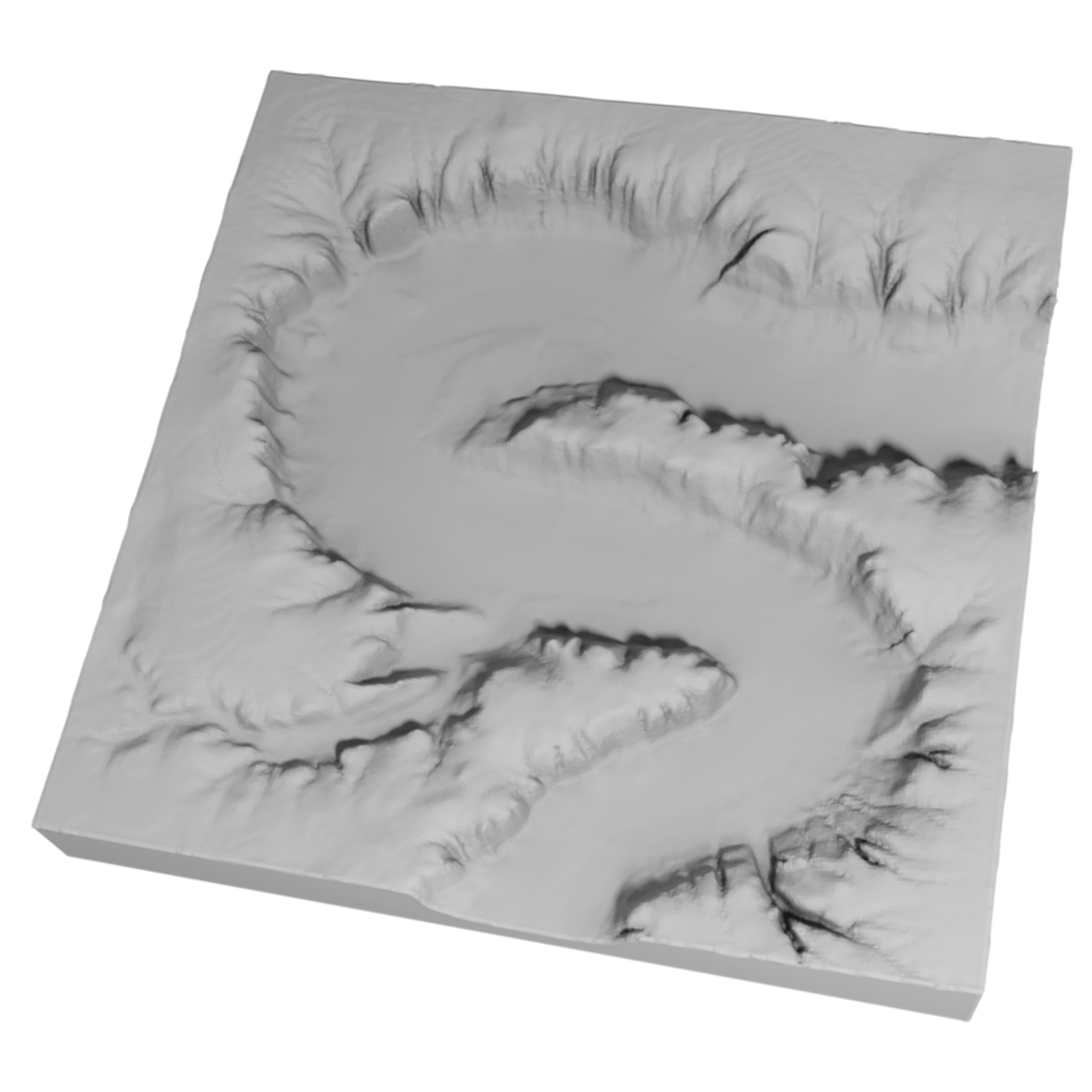}
        \caption{Input Geometry}
    \end{subfigure}
    \begin{subfigure}{.32\linewidth}
        \centering
        \includegraphics[width=\linewidth]{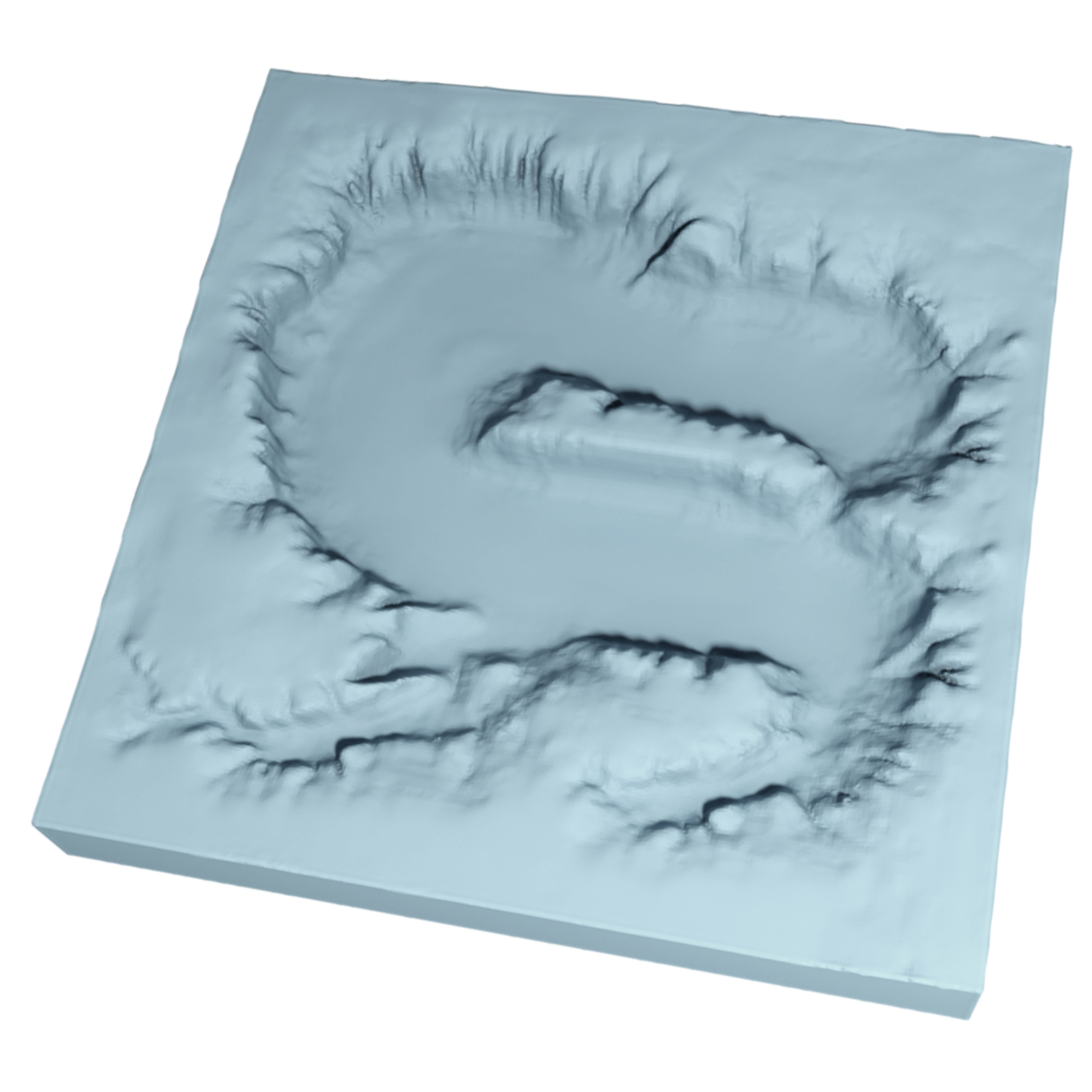}
        \caption{Ours}
    \end{subfigure}
    \begin{subfigure}{.32\linewidth}
        \centering
        \includegraphics[width=\linewidth]{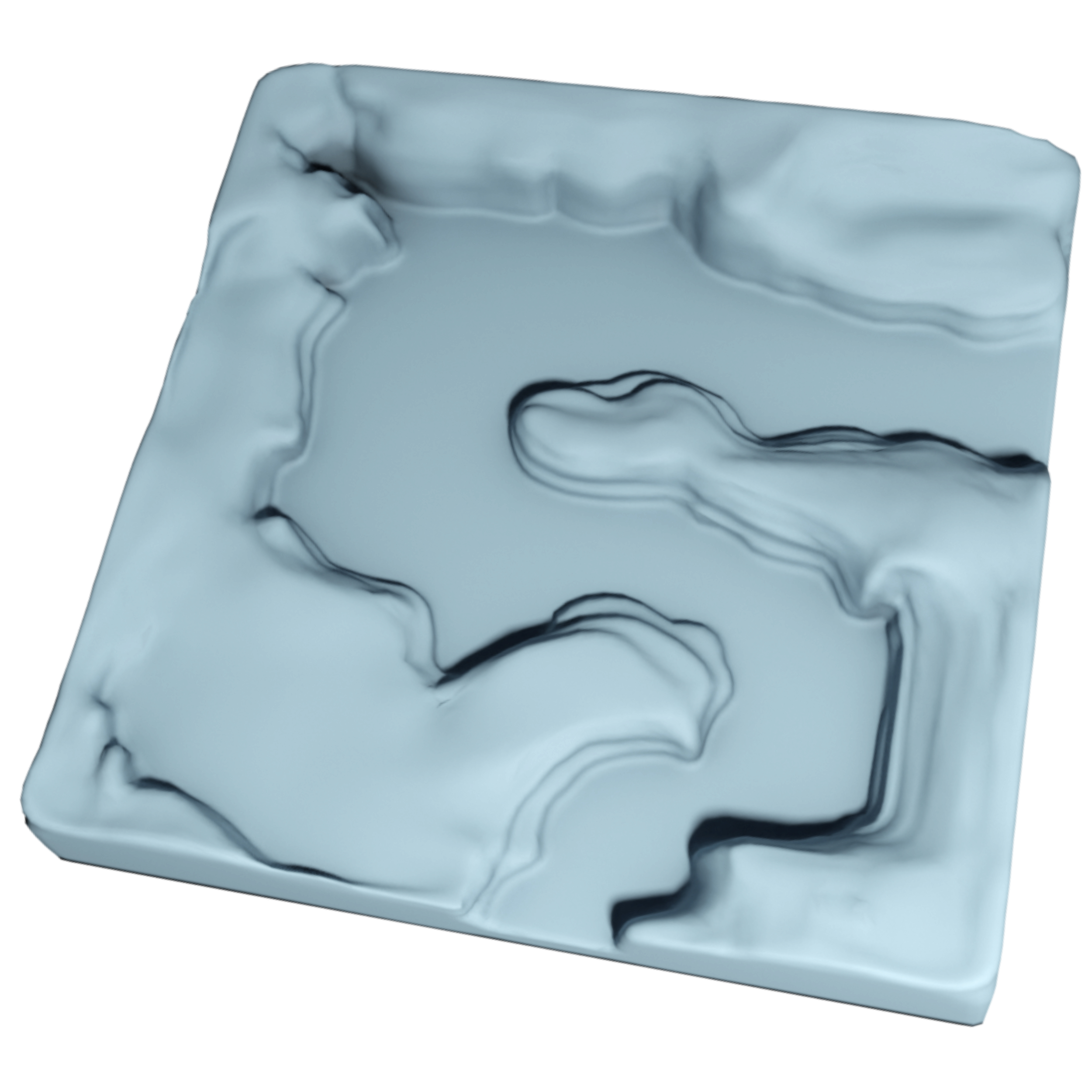}
         \caption{Rodin}
    \end{subfigure}
    
        \vspace*{-2mm}   
        \caption{\textbf{\ourmethod.} Given a 3D exemplar (left), we train a hierarchical diffusion model to create novel variations that preserve the geometric details and styles of the exemplar (center), whereas a large generative model such as Rodin~\cite{zhang2024clay}
        tends to lose the geometric details present in the input (right).}
    \label{fig:ours_0}
      \vspace*{-3mm}   
\end{figure}


\section{Additional results and renderings}

\begin{figure*}[!h]
\centering
    \begin{subfigure}{.16\linewidth}
        \centering
        \includegraphics[width=\linewidth]{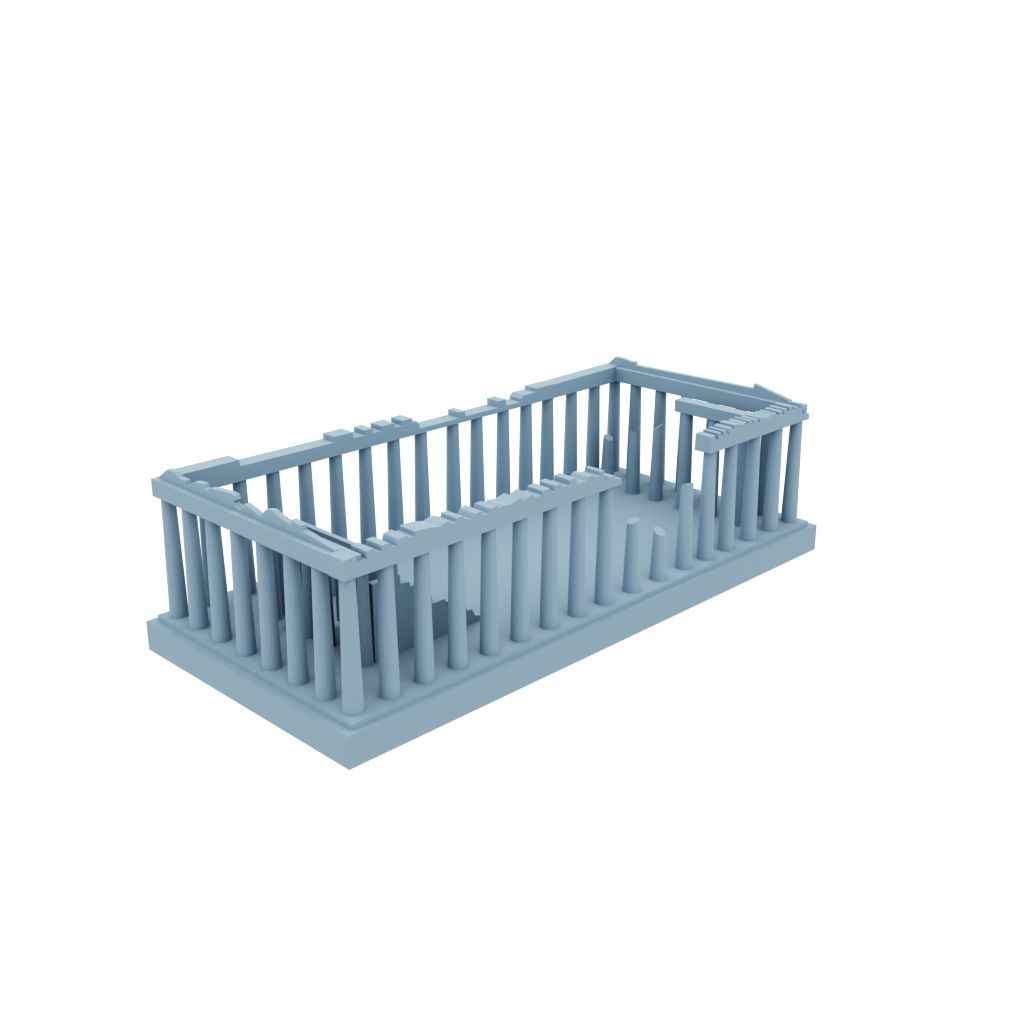}
    \end{subfigure}
    \begin{subfigure}{.16\linewidth}
        \centering
        \includegraphics[width=\linewidth]{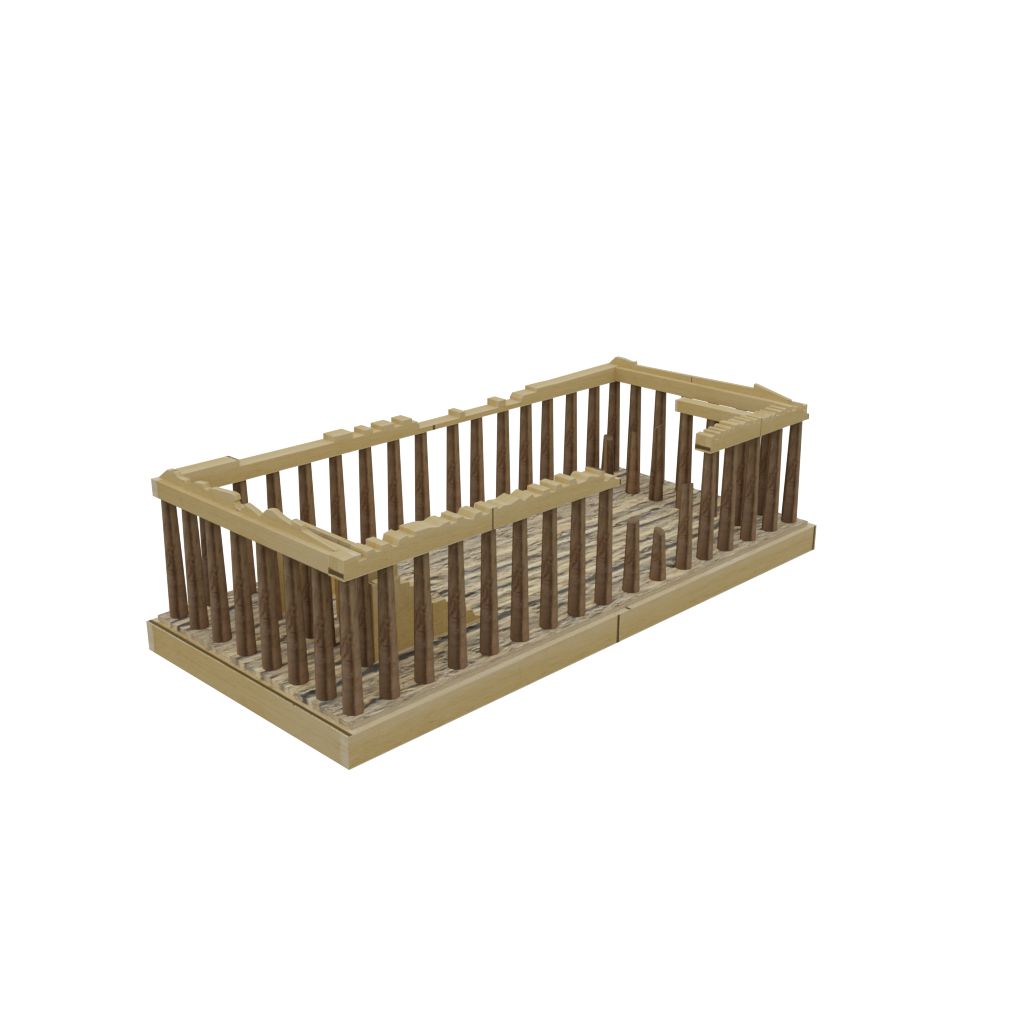}
    \end{subfigure}
    \unskip\ \vrule\ 
    \begin{subfigure}{.16\linewidth}
        \centering
        \includegraphics[width=\linewidth]{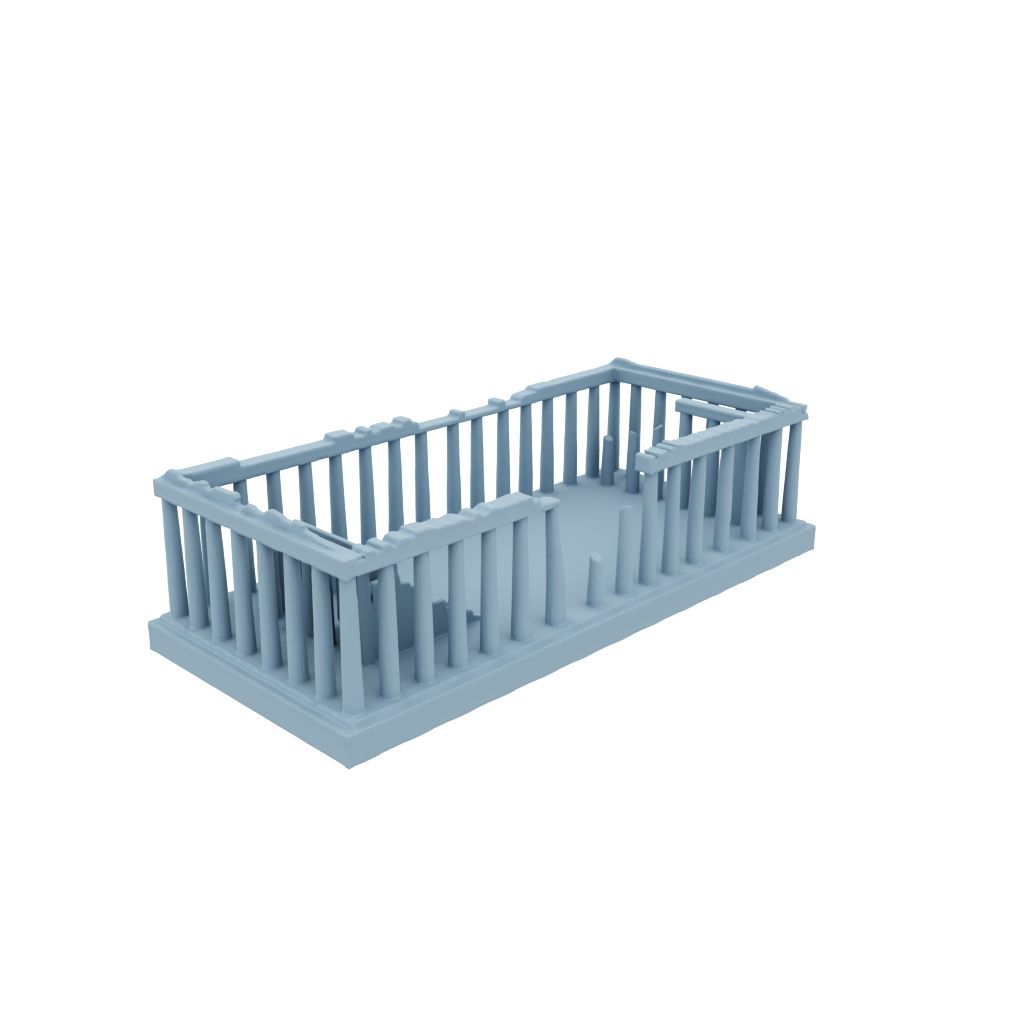}
    \end{subfigure}
    \begin{subfigure}{.16\linewidth}
        \centering
        \includegraphics[width=\linewidth]{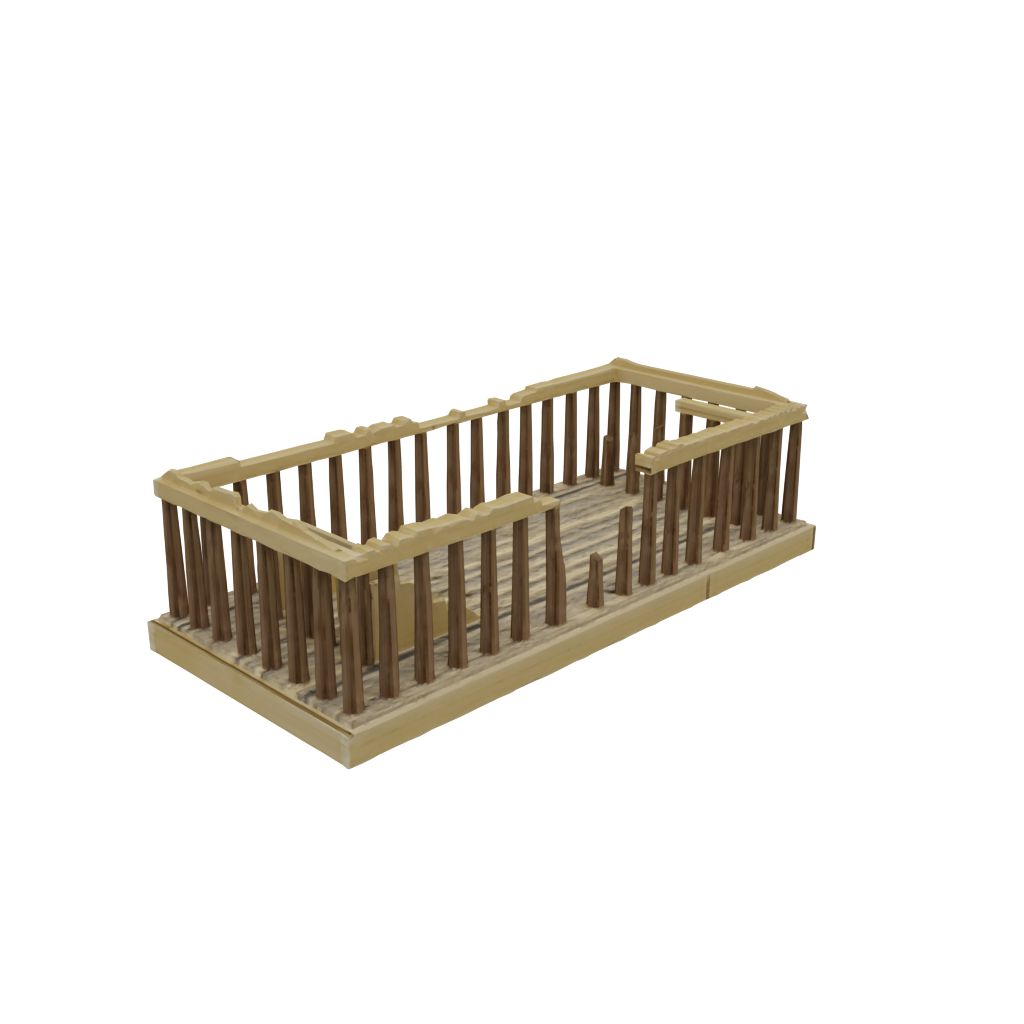}
    \end{subfigure}
    \begin{subfigure}{.16\linewidth}
        \centering
        \includegraphics[width=\linewidth]{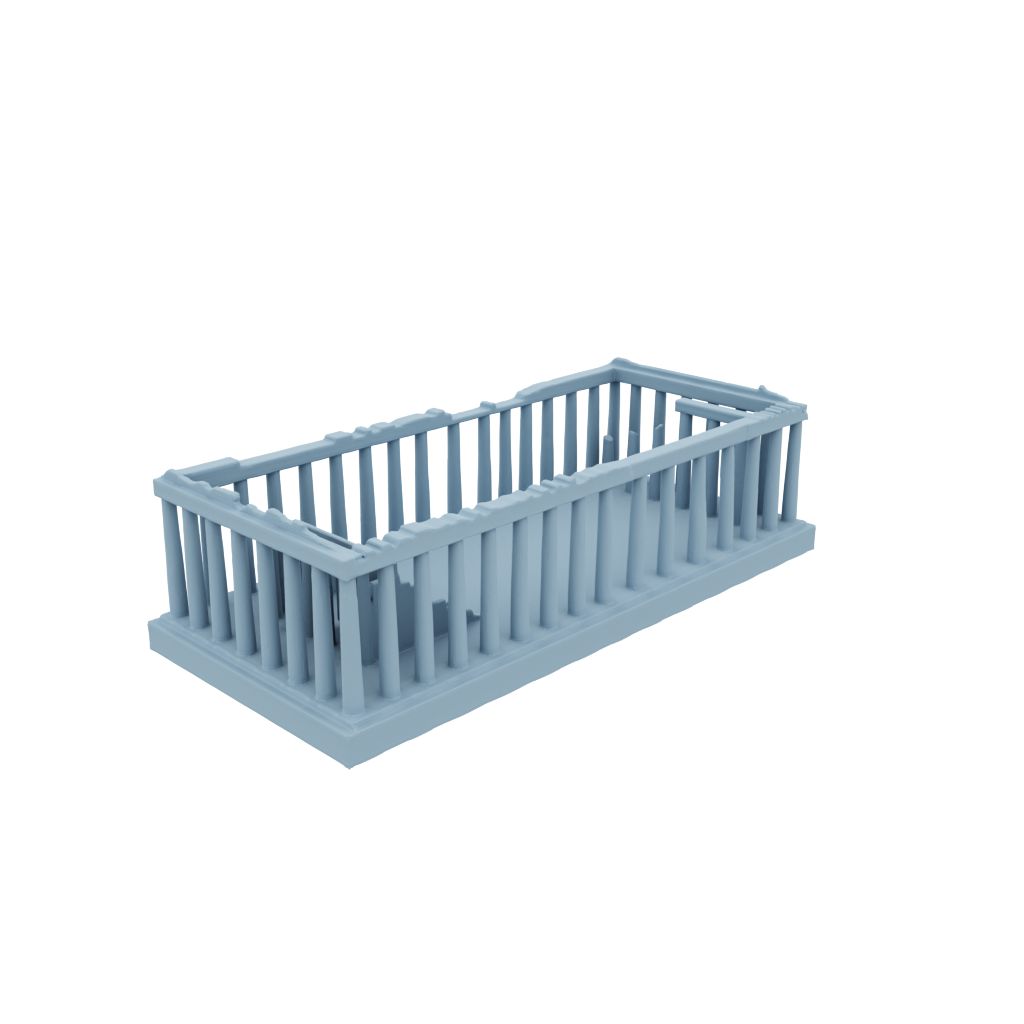}
    \end{subfigure}
    \begin{subfigure}{.16\linewidth}
        \centering
        \includegraphics[width=\linewidth]{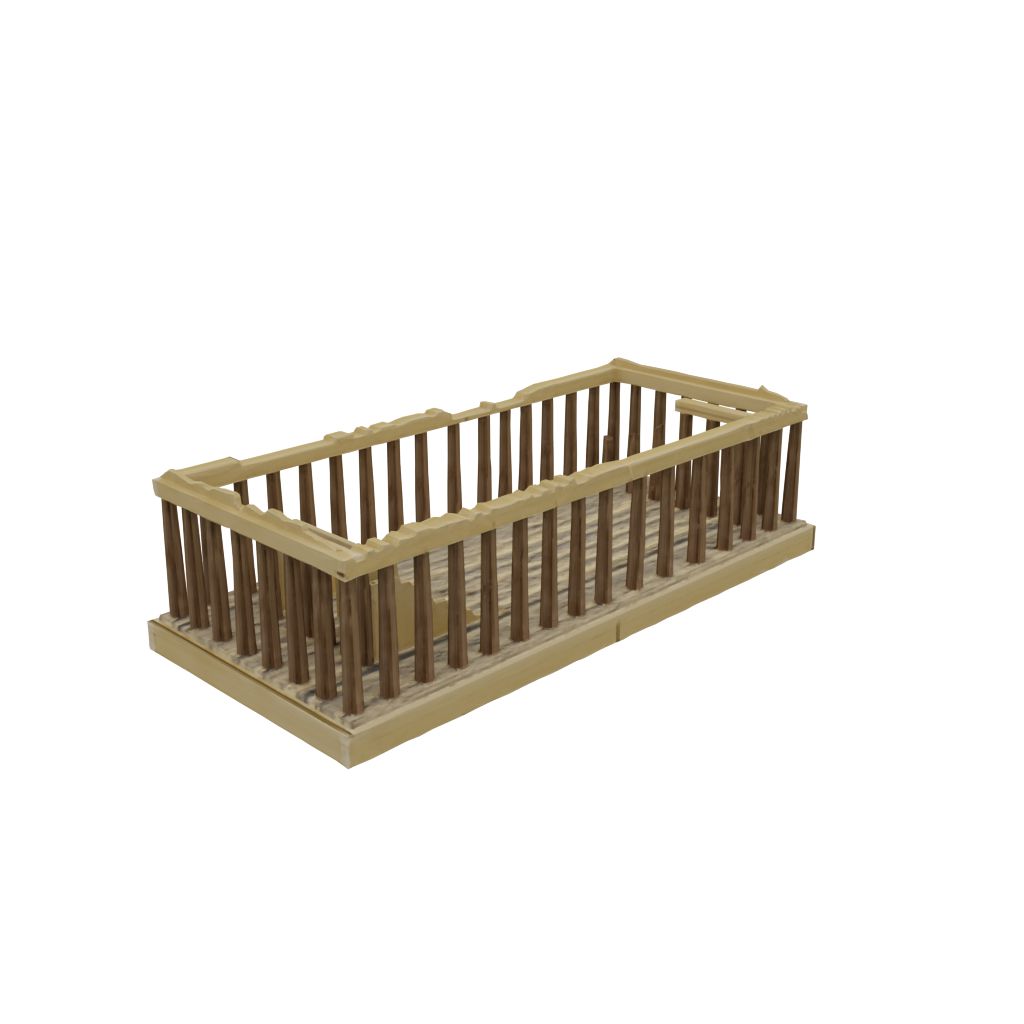}
    \end{subfigure}
    \vspace*{-5mm} 
    \begin{subfigure}{.16\linewidth}
        \centering
        \includegraphics[width=\linewidth]{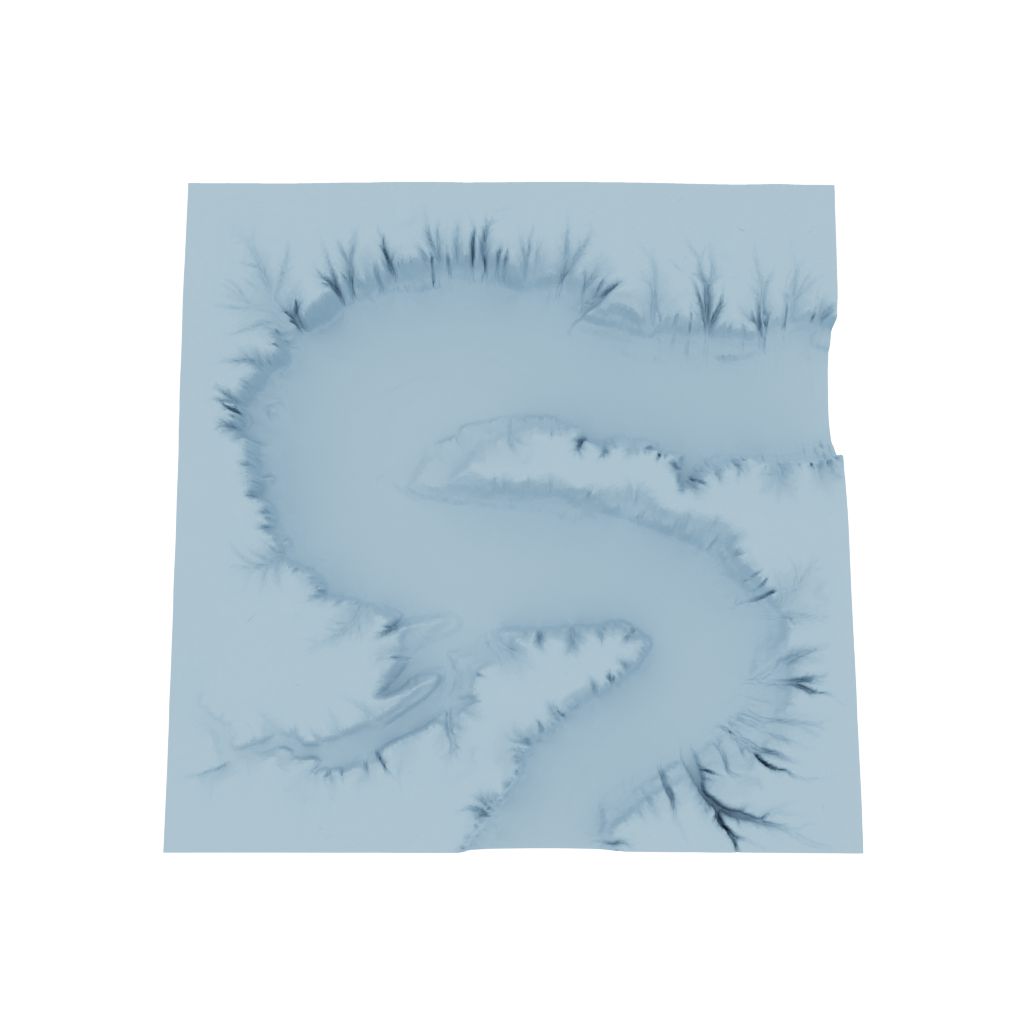}
    \end{subfigure}
    \begin{subfigure}{.16\linewidth}
        \centering
        \includegraphics[width=\linewidth]{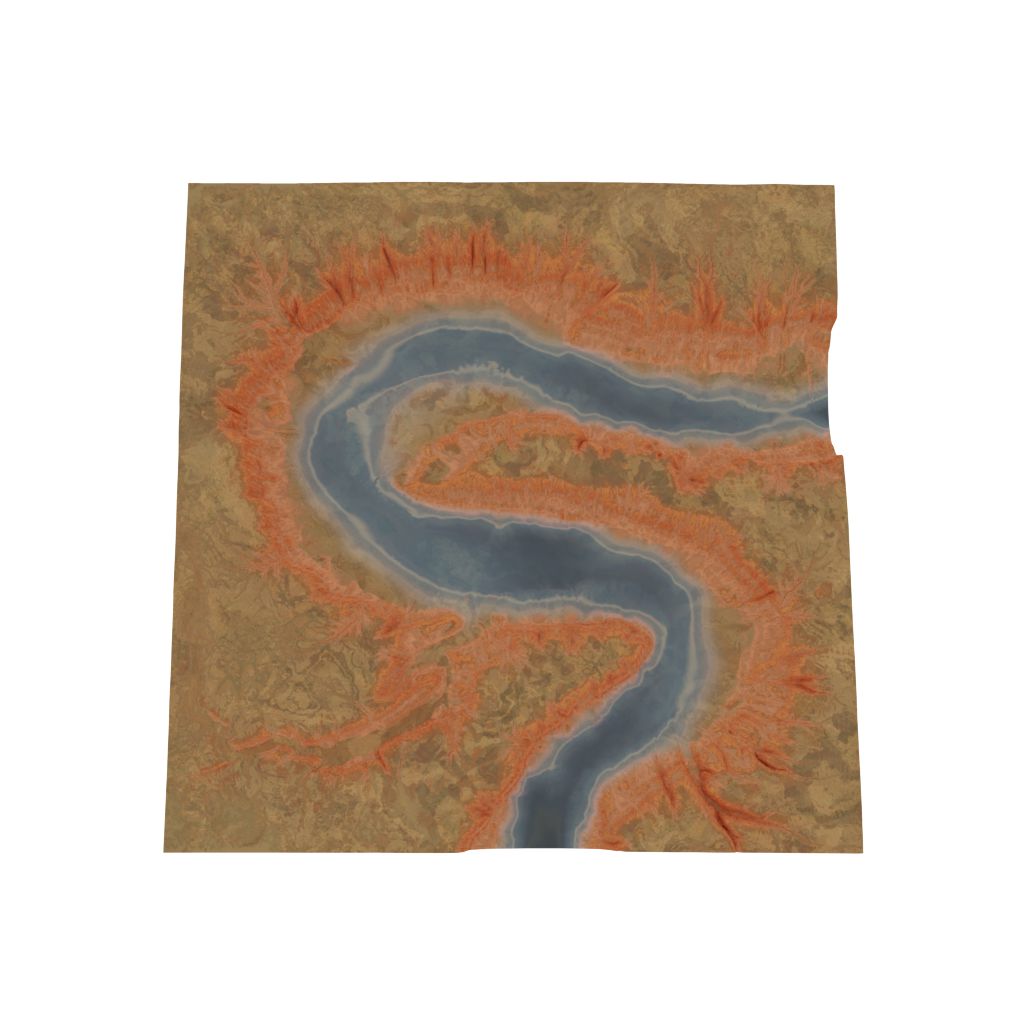}
    \end{subfigure}
    \unskip\ \vrule\ 
    \begin{subfigure}{.16\linewidth}
        \centering
        \includegraphics[width=\linewidth]{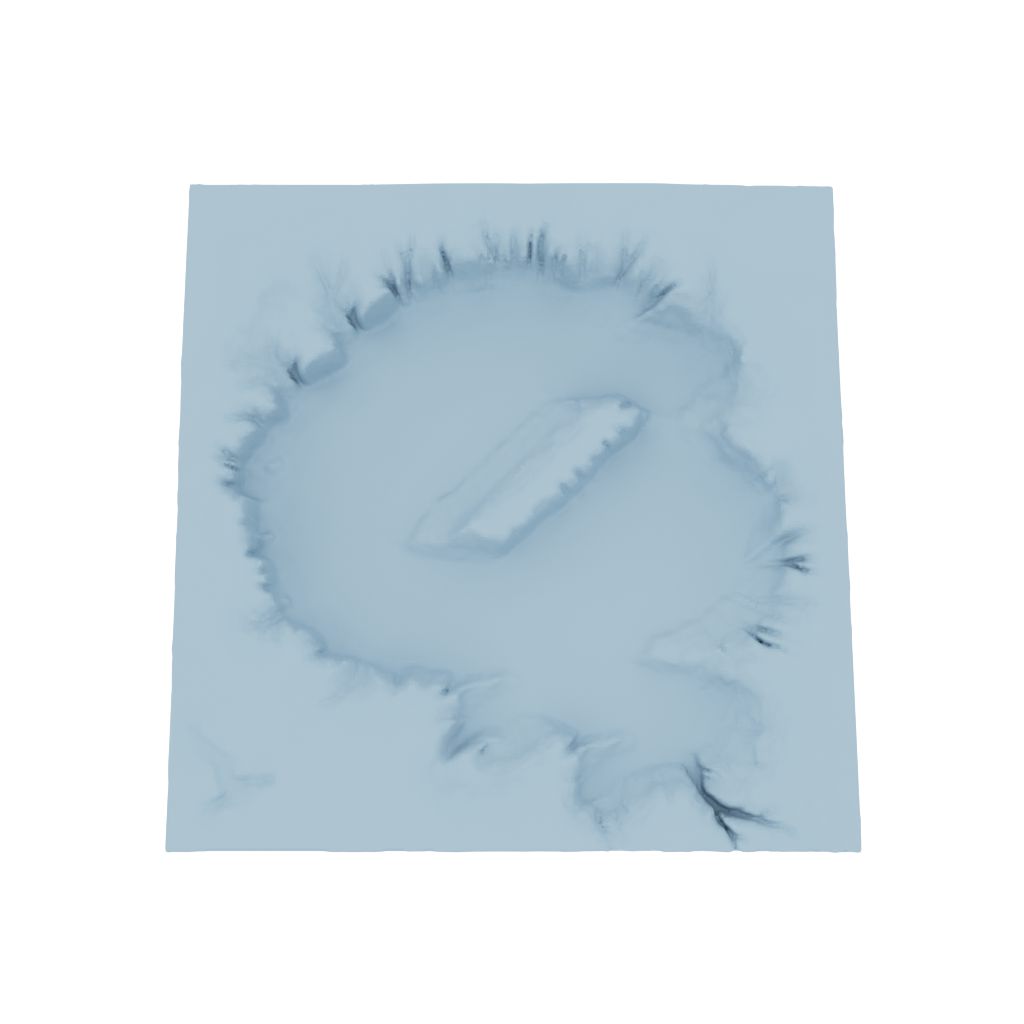}
    \end{subfigure}
    \begin{subfigure}{.16\linewidth}
        \centering
        \includegraphics[width=\linewidth]{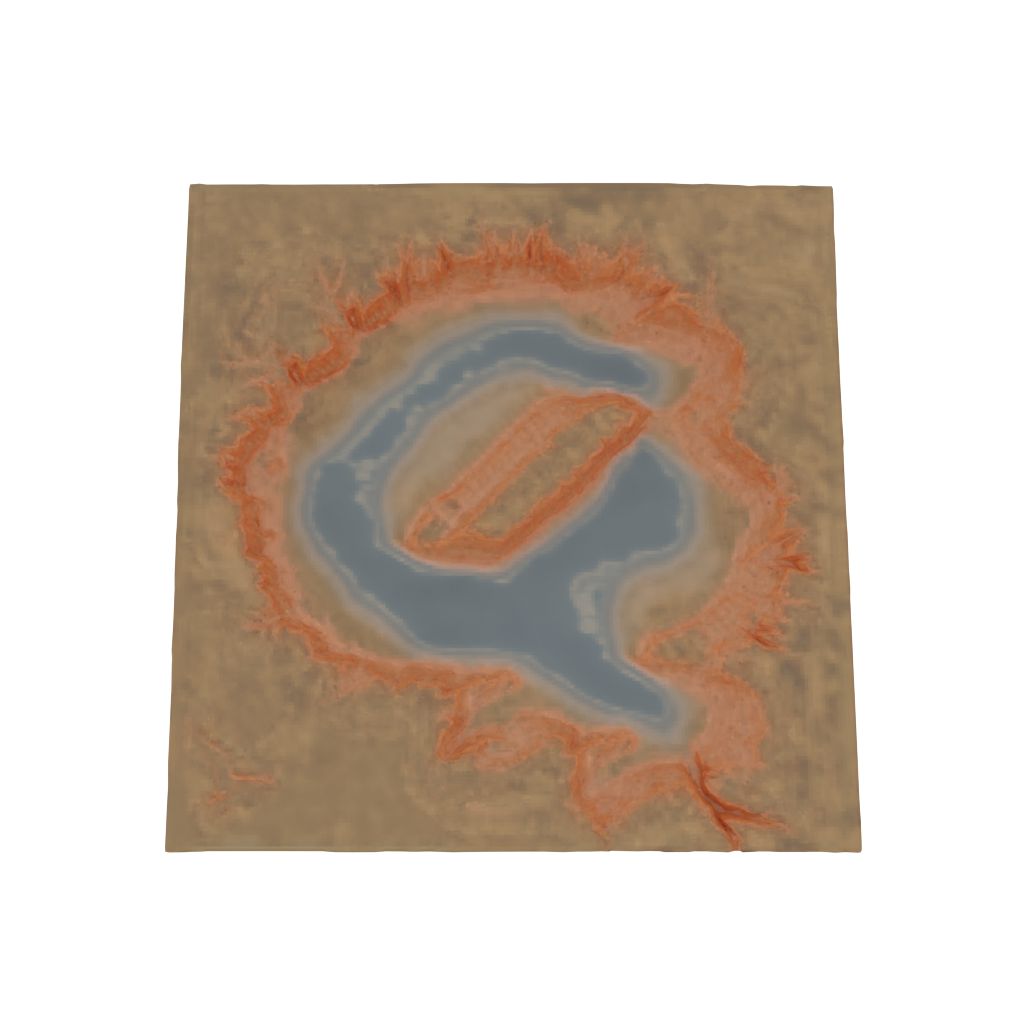}
    \end{subfigure}
    \begin{subfigure}{.16\linewidth}
        \centering
        \includegraphics[width=\linewidth]{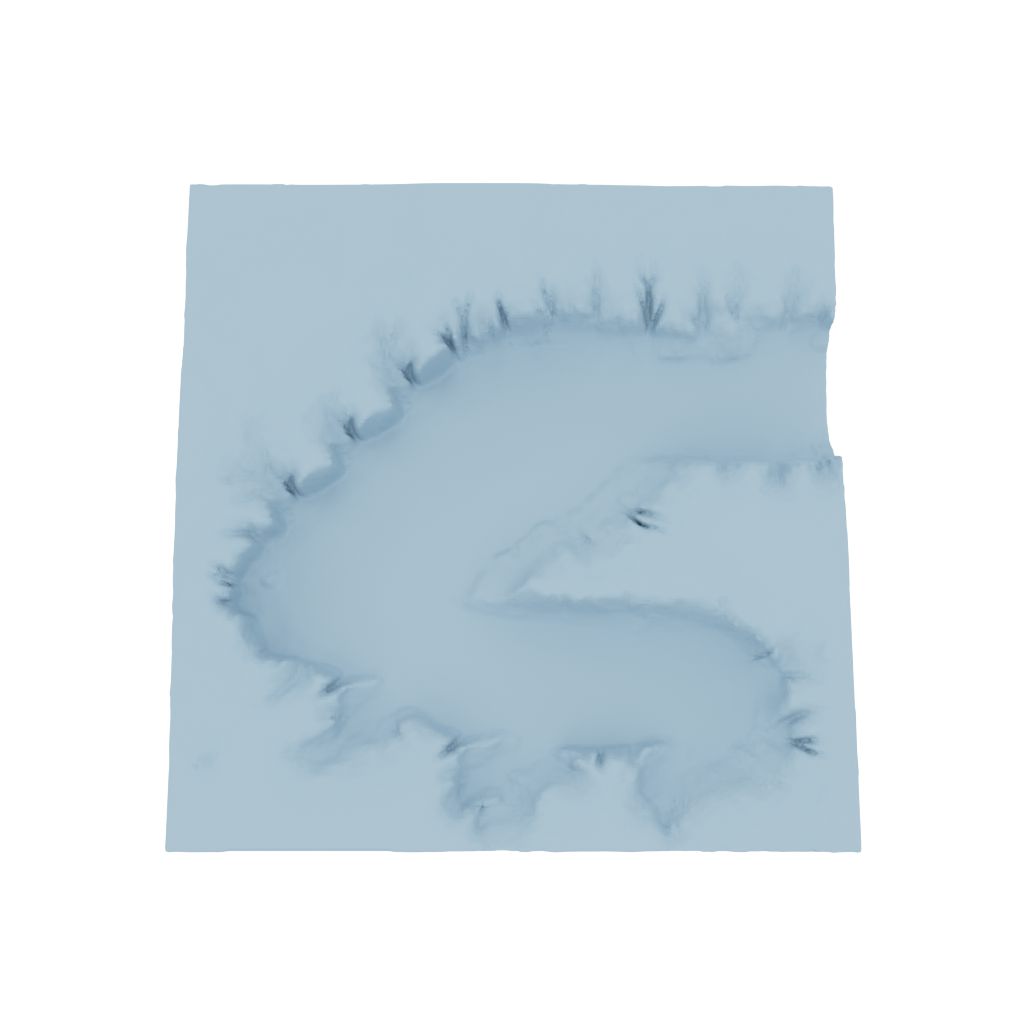}
    \end{subfigure}
    \begin{subfigure}{.16\linewidth}
        \centering
        \includegraphics[width=\linewidth]{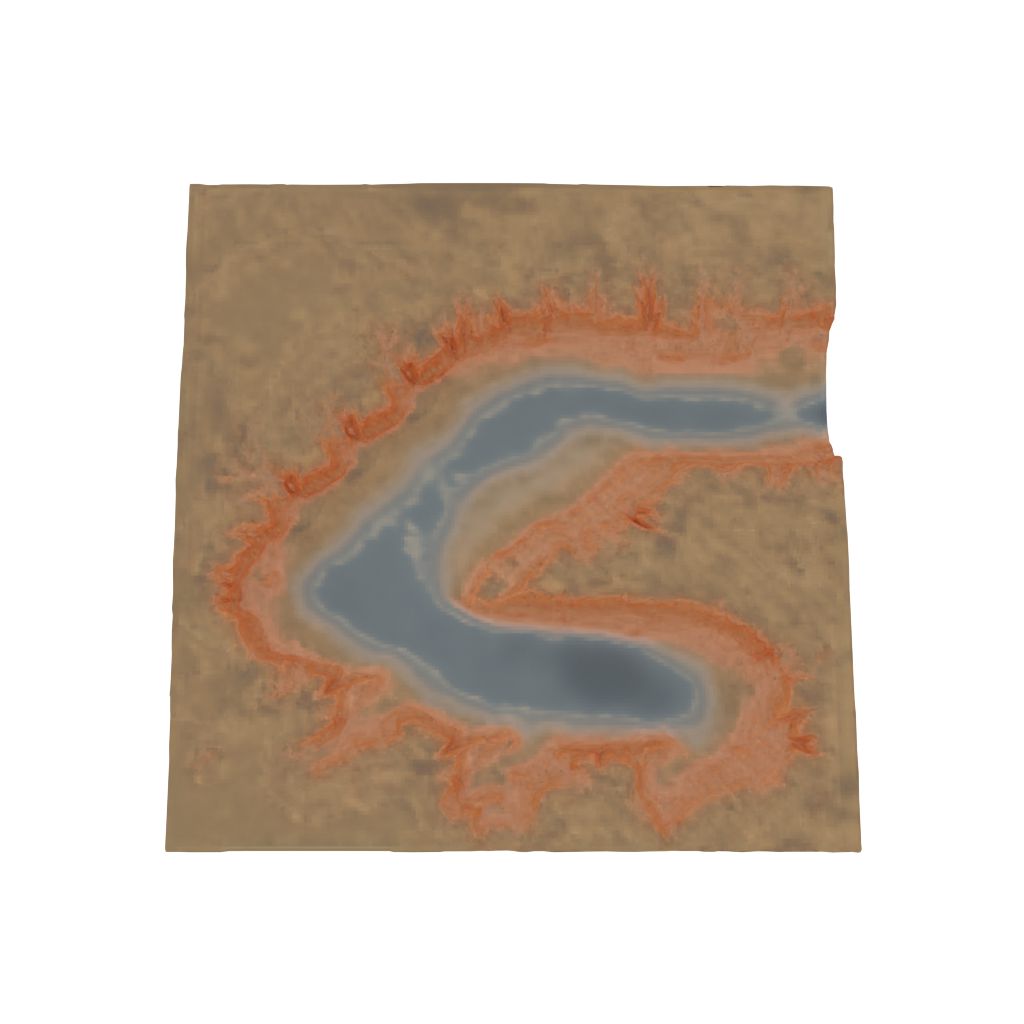}
    \end{subfigure}
    \vspace*{-5mm} 
    \begin{subfigure}{.16\linewidth}
        \centering
        \includegraphics[width=\linewidth]{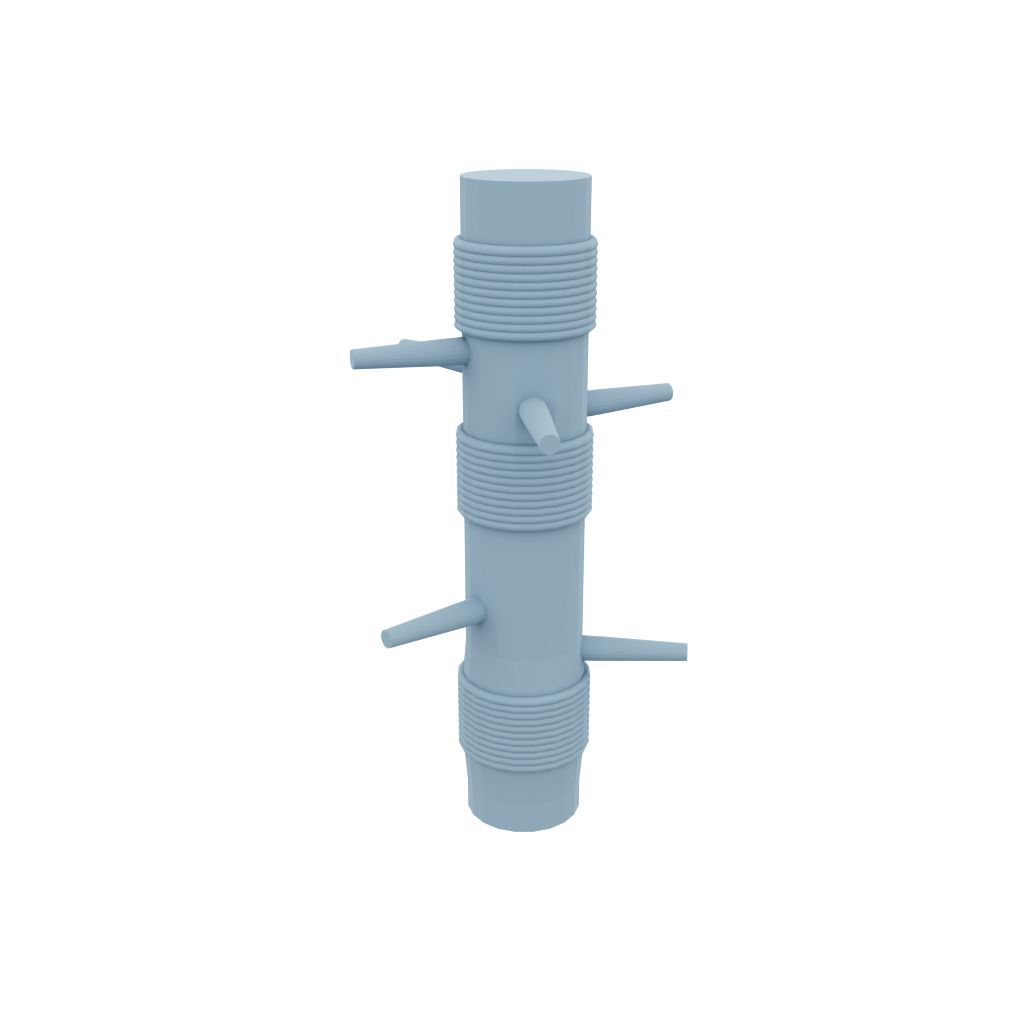}
    \end{subfigure}
    \begin{subfigure}{.16\linewidth}
        \centering
        \includegraphics[width=\linewidth]{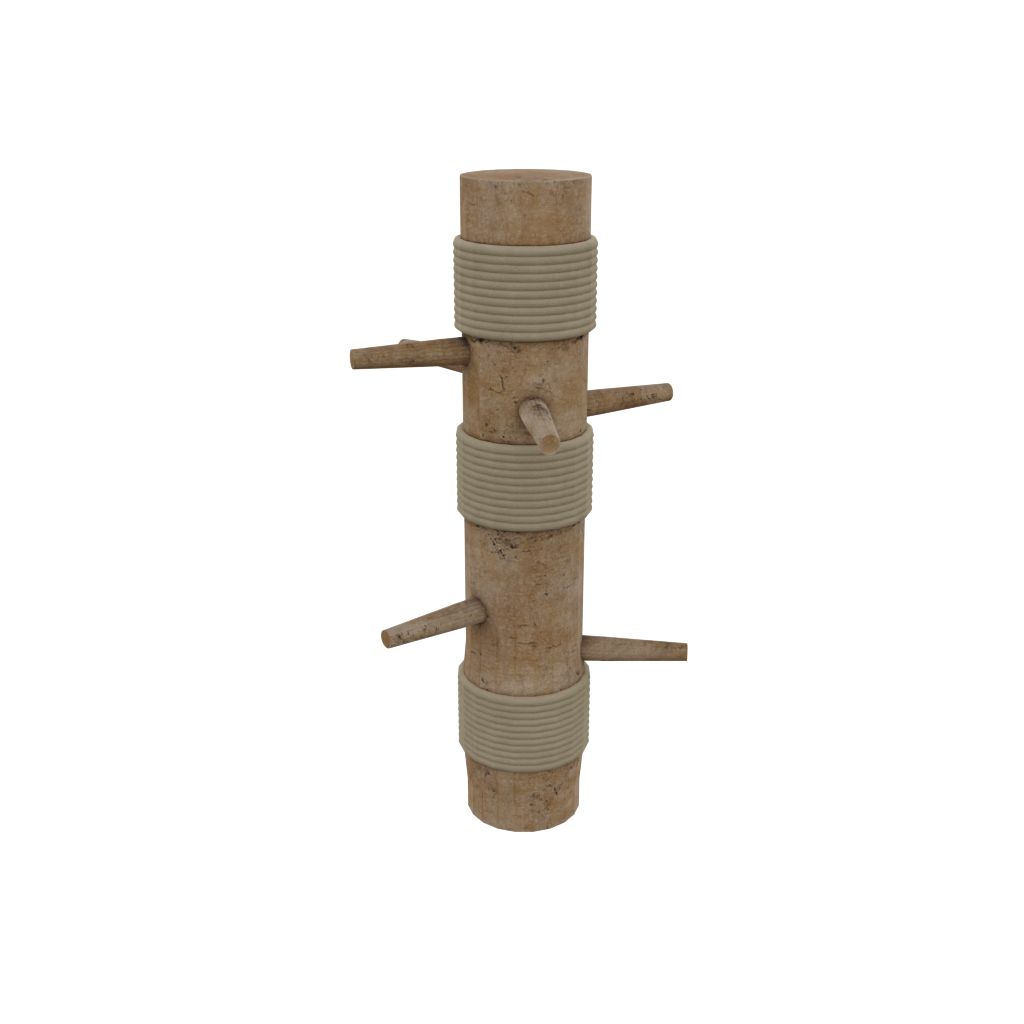}
    \end{subfigure}
    \unskip\ \vrule\ 
    \begin{subfigure}{.16\linewidth}
        \centering
        \includegraphics[width=\linewidth]{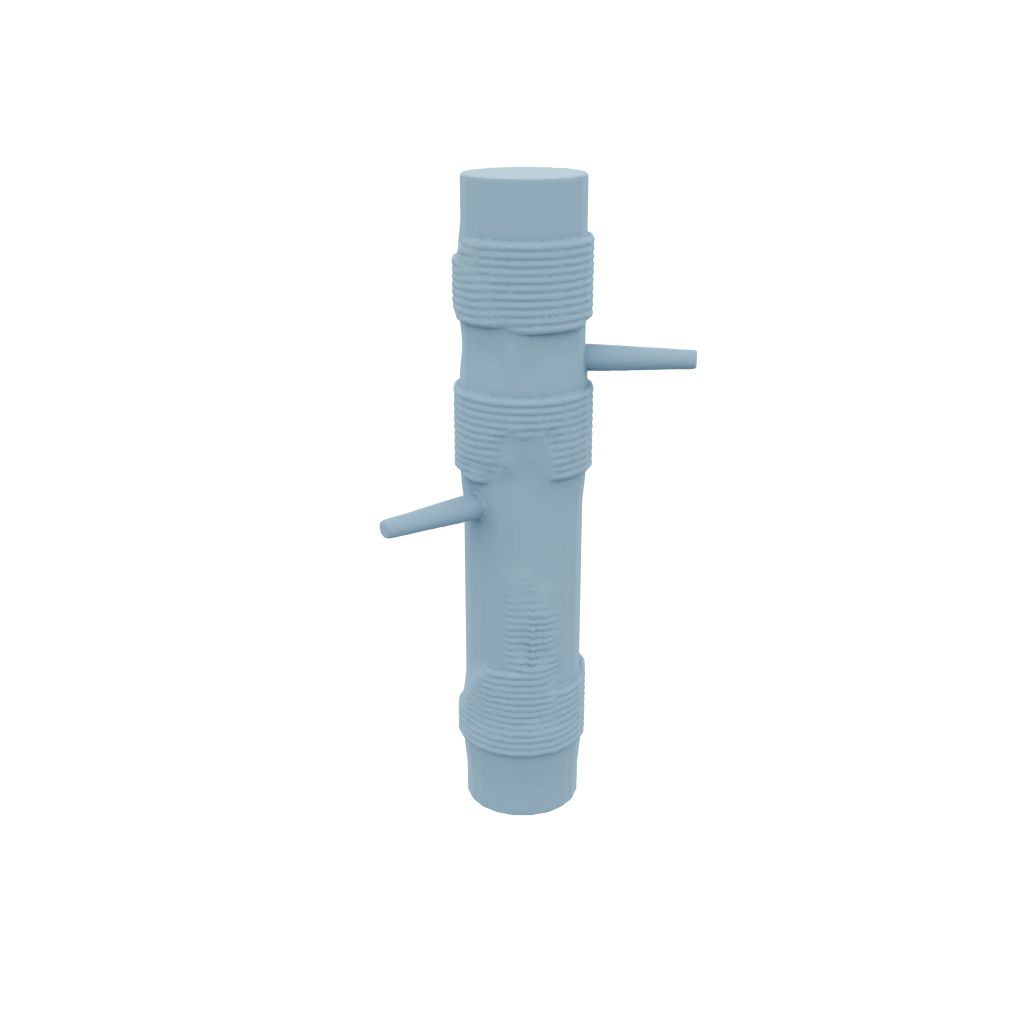}
    \end{subfigure}
    \begin{subfigure}{.16\linewidth}
        \centering
        \includegraphics[width=\linewidth]{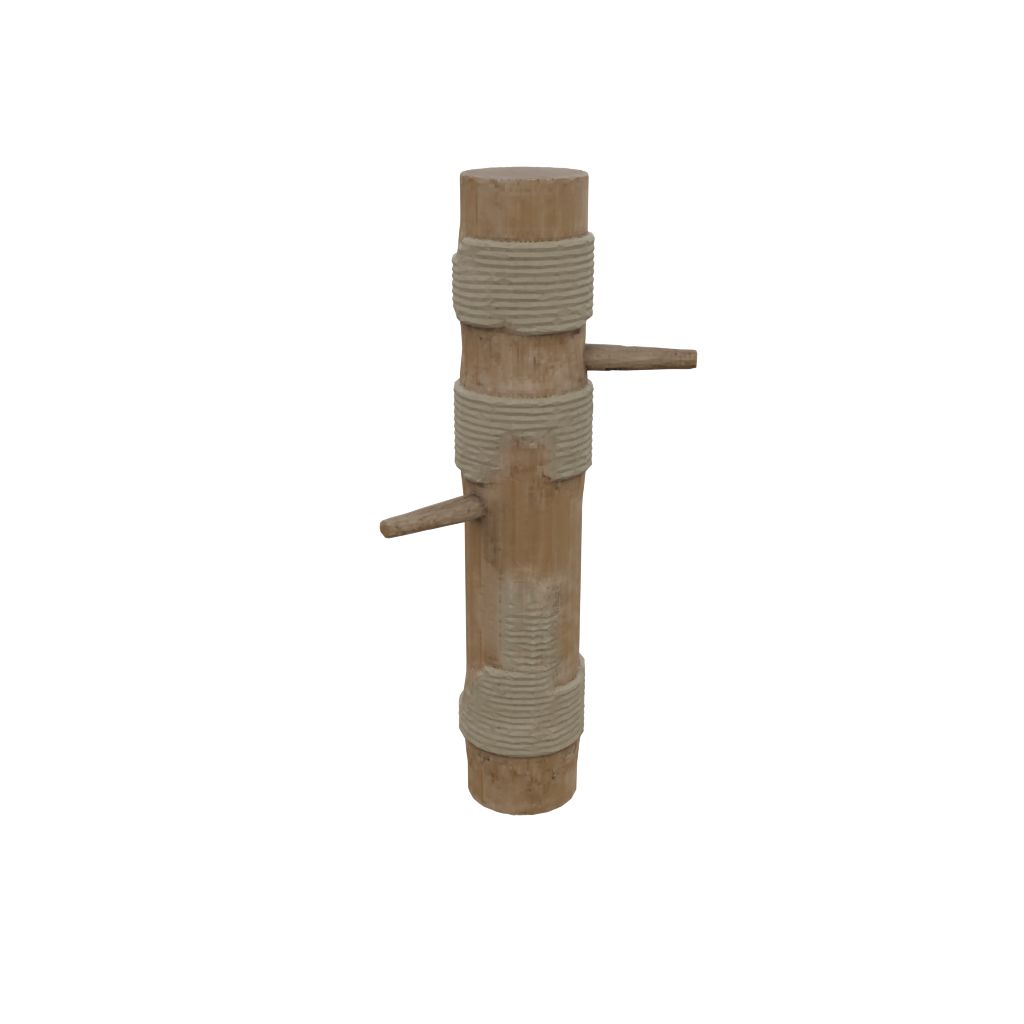}
    \end{subfigure}
    \begin{subfigure}{.16\linewidth}
        \centering
        \includegraphics[width=\linewidth]{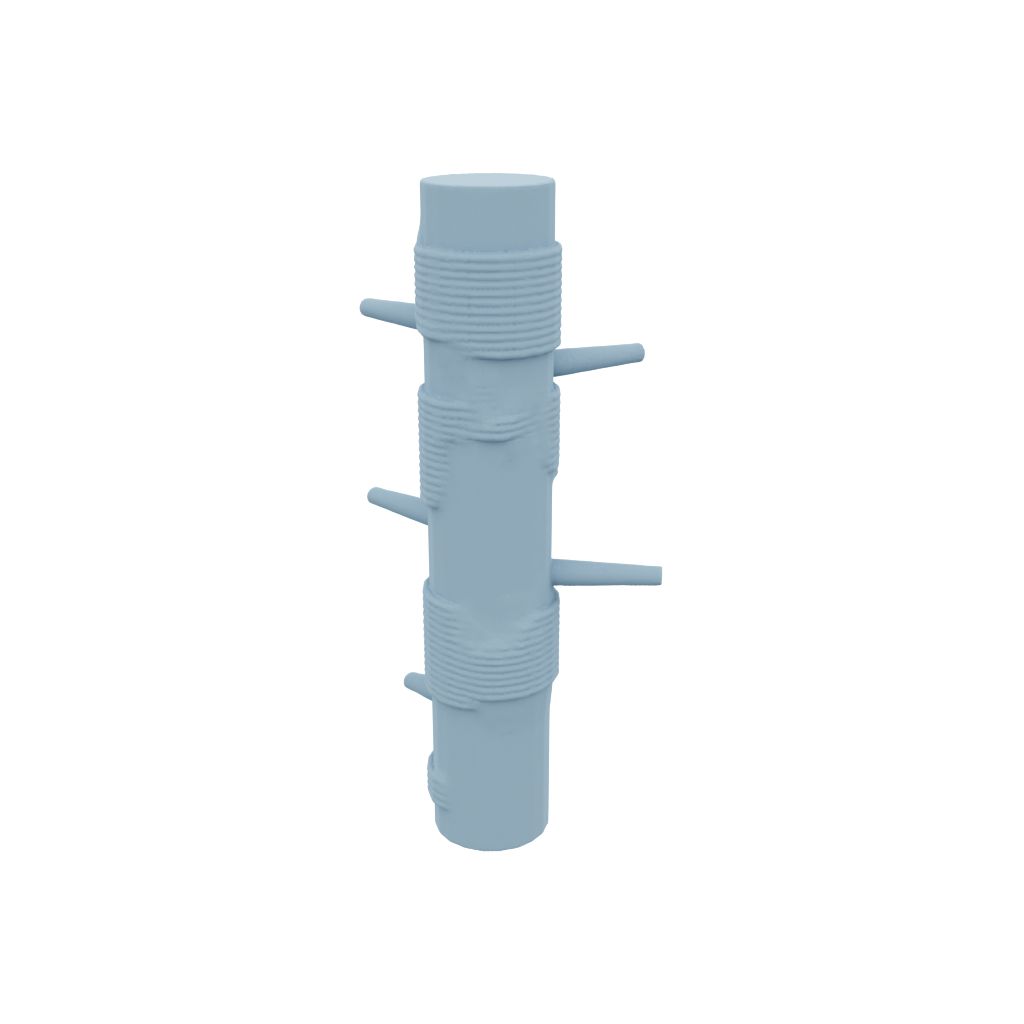}
    \end{subfigure}
    \begin{subfigure}{.16\linewidth}
        \centering
        \includegraphics[width=\linewidth]{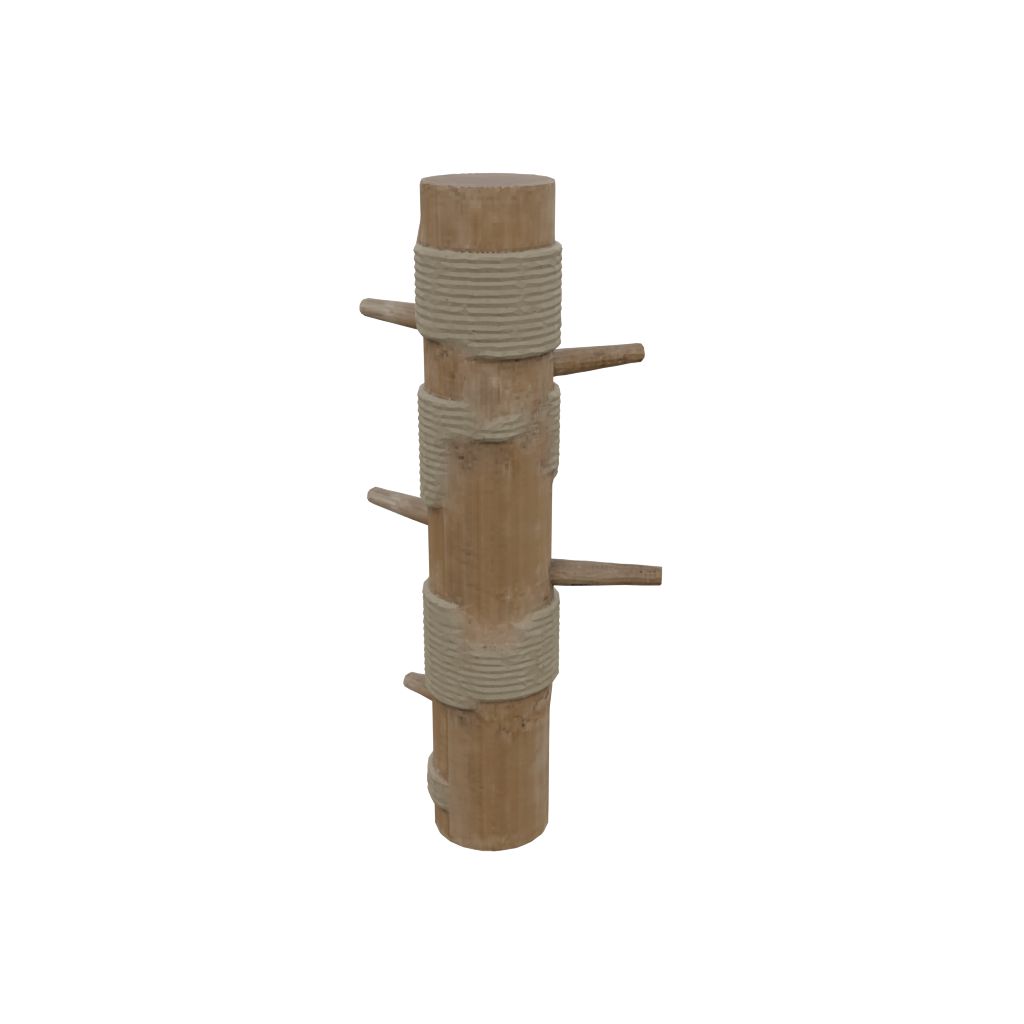}
    \end{subfigure}
    \vspace*{-5mm} 
    \begin{subfigure}{.16\linewidth}
        \centering
        \includegraphics[width=\linewidth]{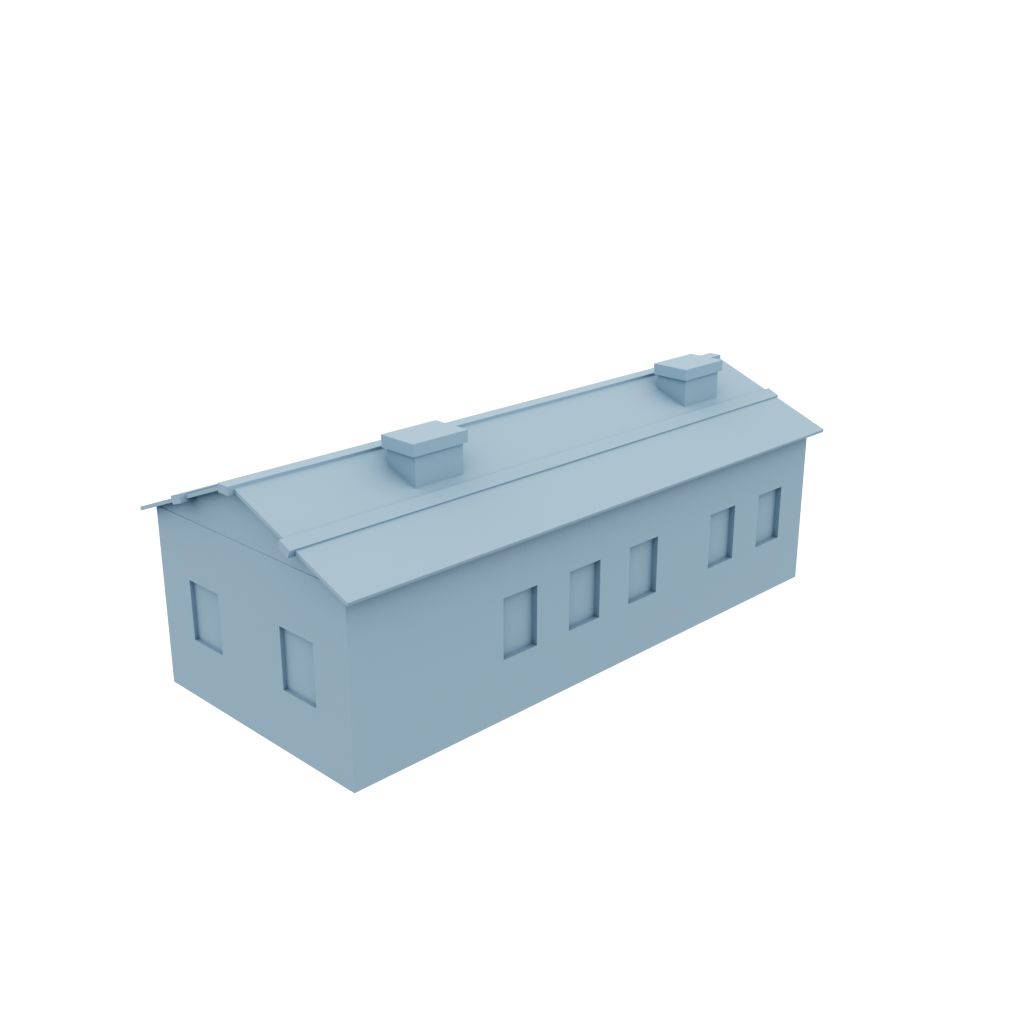}
    \end{subfigure}
    \begin{subfigure}{.16\linewidth}
        \centering
        \includegraphics[width=\linewidth]{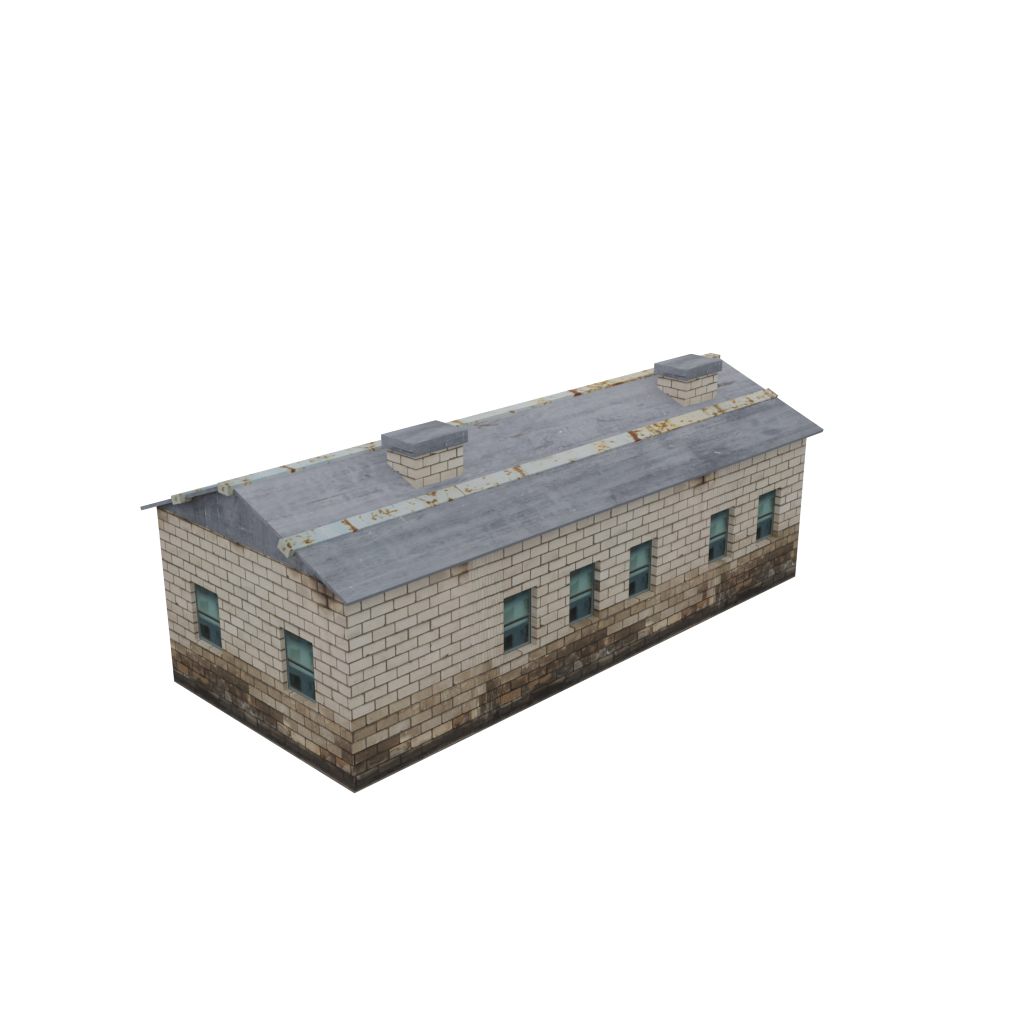}
    \end{subfigure}
    \unskip\ \vrule\ 
    \begin{subfigure}{.16\linewidth}
        \centering
        \includegraphics[width=\linewidth]{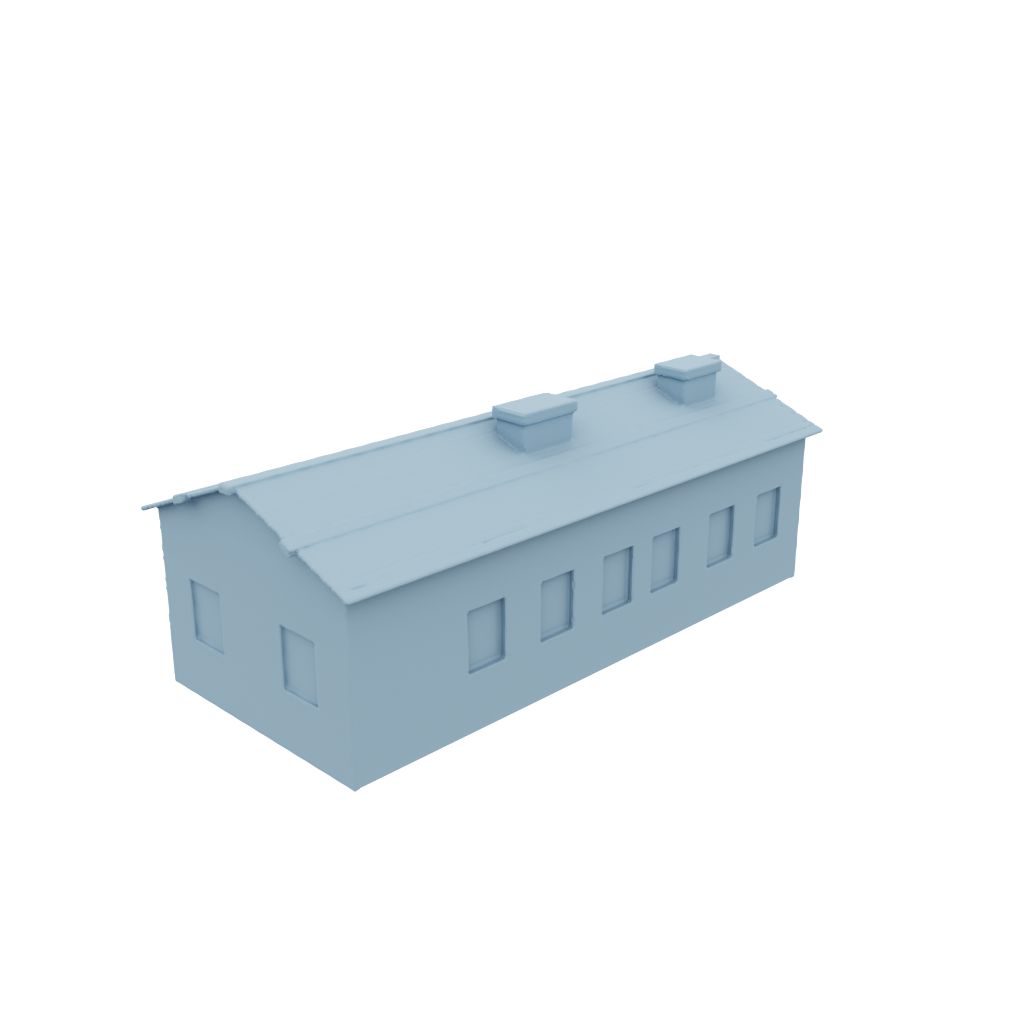}
    \end{subfigure}
    \begin{subfigure}{.16\linewidth}
        \centering
        \includegraphics[width=\linewidth]{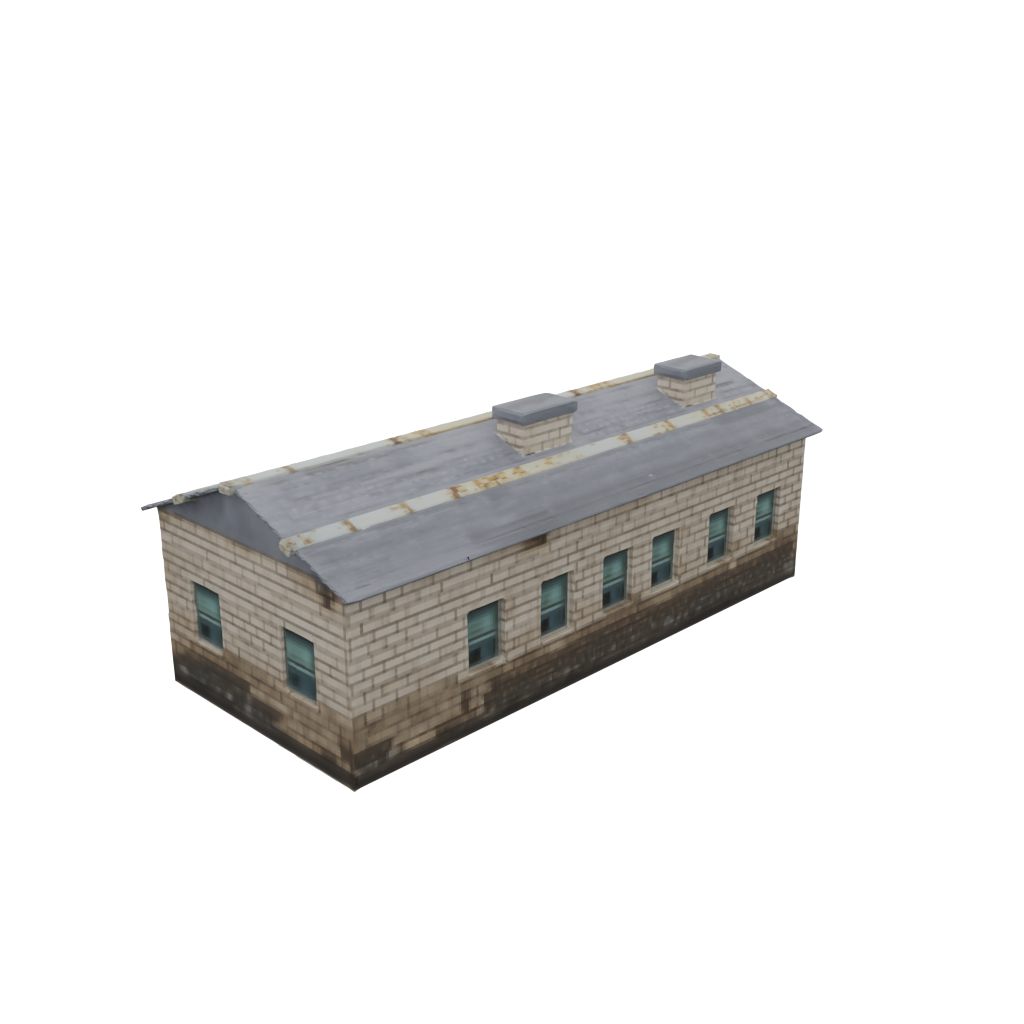}
    \end{subfigure}
    \begin{subfigure}{.16\linewidth}
        \centering
        \includegraphics[width=\linewidth]{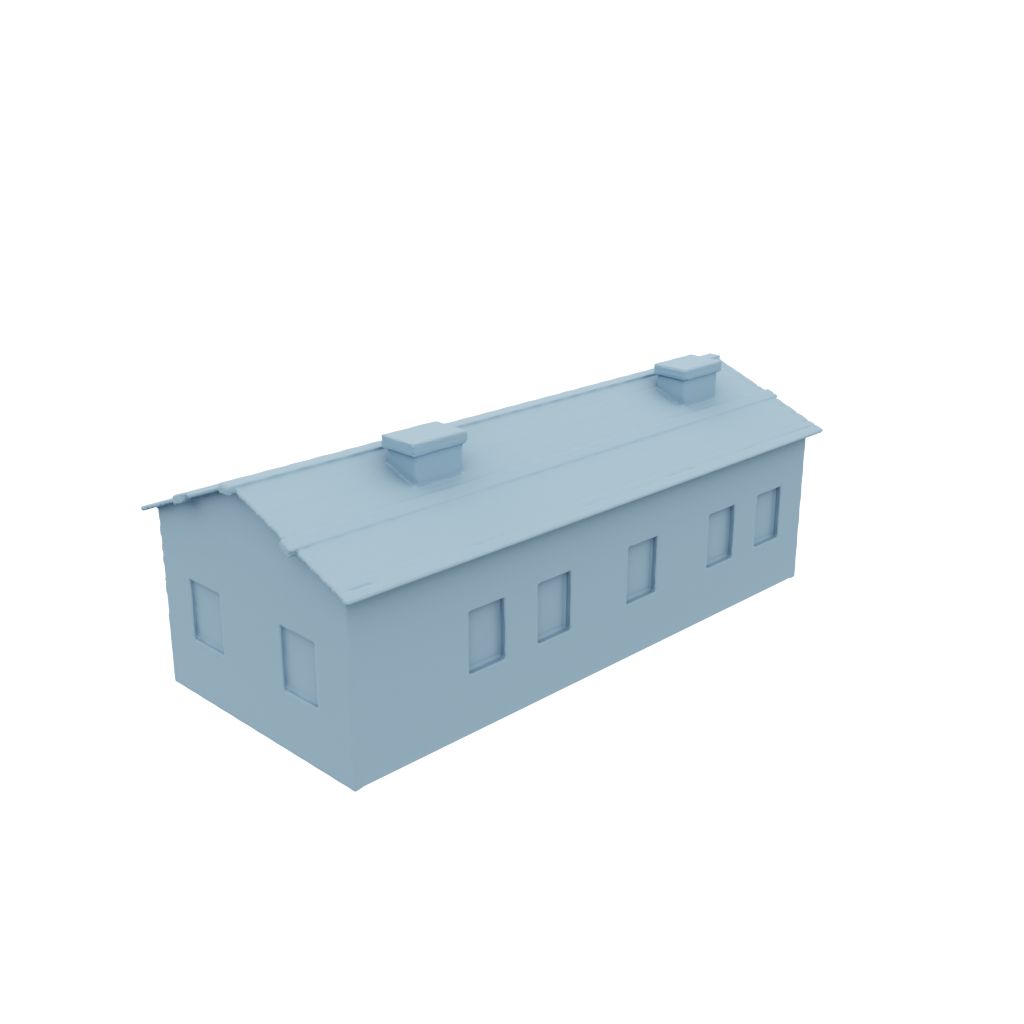}
    \end{subfigure}
    \begin{subfigure}{.16\linewidth}
        \centering
        \includegraphics[width=\linewidth]{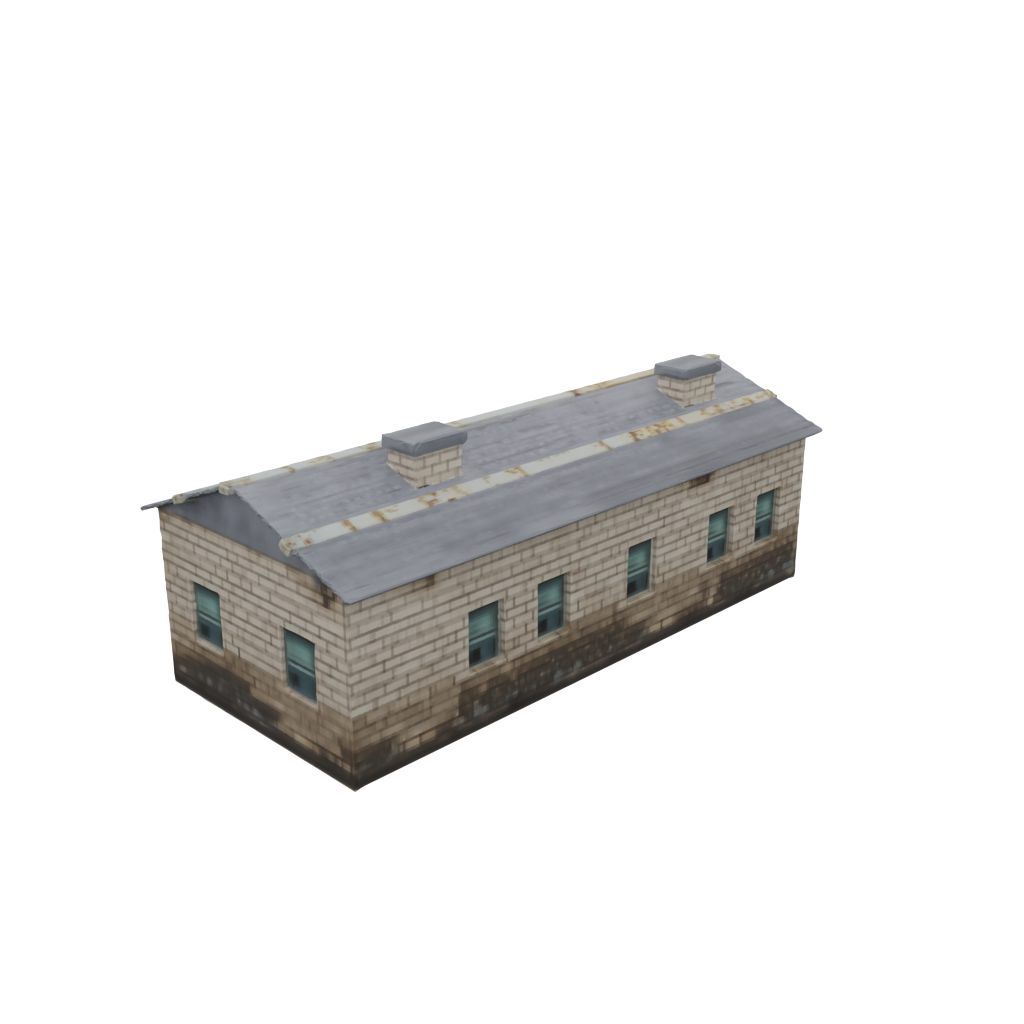}
    \end{subfigure}
    \begin{subfigure}{.16\linewidth}
        \centering
        \includegraphics[width=\linewidth]{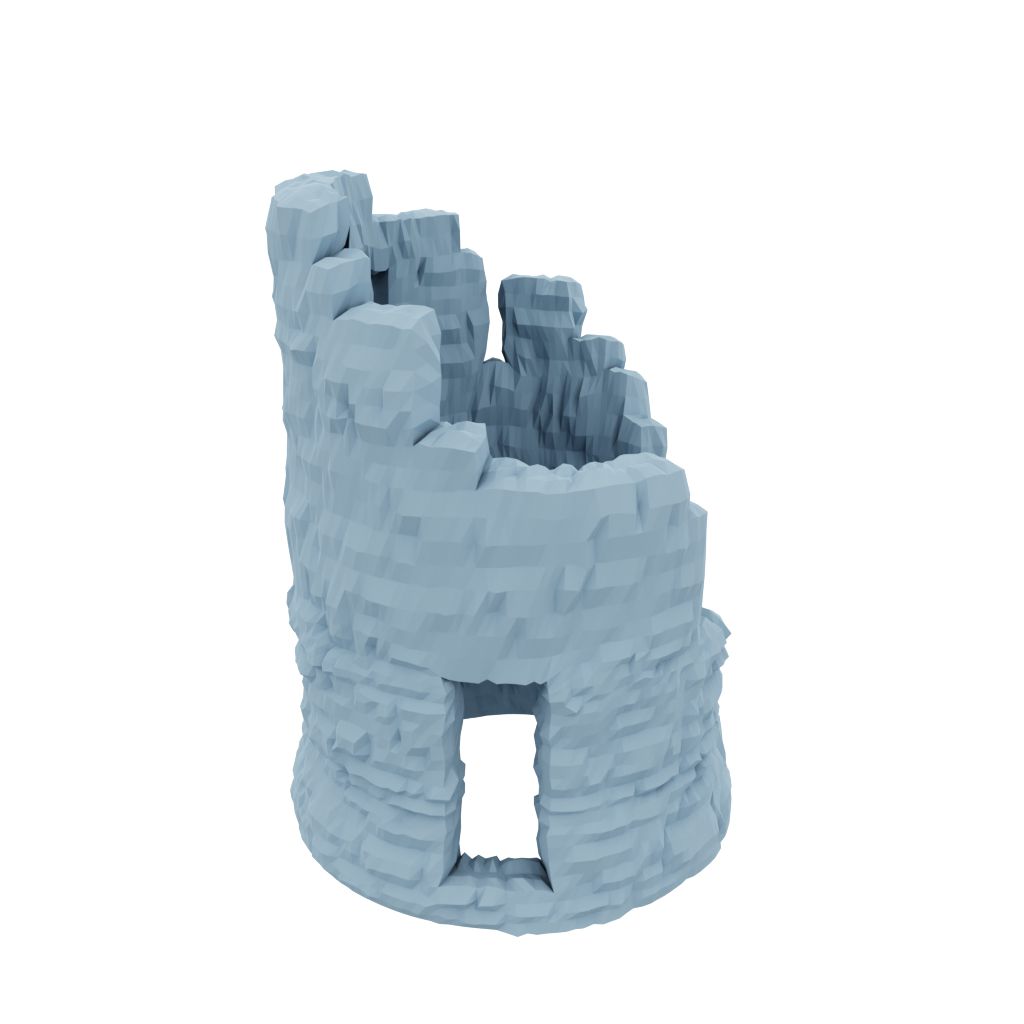}
    \end{subfigure}
    \begin{subfigure}{.16\linewidth}
        \centering
        \includegraphics[width=\linewidth]{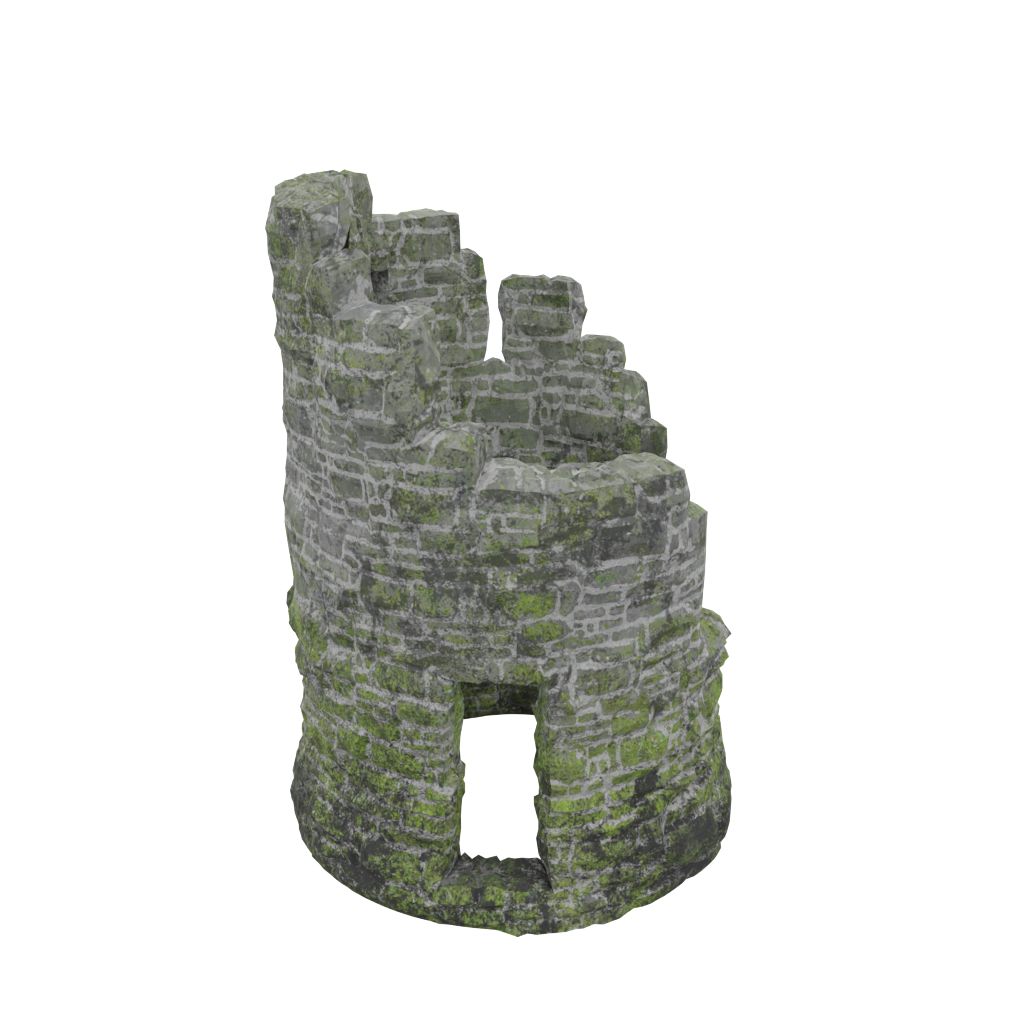}
    \end{subfigure}
    \unskip\ \vrule\ 
    \begin{subfigure}{.16\linewidth}
        \centering
        \includegraphics[width=\linewidth]{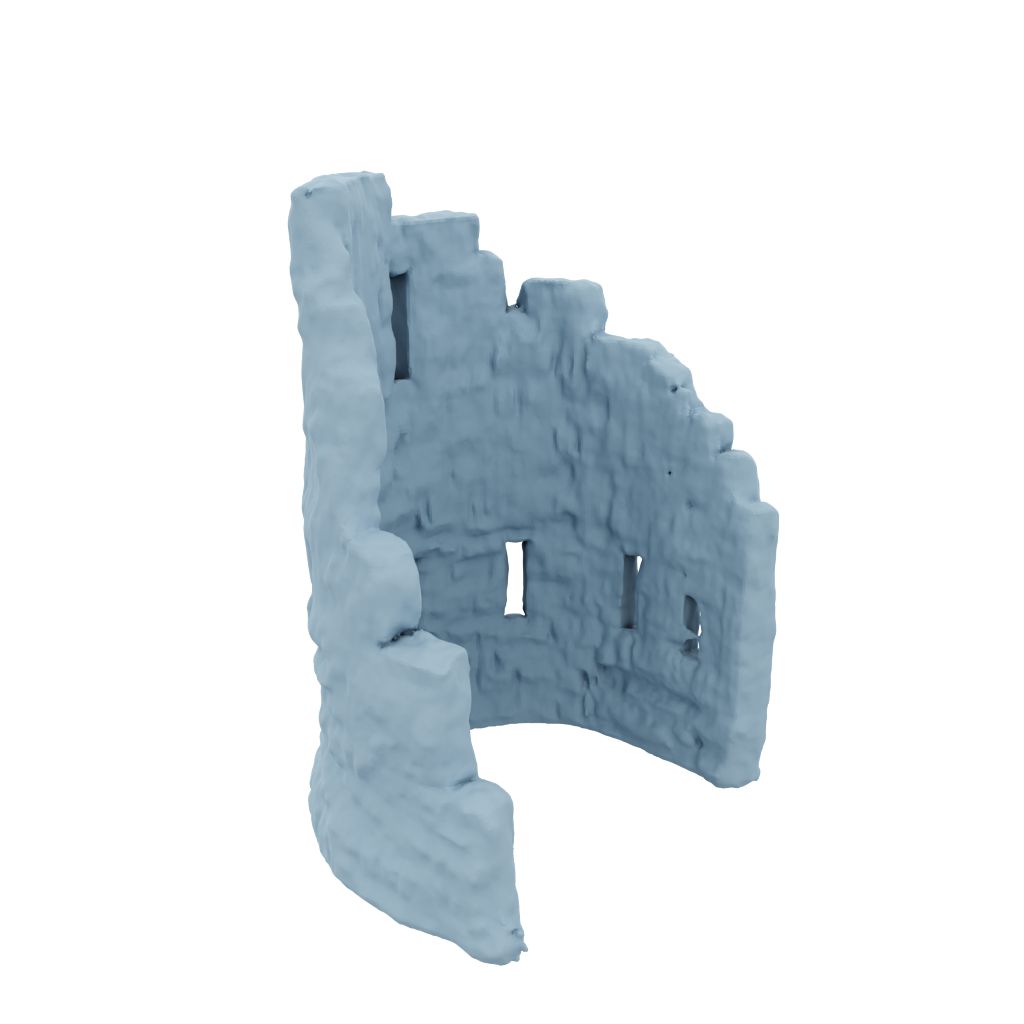}
    \end{subfigure}
    \begin{subfigure}{.16\linewidth}
        \centering
        \includegraphics[width=\linewidth]{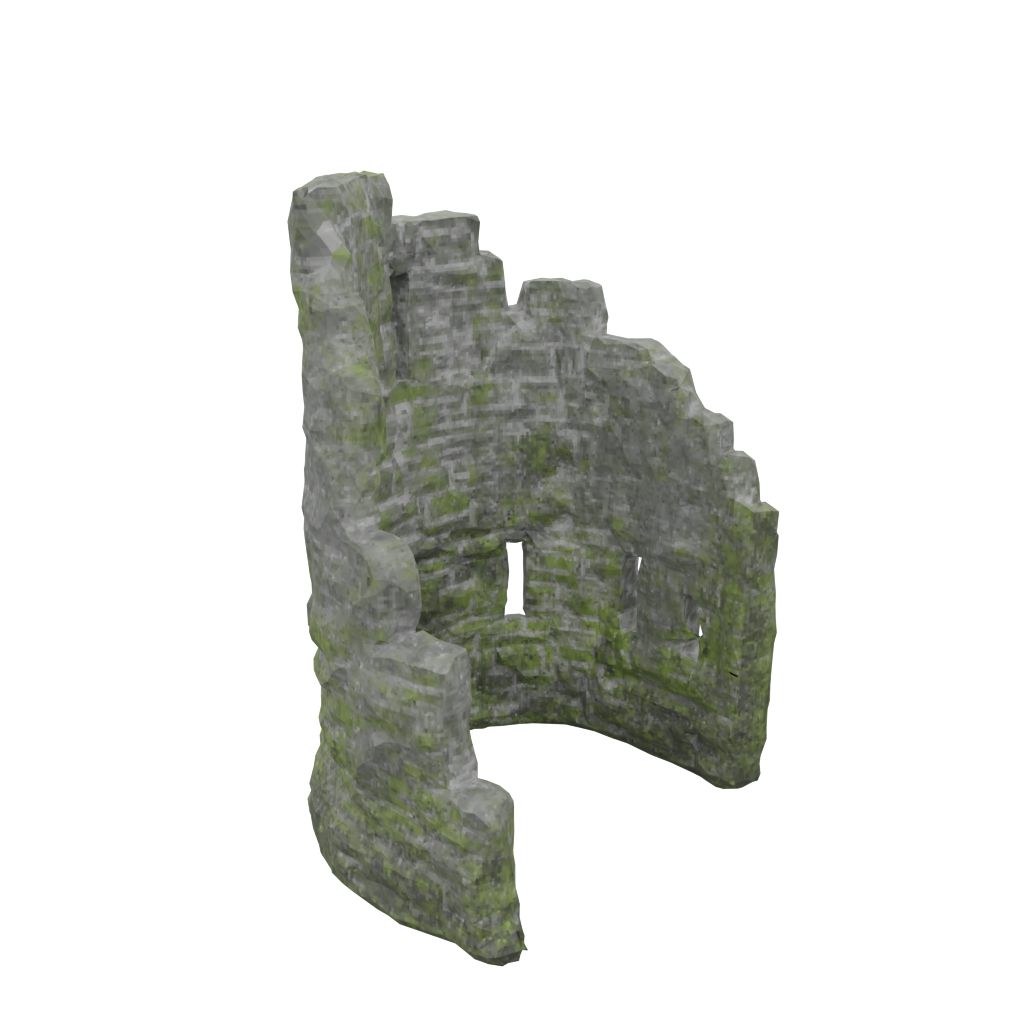}
    \end{subfigure}
    \begin{subfigure}{.16\linewidth}
        \centering
        \includegraphics[width=\linewidth]{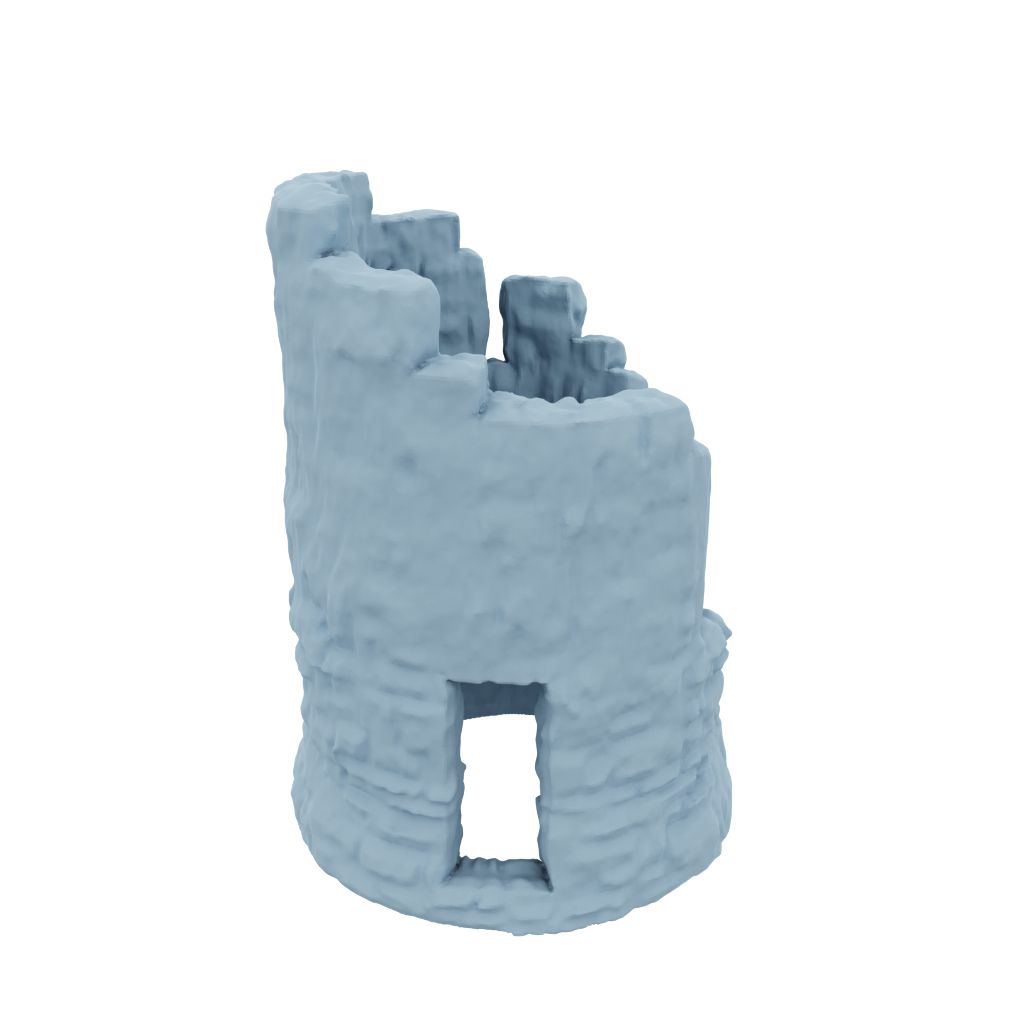}
    \end{subfigure}
    \begin{subfigure}{.16\linewidth}
        \centering
        \includegraphics[width=\linewidth]{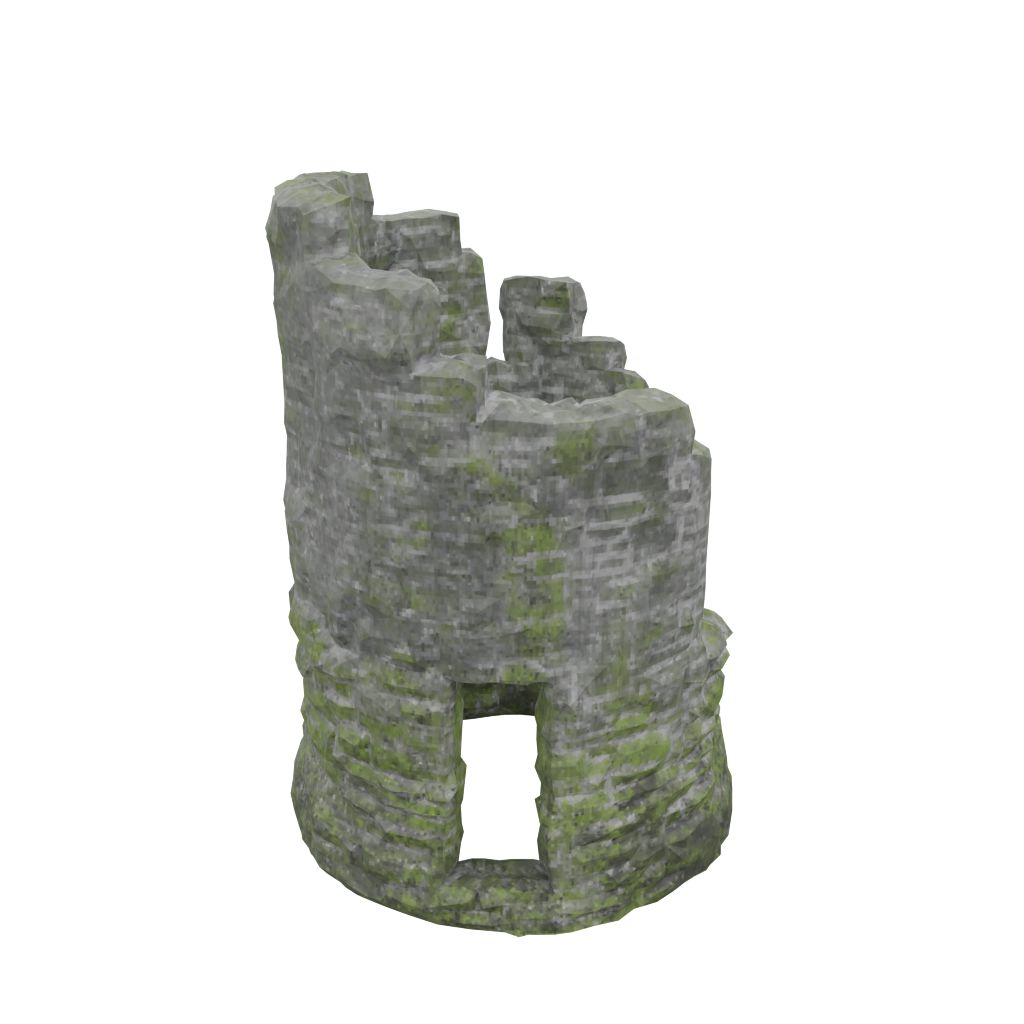}
    \end{subfigure}
    \begin{subfigure}{.16\linewidth}
        \centering
        \includegraphics[width=\linewidth]{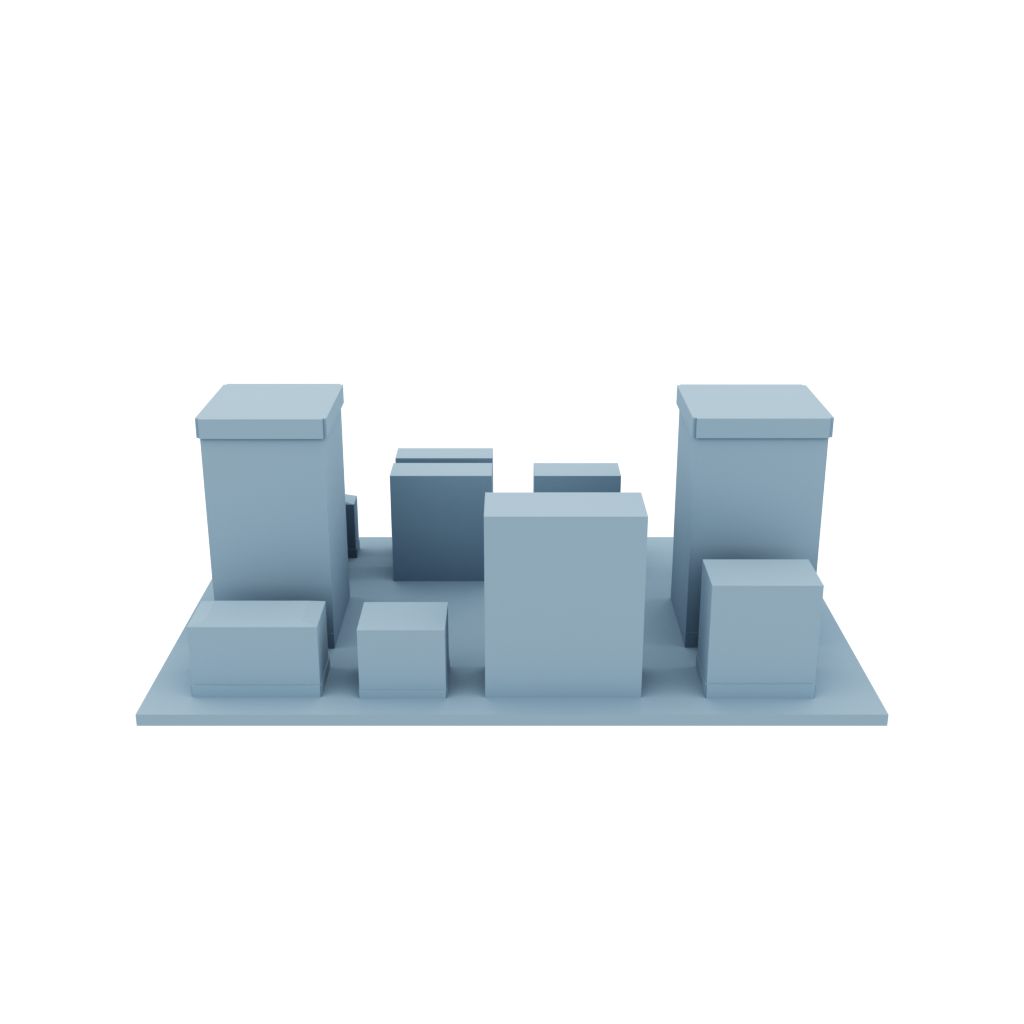}
    \end{subfigure}
    \begin{subfigure}{.16\linewidth}
        \centering
        \includegraphics[width=\linewidth]{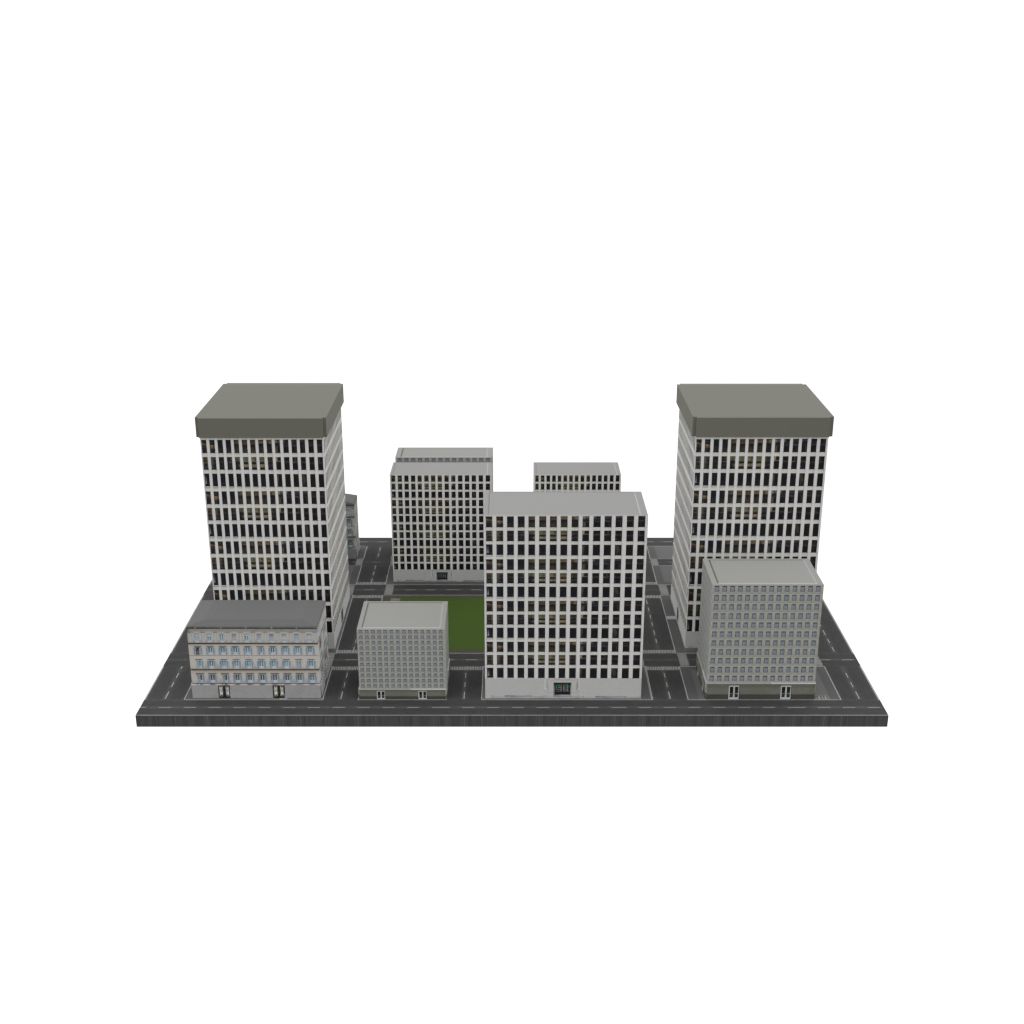}
    \end{subfigure}
    \unskip\ \vrule\ 
    \begin{subfigure}{.16\linewidth}
        \centering
        \includegraphics[width=\linewidth]{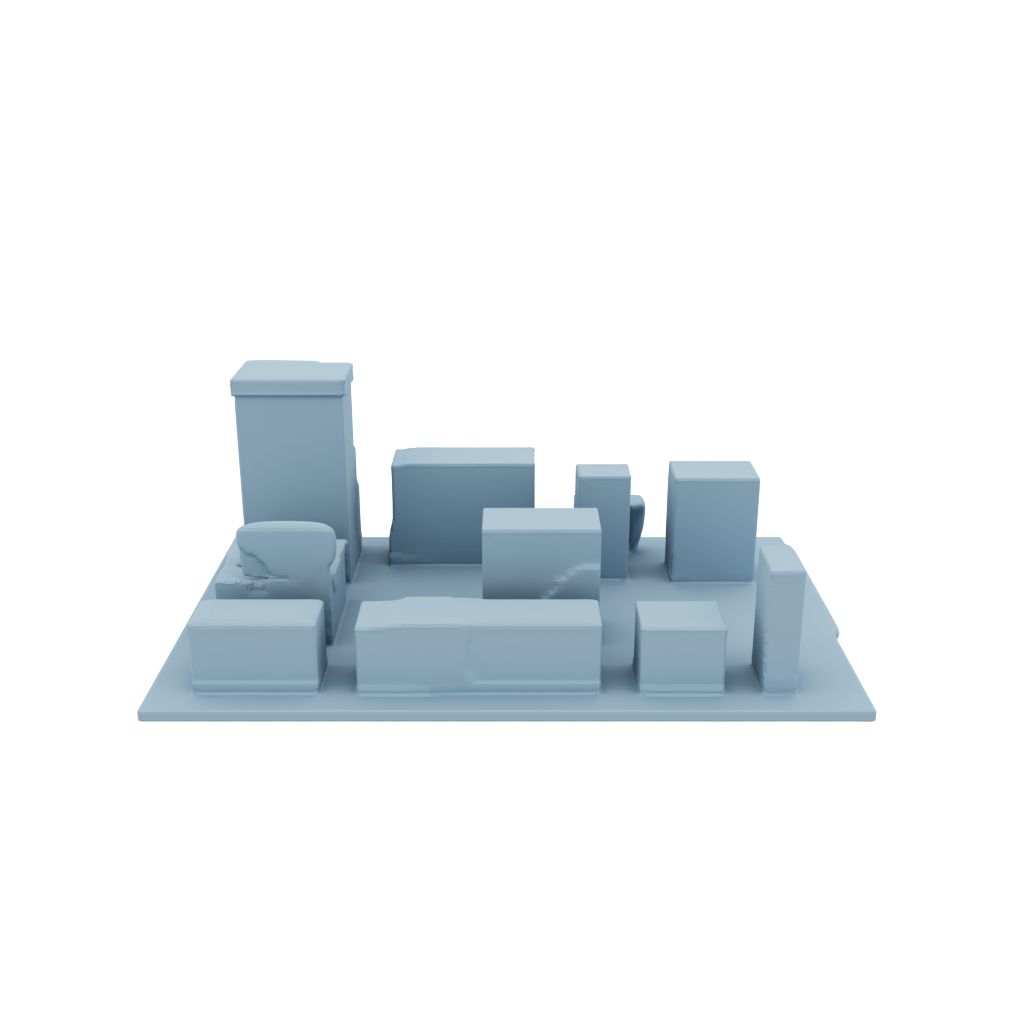}
    \end{subfigure}
    \begin{subfigure}{.16\linewidth}
        \centering
        \includegraphics[width=\linewidth]{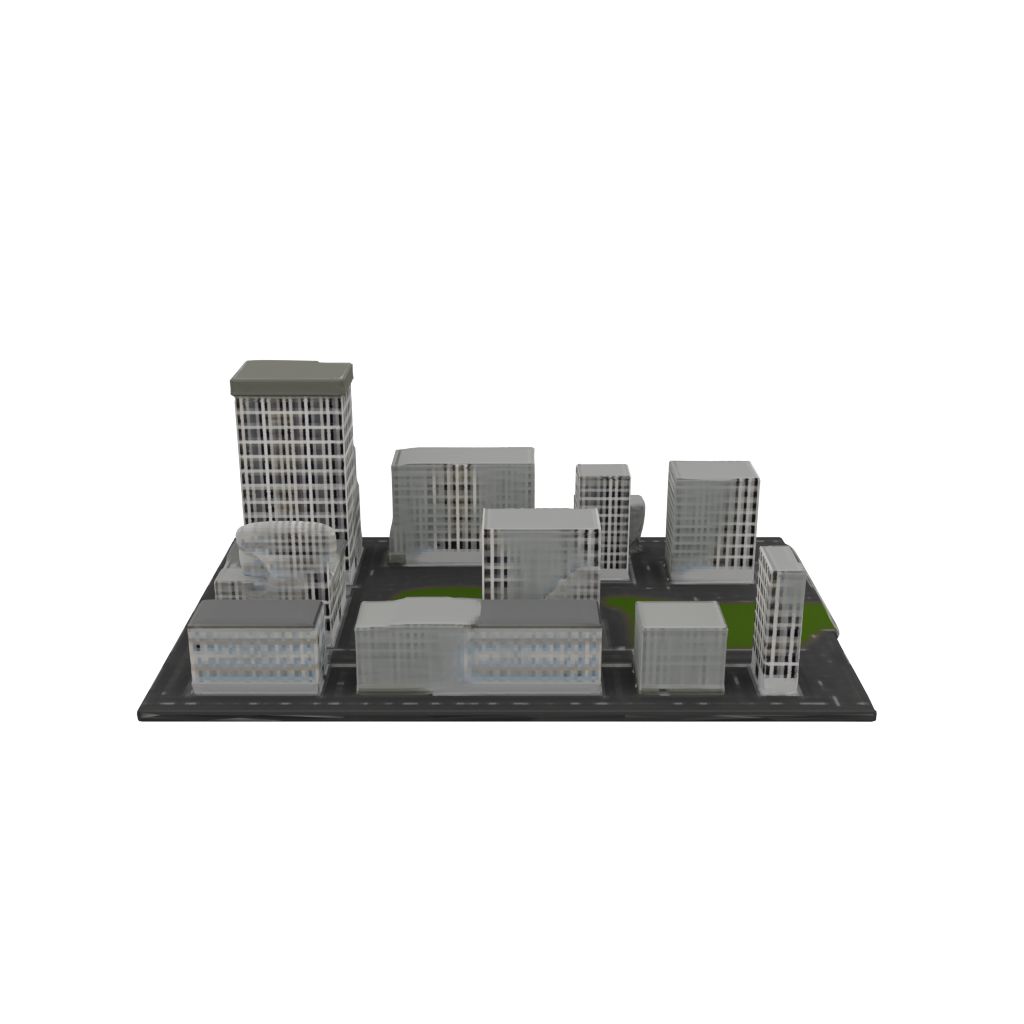}
    \end{subfigure}
    \begin{subfigure}{.16\linewidth}
        \centering
        \includegraphics[width=\linewidth]{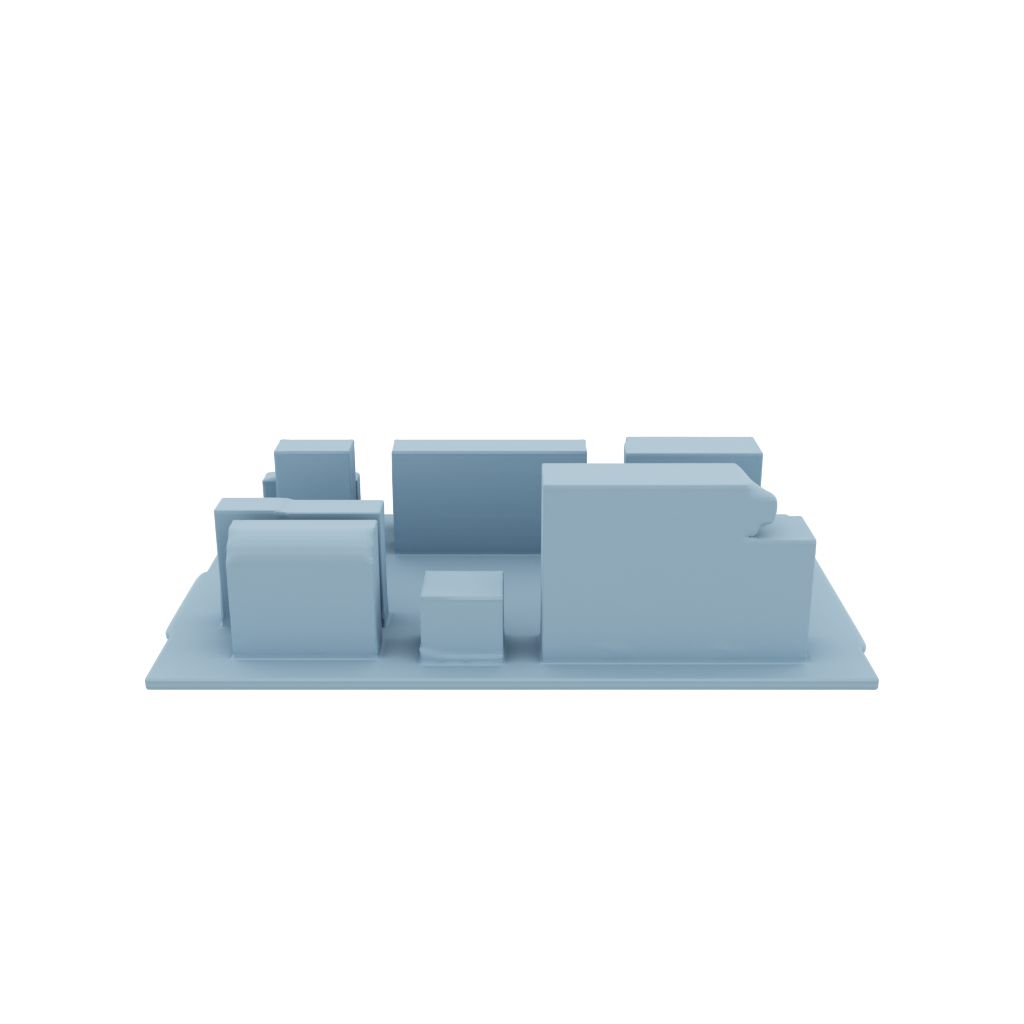}
    \end{subfigure}
    \begin{subfigure}{.16\linewidth}
        \centering
        \includegraphics[width=\linewidth]{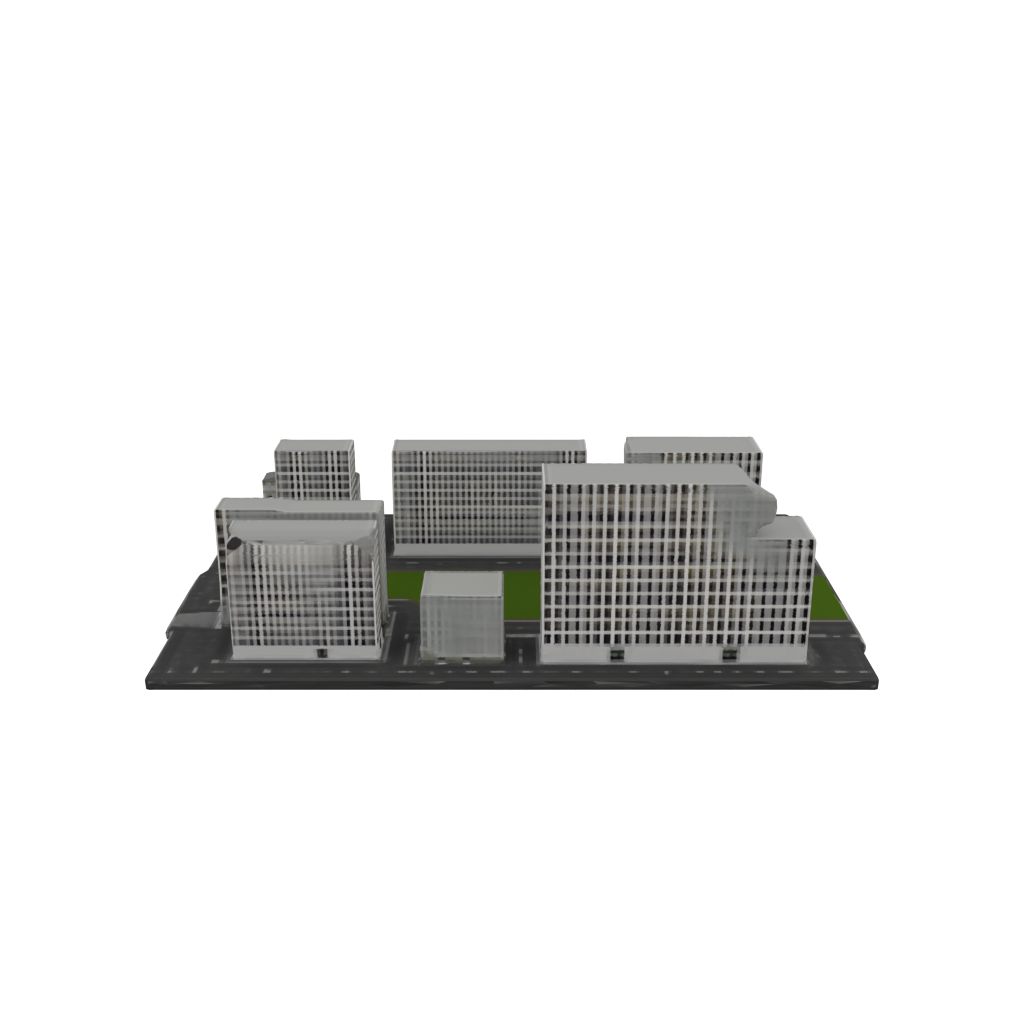}
    \end{subfigure}
    \begin{subfigure}{.16\linewidth}
        \centering
        \includegraphics[width=\linewidth]{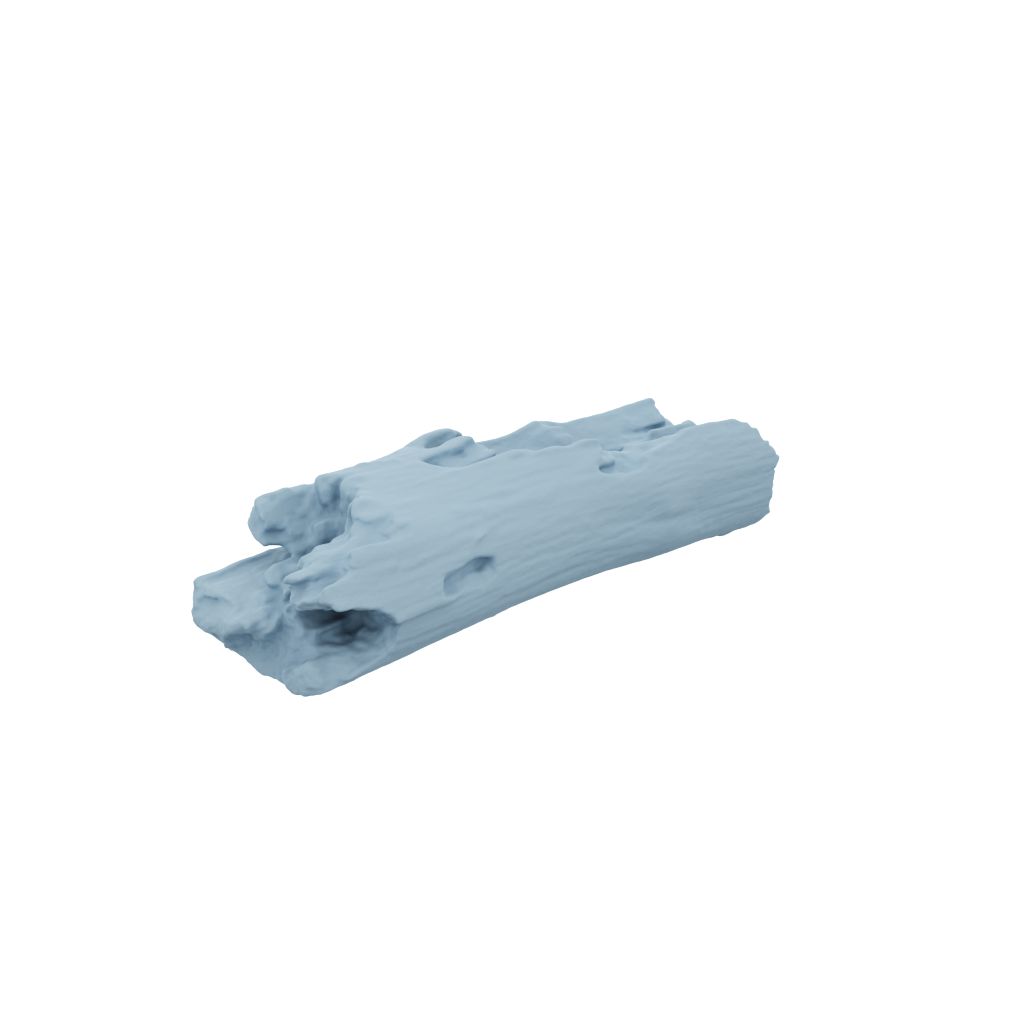}
    \end{subfigure}
    \begin{subfigure}{.16\linewidth}
        \centering
        \includegraphics[width=\linewidth]{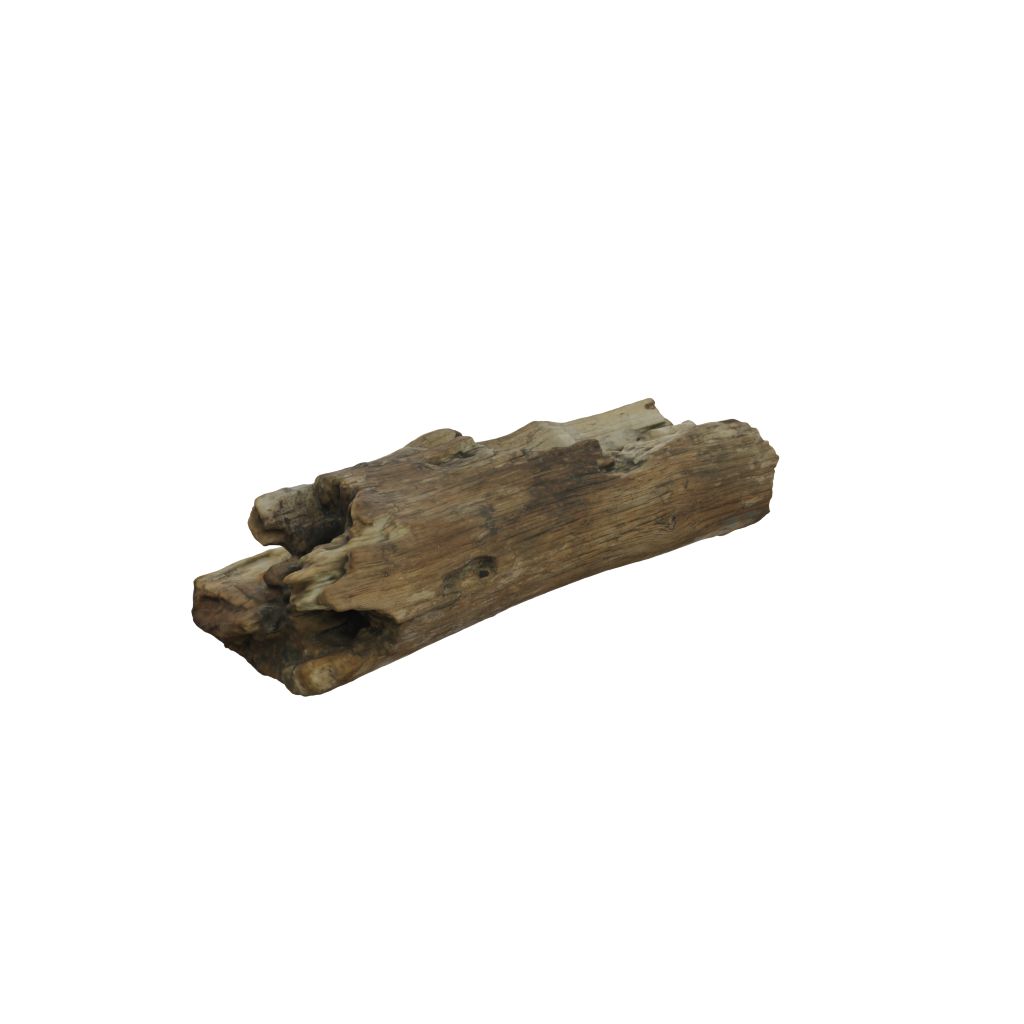}
    \end{subfigure}
    \unskip\ \vrule\ 
    \begin{subfigure}{.16\linewidth}
        \centering
        \includegraphics[width=\linewidth]{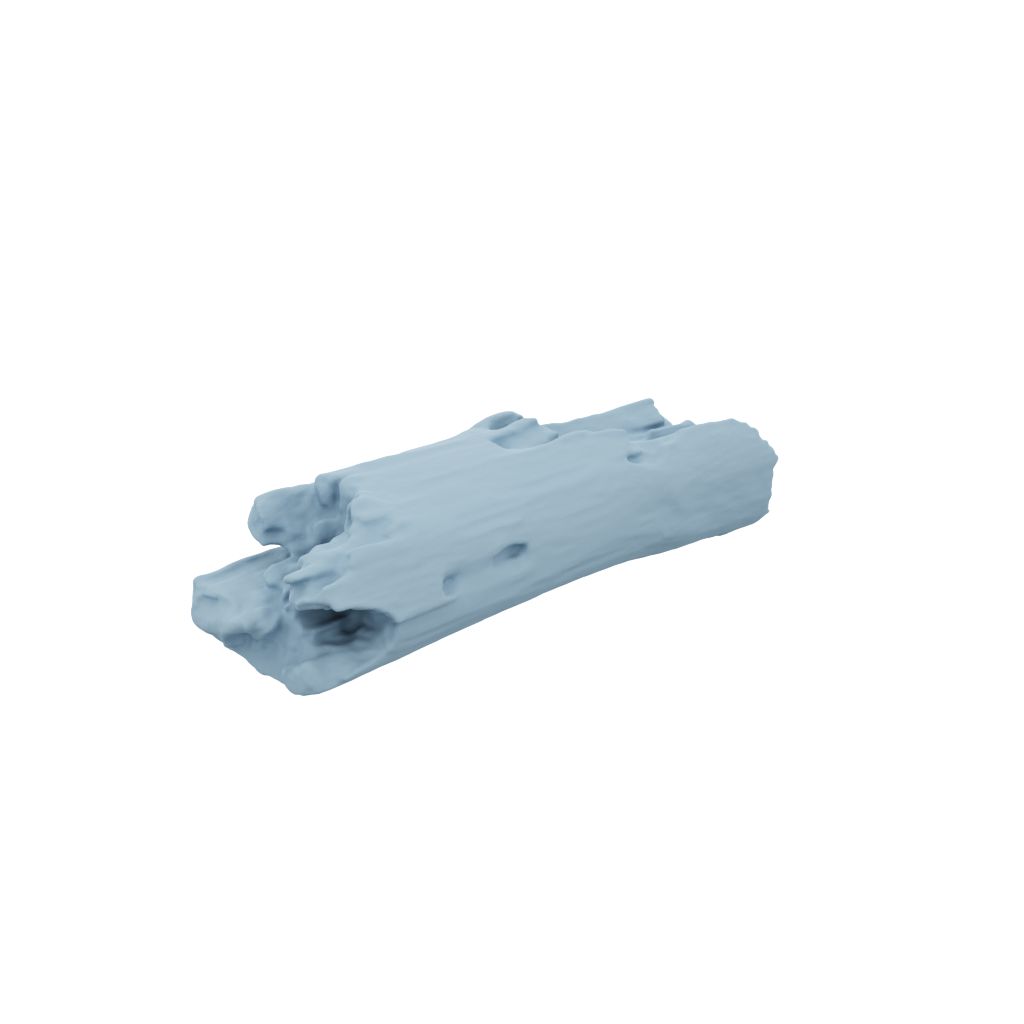}
    \end{subfigure}
    \begin{subfigure}{.16\linewidth}
        \centering
        \includegraphics[width=\linewidth]{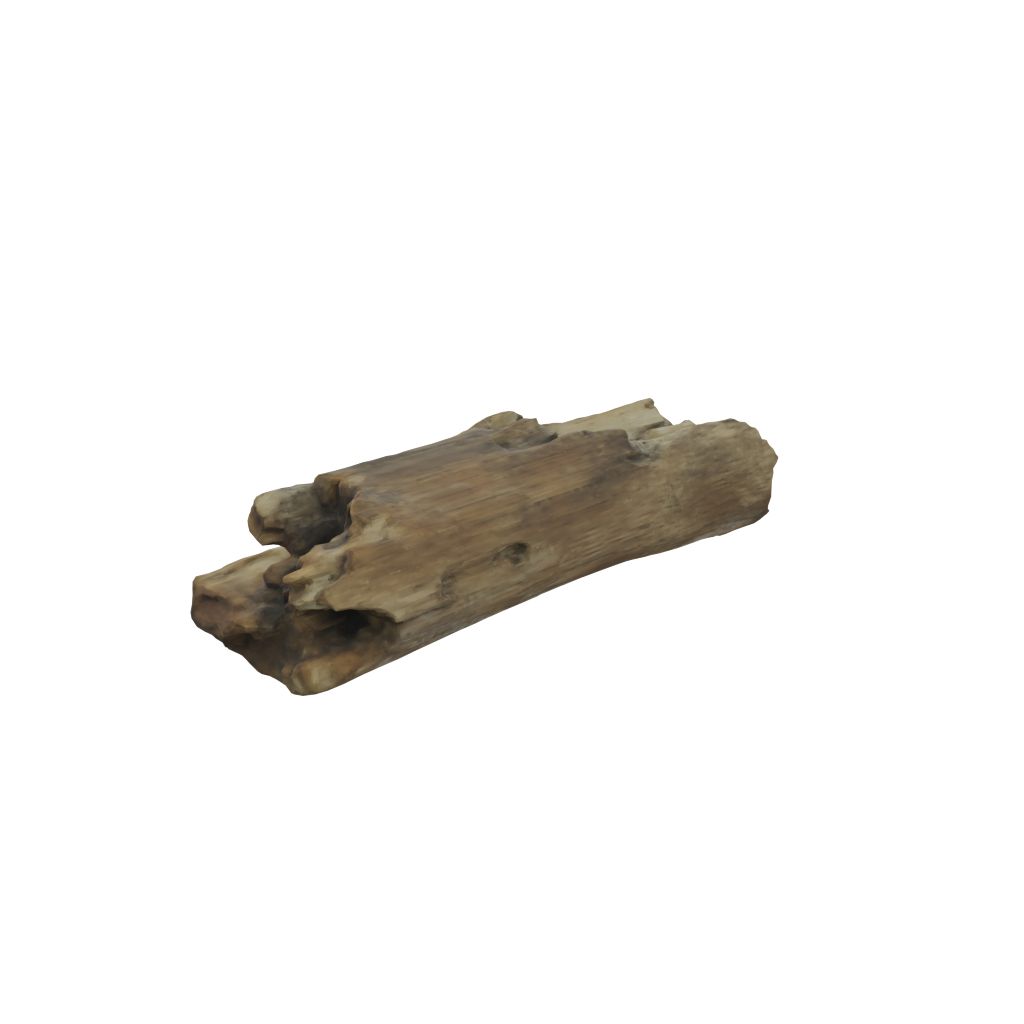}
    \end{subfigure}
    \begin{subfigure}{.16\linewidth}
        \centering
        \includegraphics[width=\linewidth]{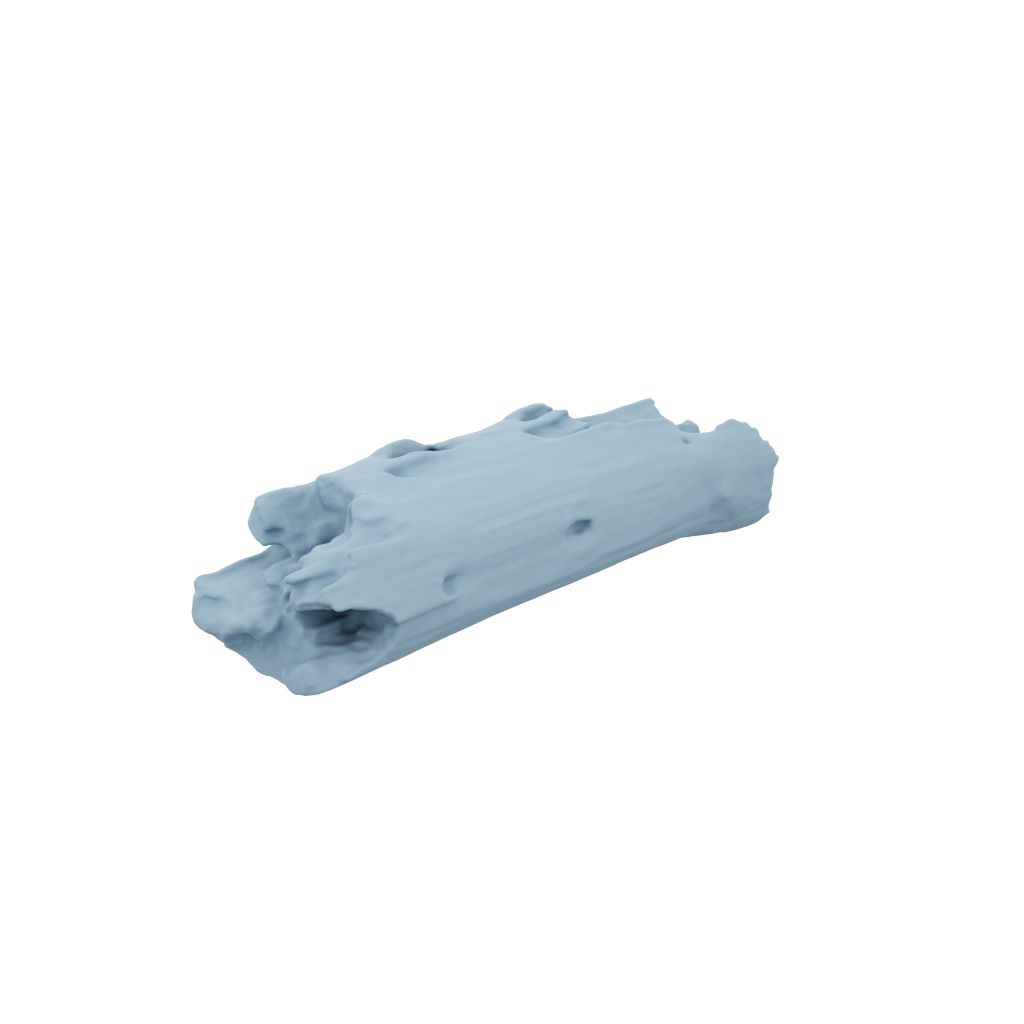}
    \end{subfigure}
    \begin{subfigure}{.16\linewidth}
        \centering
        \includegraphics[width=\linewidth]{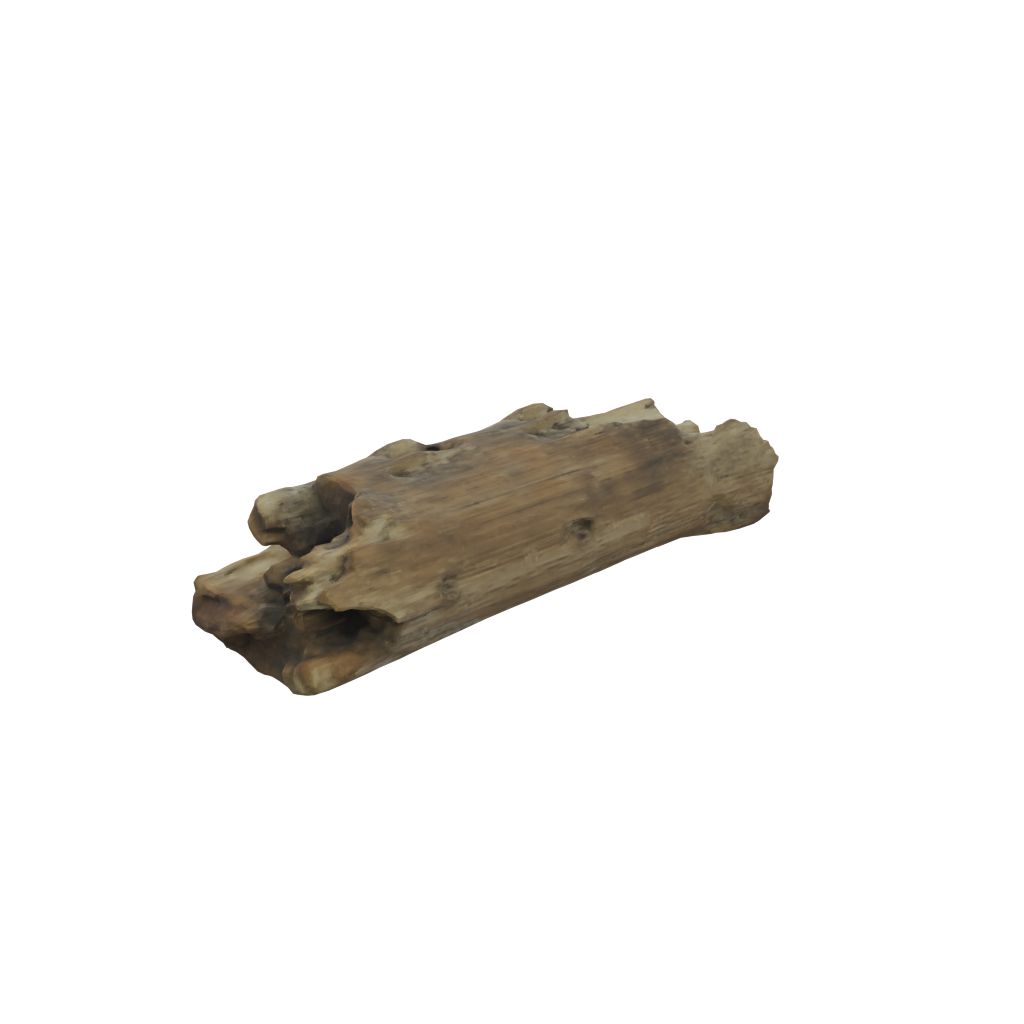}
    \end{subfigure}
        \vspace*{-5mm} 

    \subfloat[Input Shape]{\hspace{.333\linewidth}}
      \subfloat[Generated Shape (ours)]{\hspace{.333\linewidth}}
        \subfloat[Generated Shape (ours)]{\hspace{.333\linewidth}}
    \caption{\textbf{Samples of our results I.} This figure shows a variety of input models and some of the generated variants (both shown without and with texture to facilitate visual inspection) ShapeShifter outputs.}
\label{fig:all_renderings0}
\end{figure*}

\begin{figure*}[!h]
    \centering
    \begin{subfigure}{.16\linewidth}
        \centering
        \includegraphics[width=\linewidth]{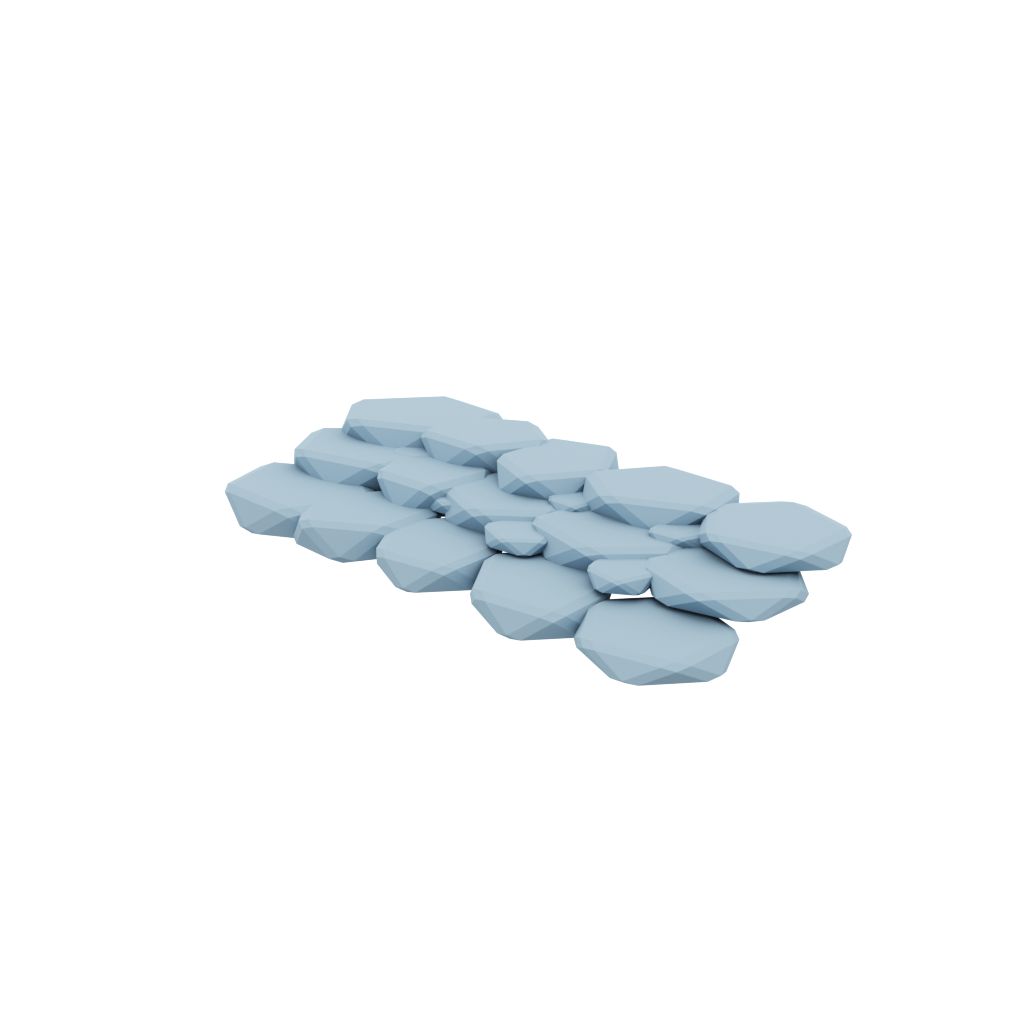}
    \end{subfigure}
    \begin{subfigure}{.16\linewidth}
        \centering
        \includegraphics[width=\linewidth]{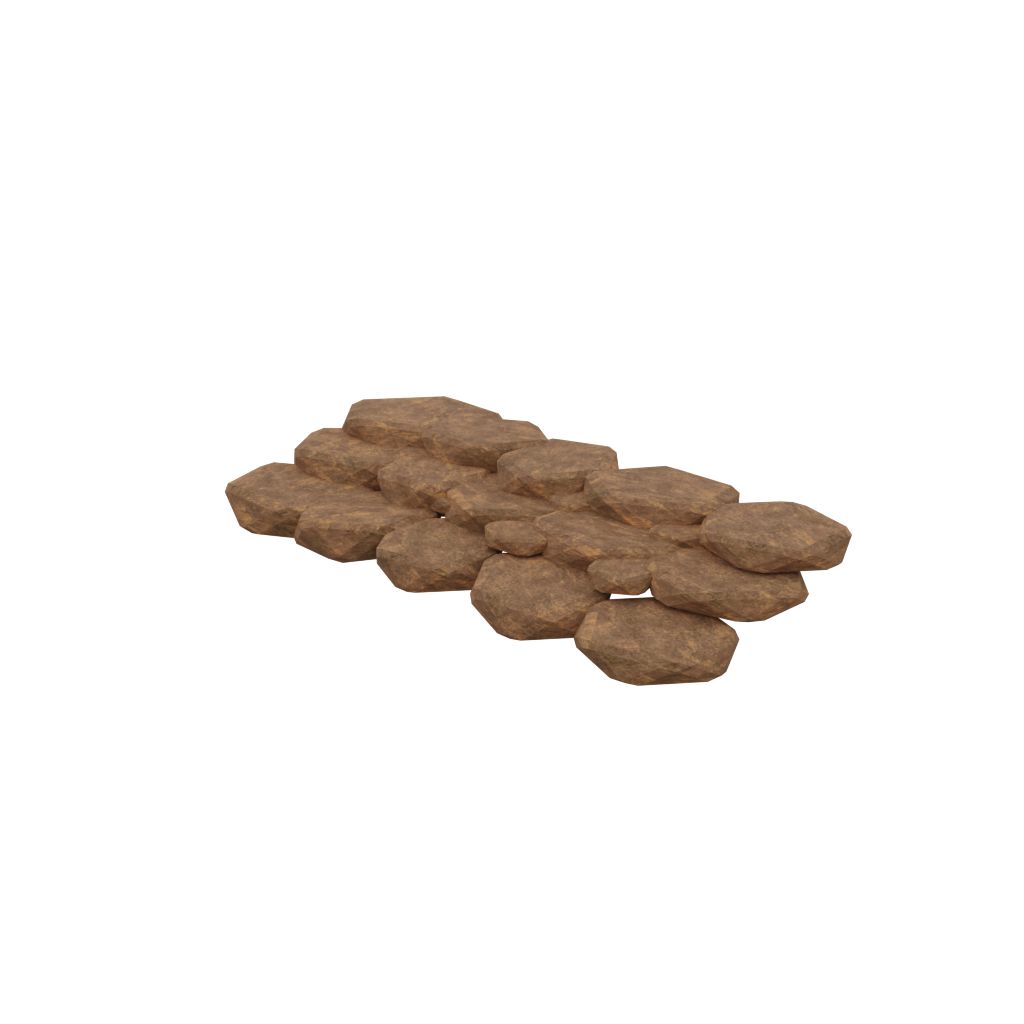}
    \end{subfigure}
    \unskip\ \vrule\ 
    \begin{subfigure}{.16\linewidth}
        \centering
        \includegraphics[width=\linewidth]{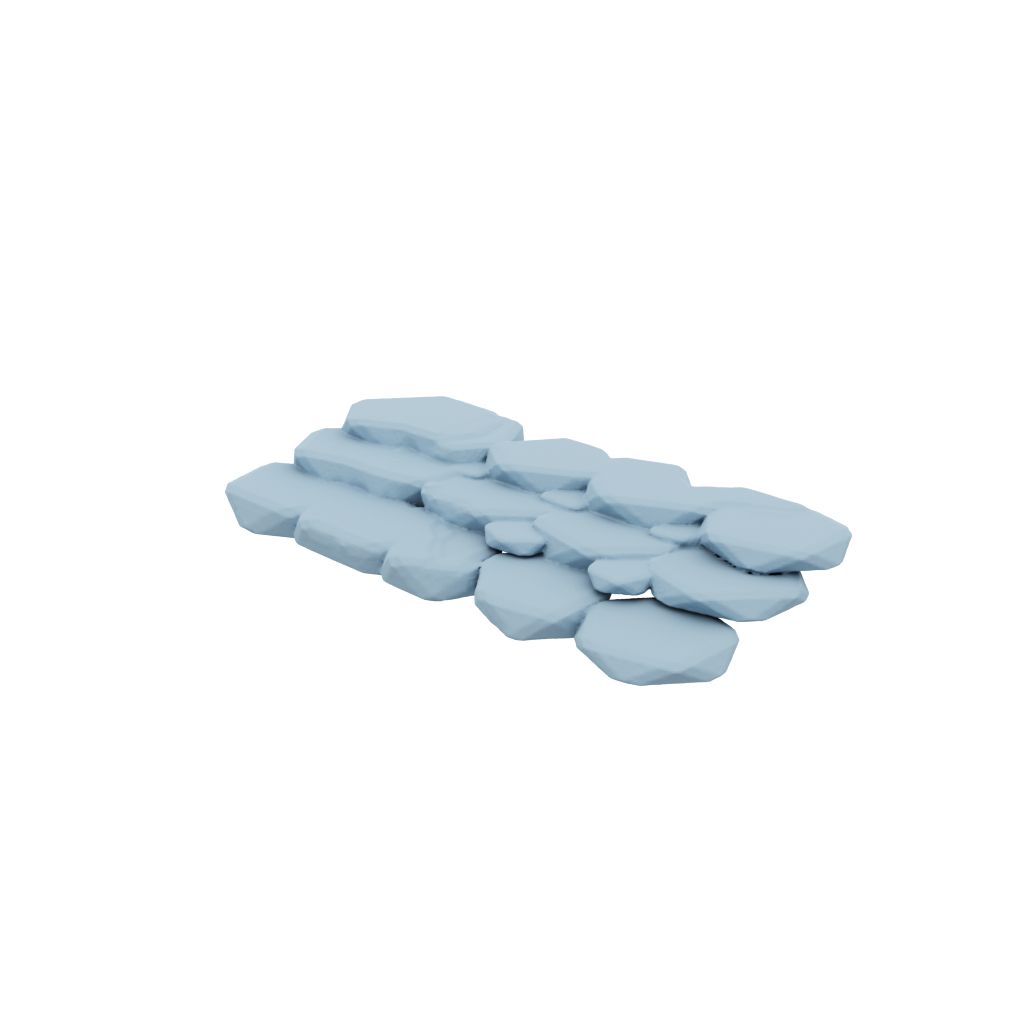}
    \end{subfigure}
    \begin{subfigure}{.16\linewidth}
        \centering
        \includegraphics[width=\linewidth]{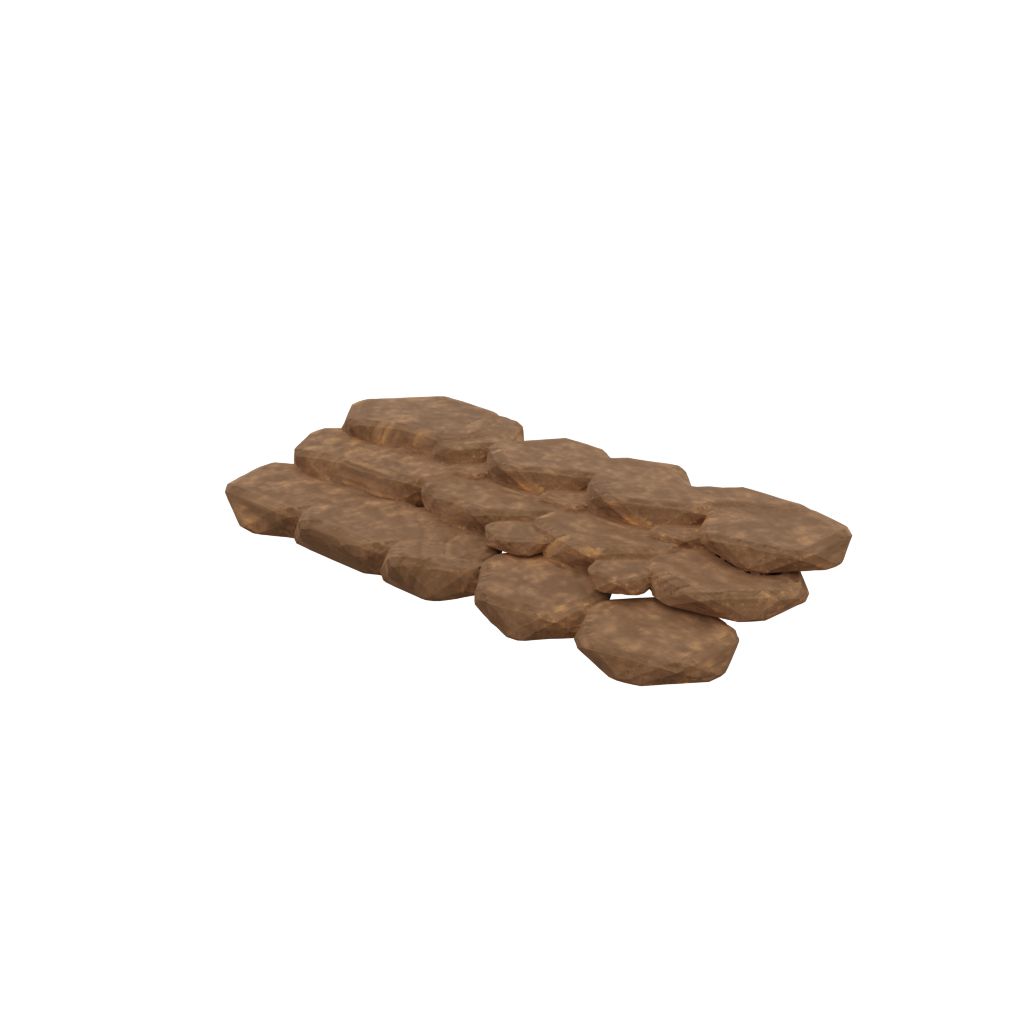}
    \end{subfigure}
    \begin{subfigure}{.16\linewidth}
        \centering
        \includegraphics[width=\linewidth]{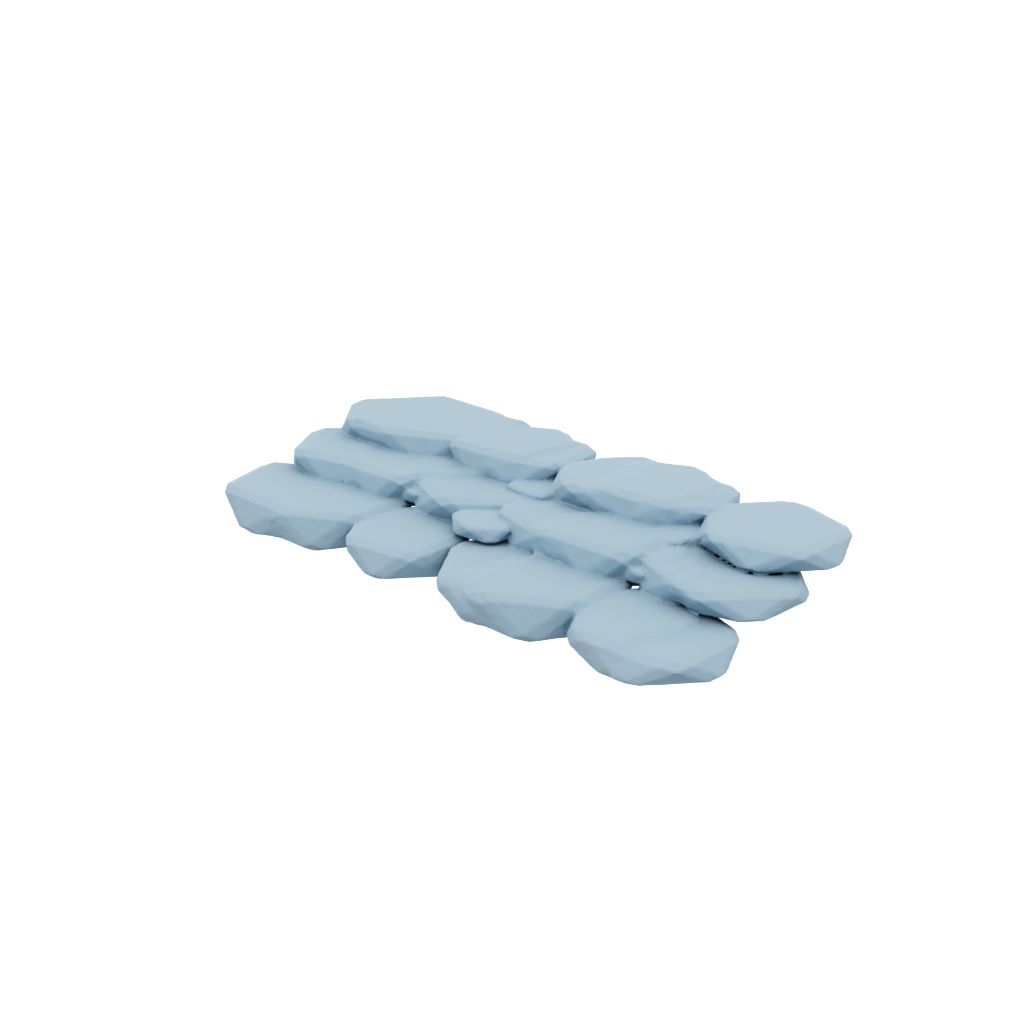}
    \end{subfigure}
    \begin{subfigure}{.16\linewidth}
        \centering
        \includegraphics[width=\linewidth]{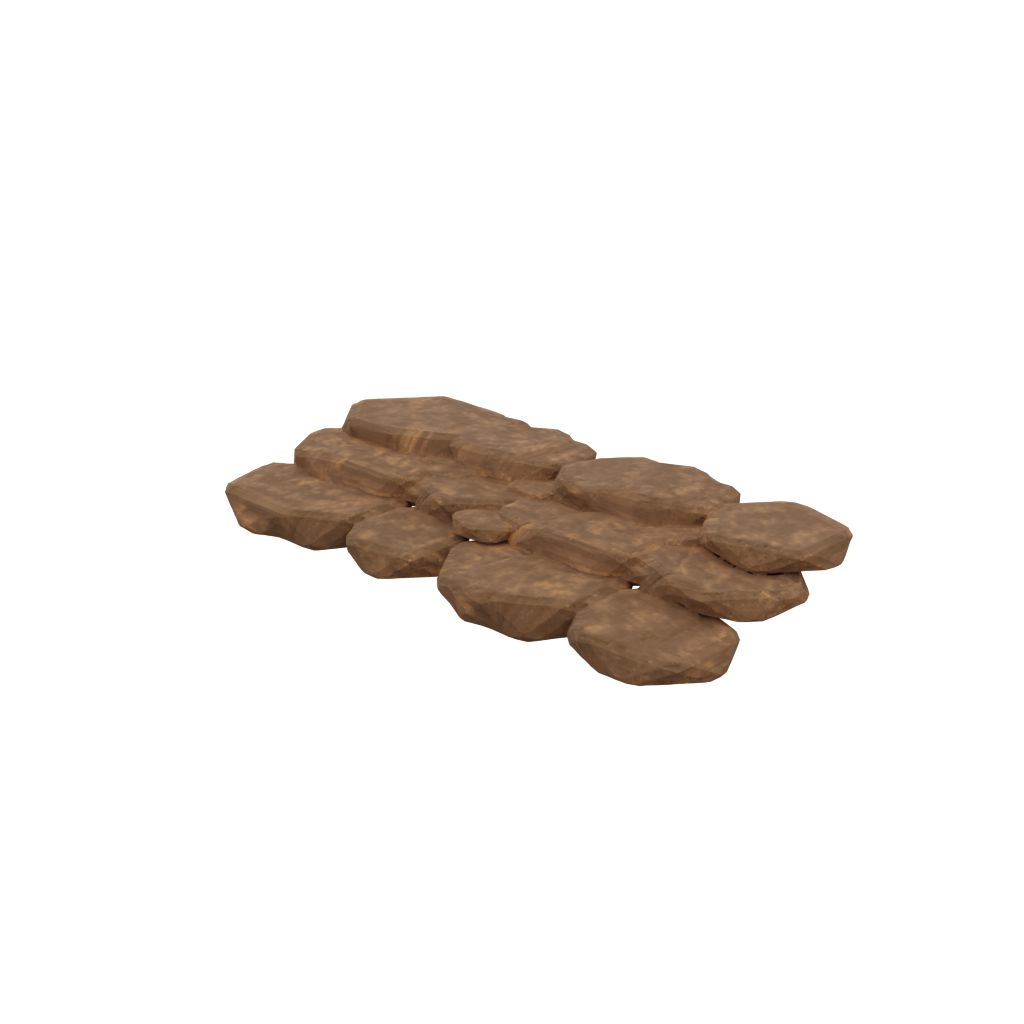}
    \end{subfigure}
    \vspace*{-5mm} 
    \begin{subfigure}{.16\linewidth}
        \centering
        \includegraphics[width=\linewidth]{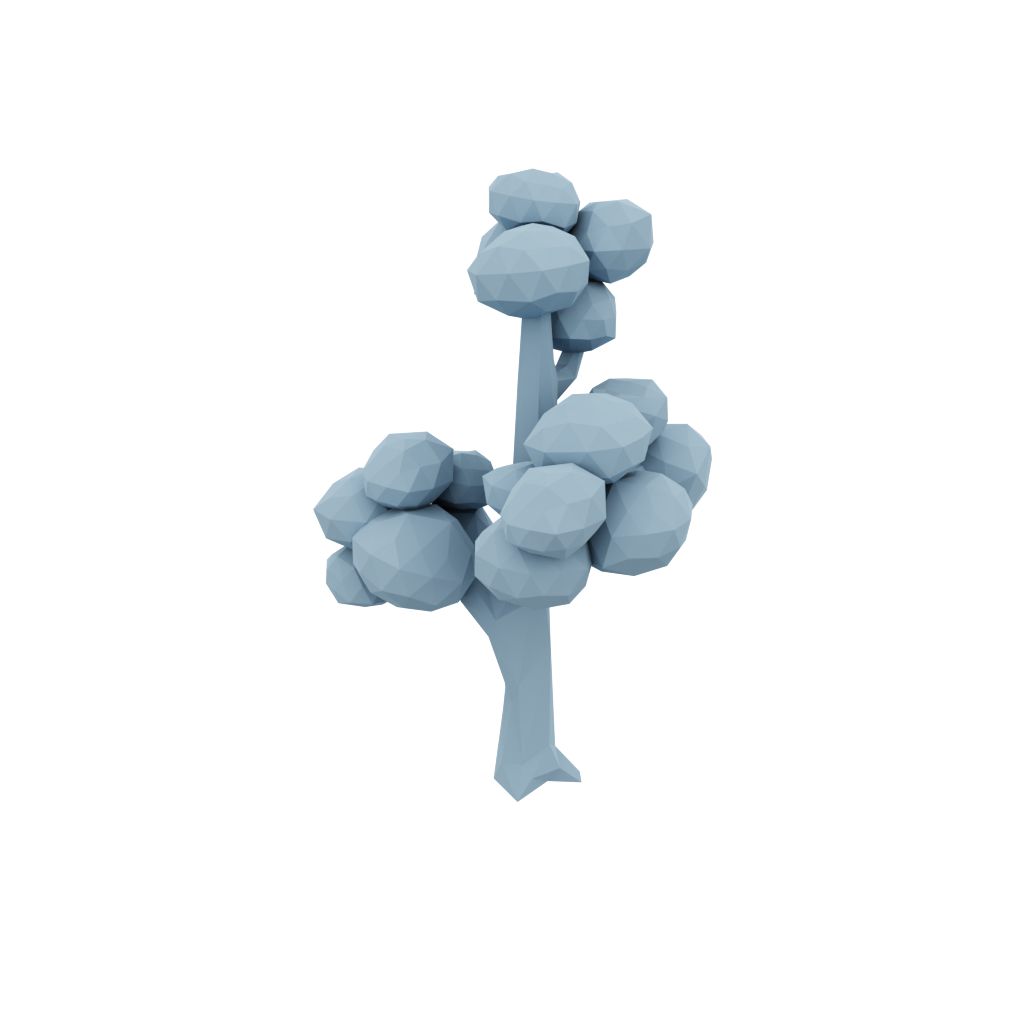}
    \end{subfigure}
    \begin{subfigure}{.16\linewidth}
        \centering
        \includegraphics[width=\linewidth]{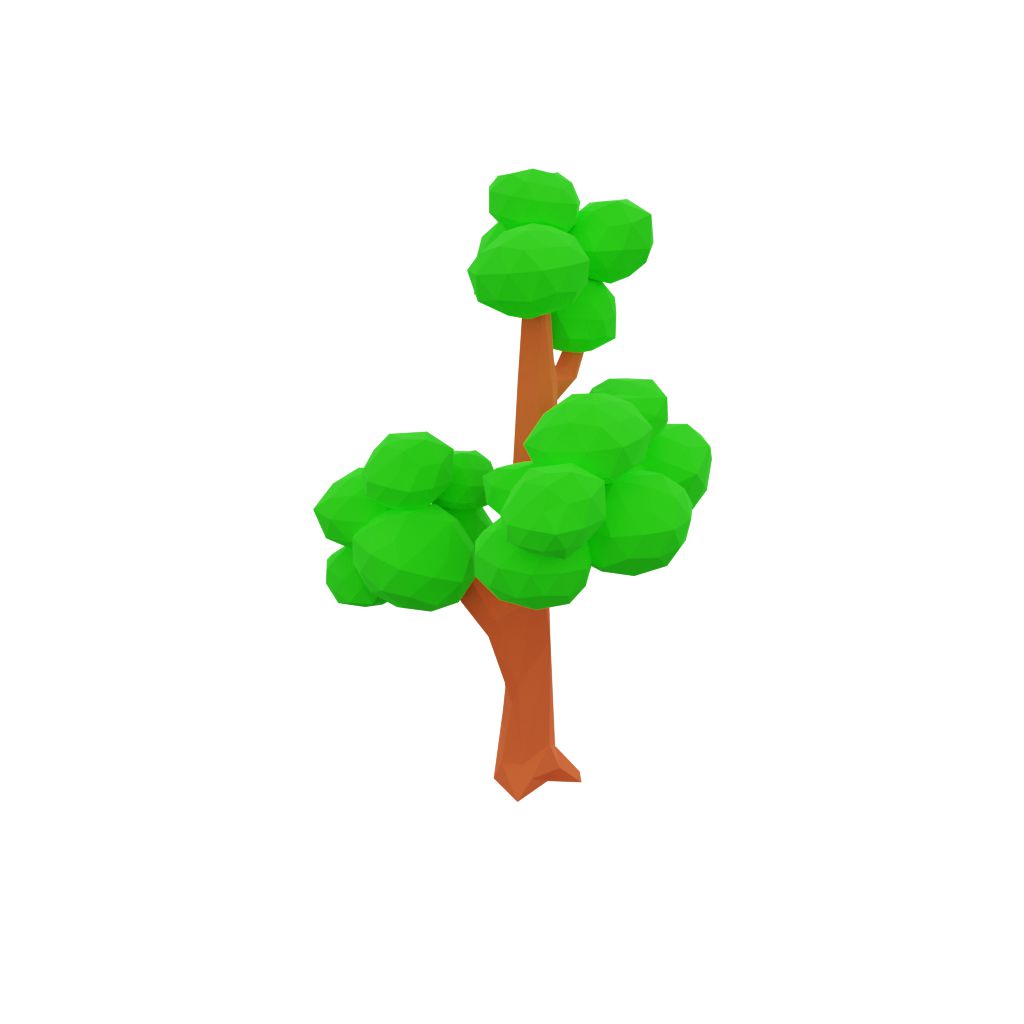}
    \end{subfigure}
    \unskip\ \vrule\ 
    \begin{subfigure}{.16\linewidth}
        \centering
        \includegraphics[width=\linewidth]{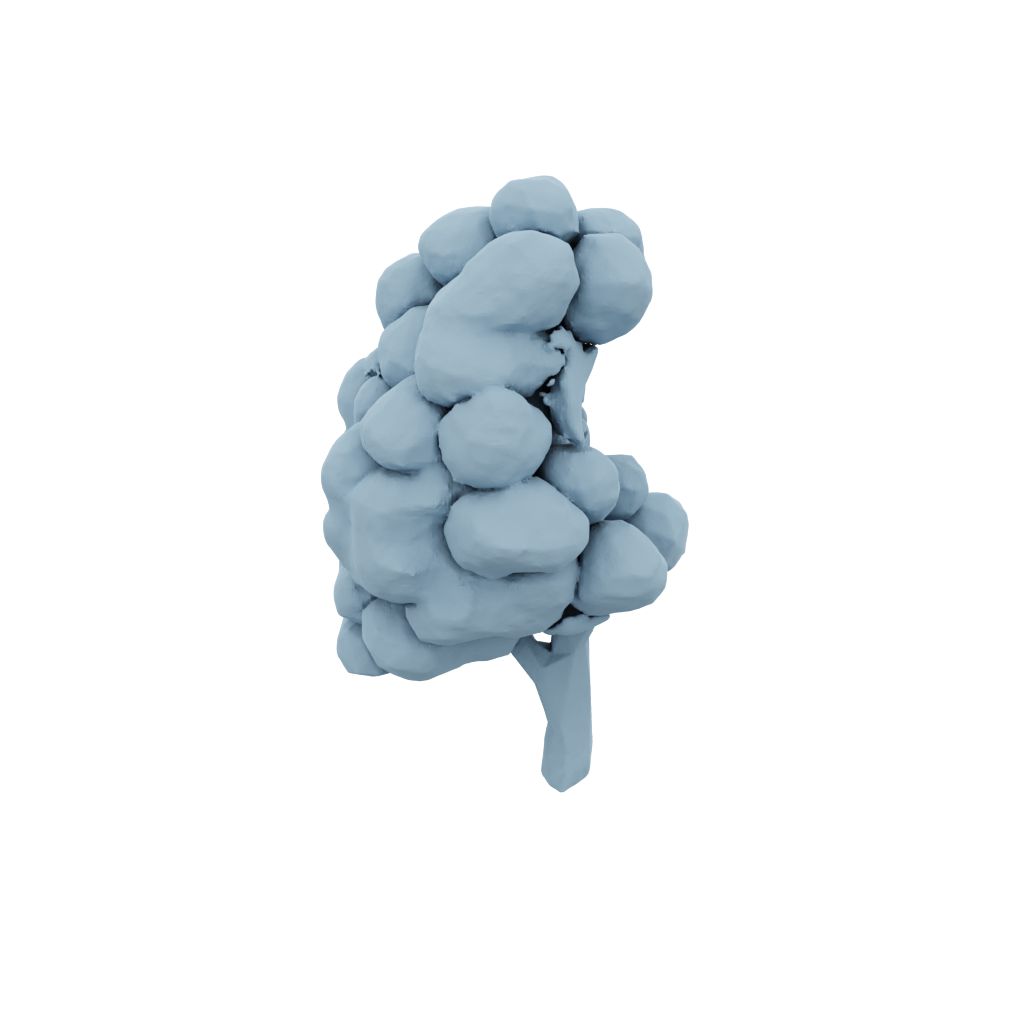}
    \end{subfigure}
    \begin{subfigure}{.16\linewidth}
        \centering
        \includegraphics[width=\linewidth]{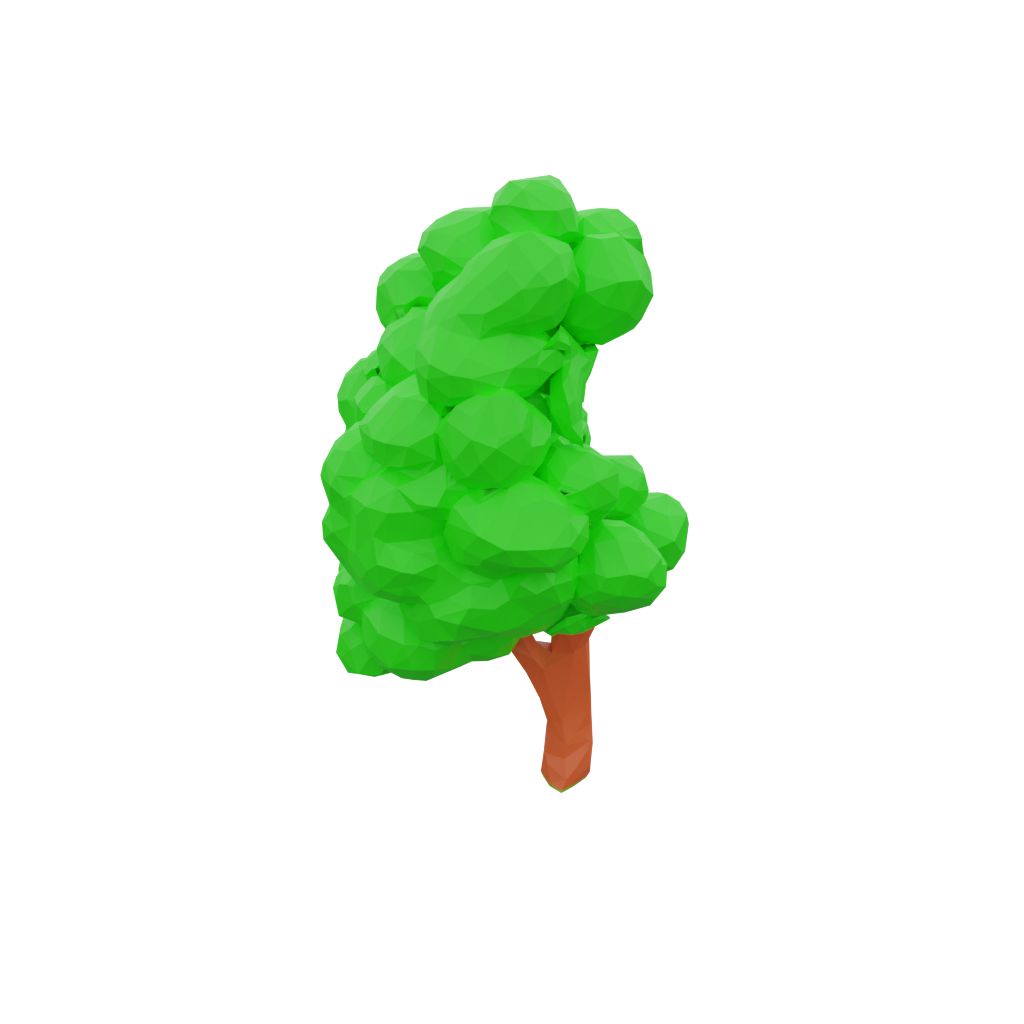}
    \end{subfigure}
    \begin{subfigure}{.16\linewidth}
        \centering
        \includegraphics[width=\linewidth]{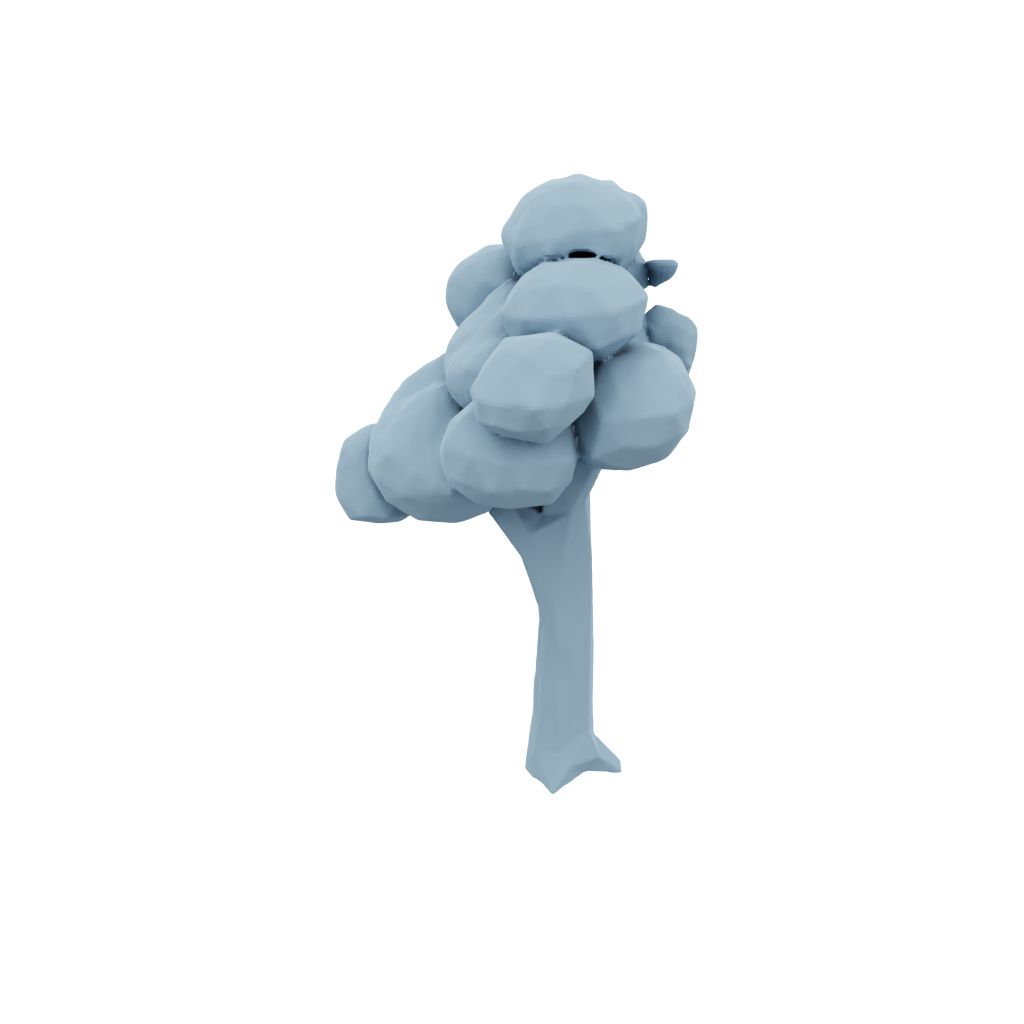}
    \end{subfigure}
    \begin{subfigure}{.16\linewidth}
        \centering
        \includegraphics[width=\linewidth]{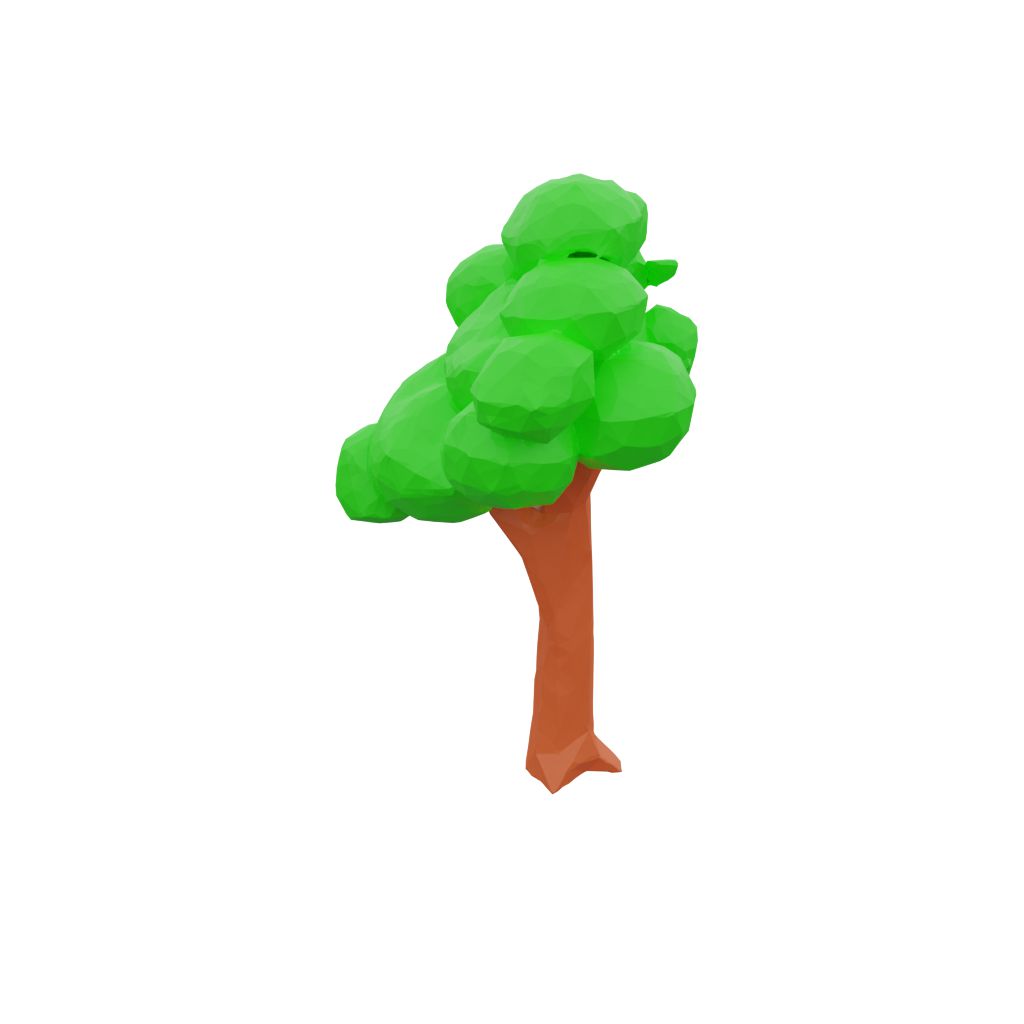}
    \end{subfigure}
    \vspace*{-5mm} 
    \begin{subfigure}{.315\linewidth}
        \centering
        \includegraphics[width=0.7\linewidth]{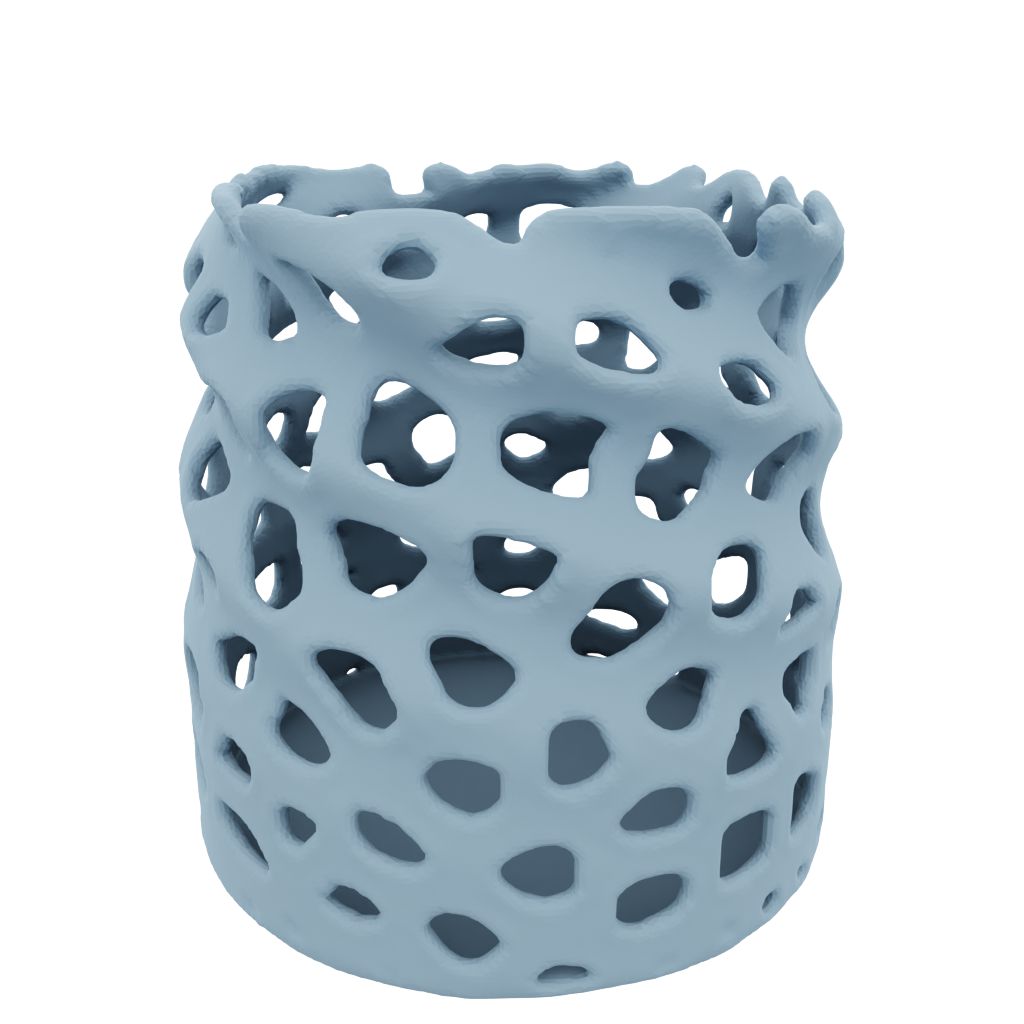}
    \end{subfigure}
    \unskip\ \vrule\ 
    \begin{subfigure}{.32\linewidth}
        \centering
        \includegraphics[width=0.7\linewidth]{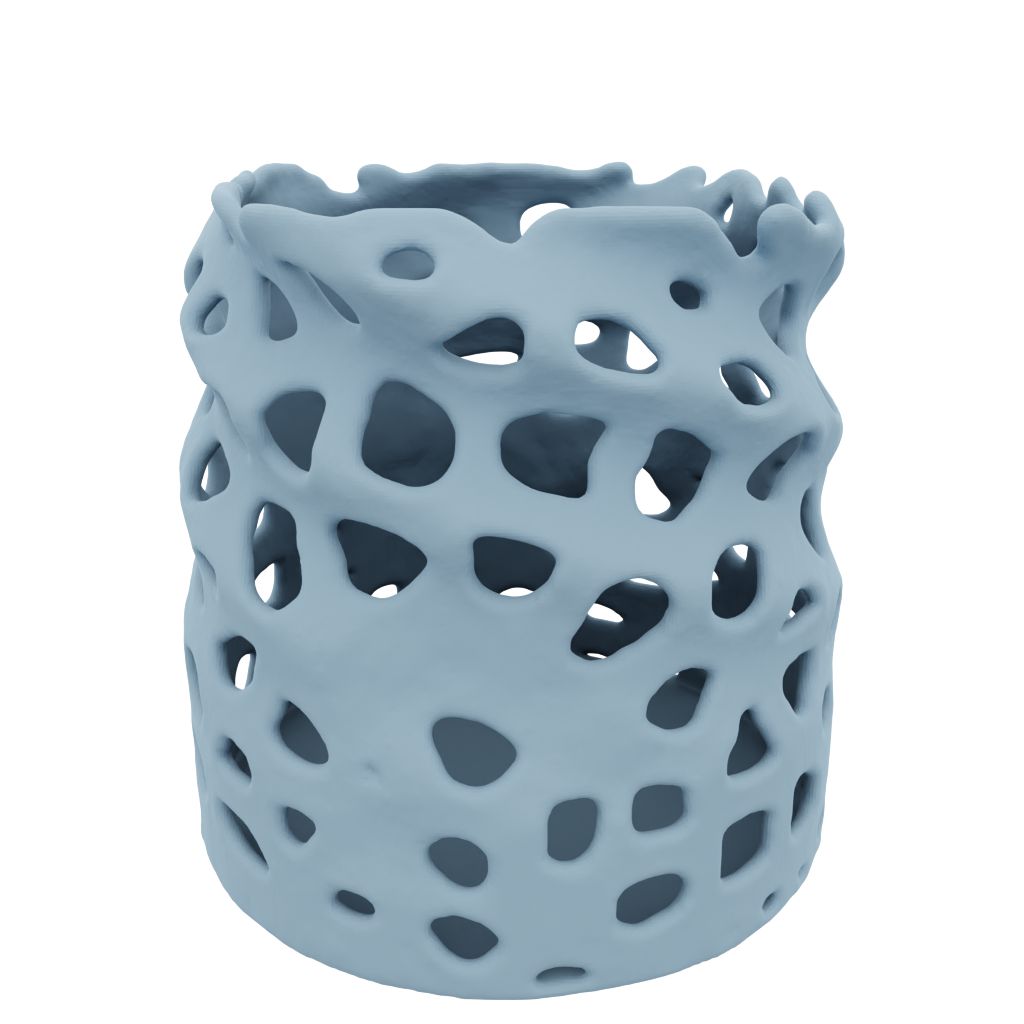}
    \end{subfigure}
    \begin{subfigure}{.32\linewidth}
        \centering
        \includegraphics[width=0.7\linewidth]{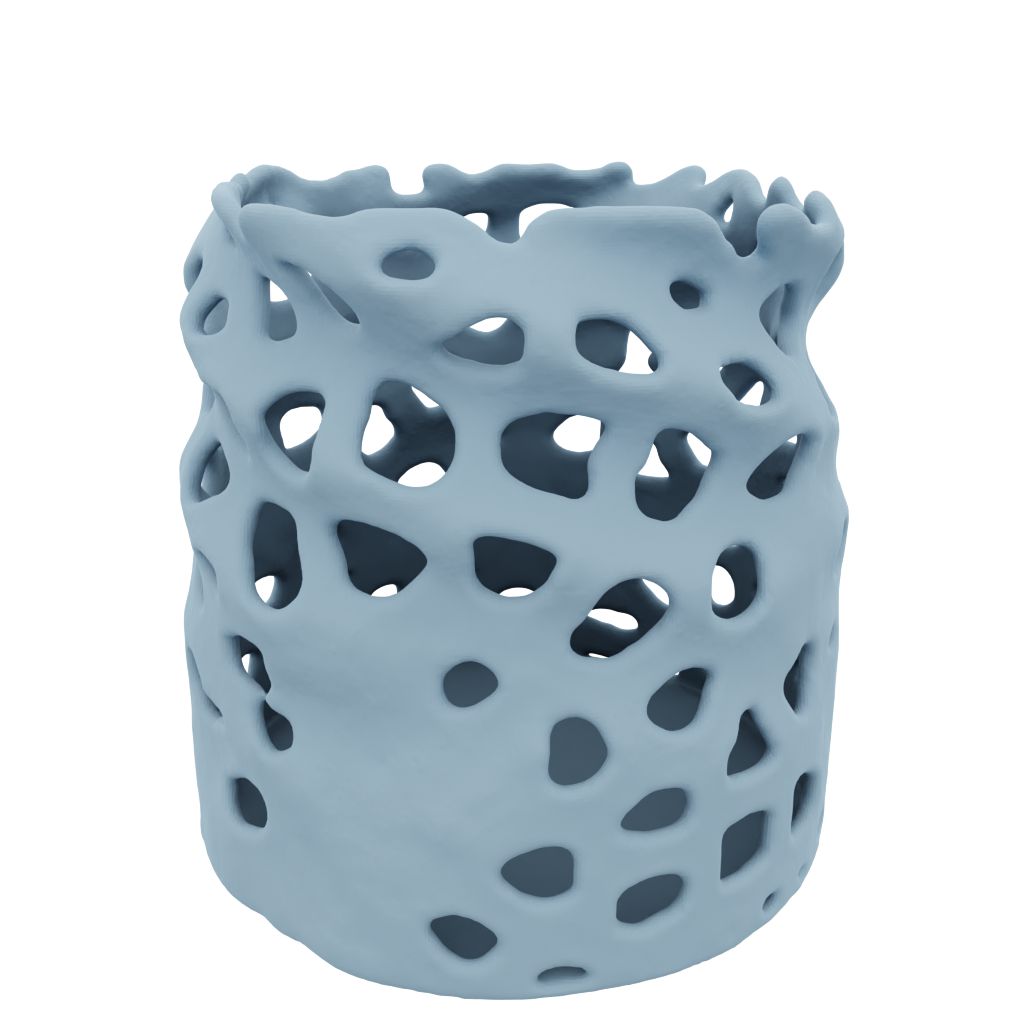}
    \end{subfigure}
     \centering
    \begin{subfigure}{.32\linewidth}
        \centering
        \includegraphics[width=\linewidth]{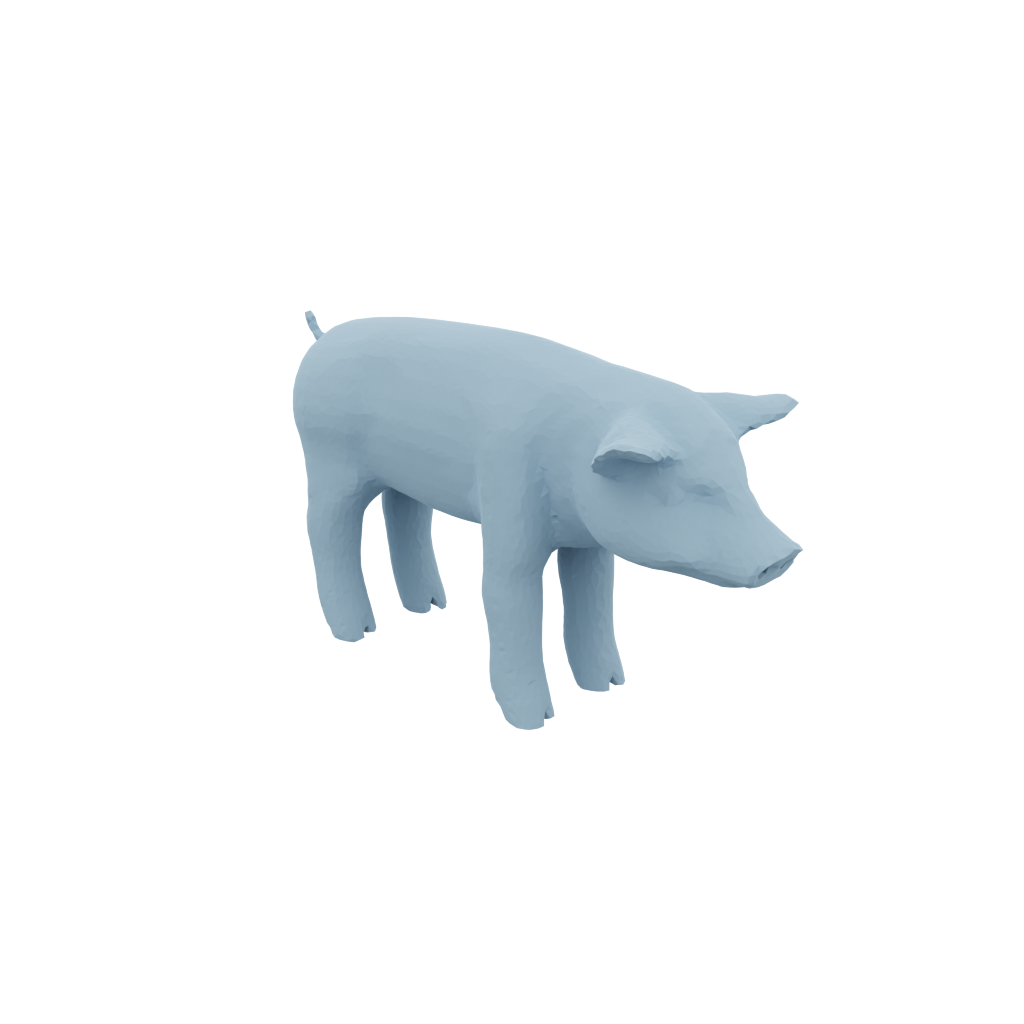}
    \end{subfigure}
    \unskip\ \vrule\ 
    \begin{subfigure}{.32\linewidth}
        \centering
        \includegraphics[width=\linewidth]{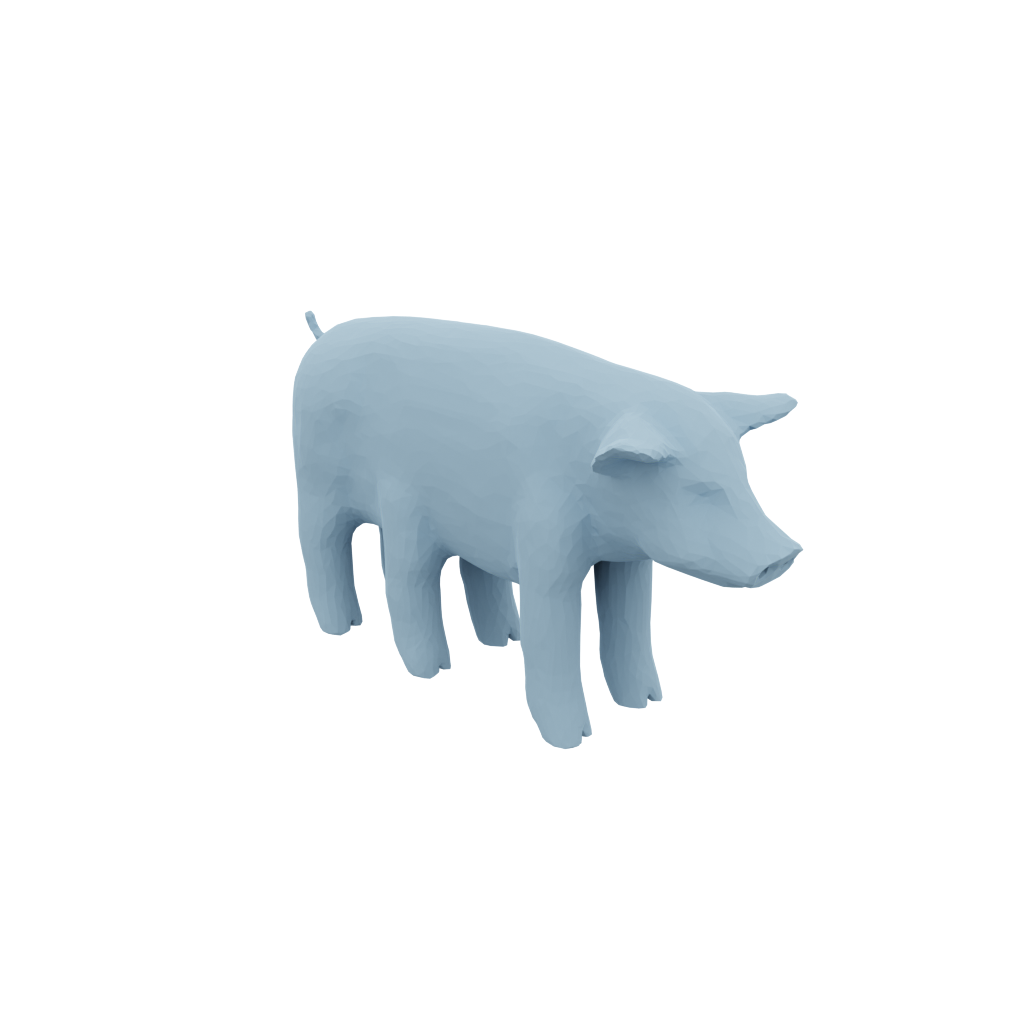}
    \end{subfigure}
    \begin{subfigure}{.32\linewidth}
        \centering
        \includegraphics[width=\linewidth]{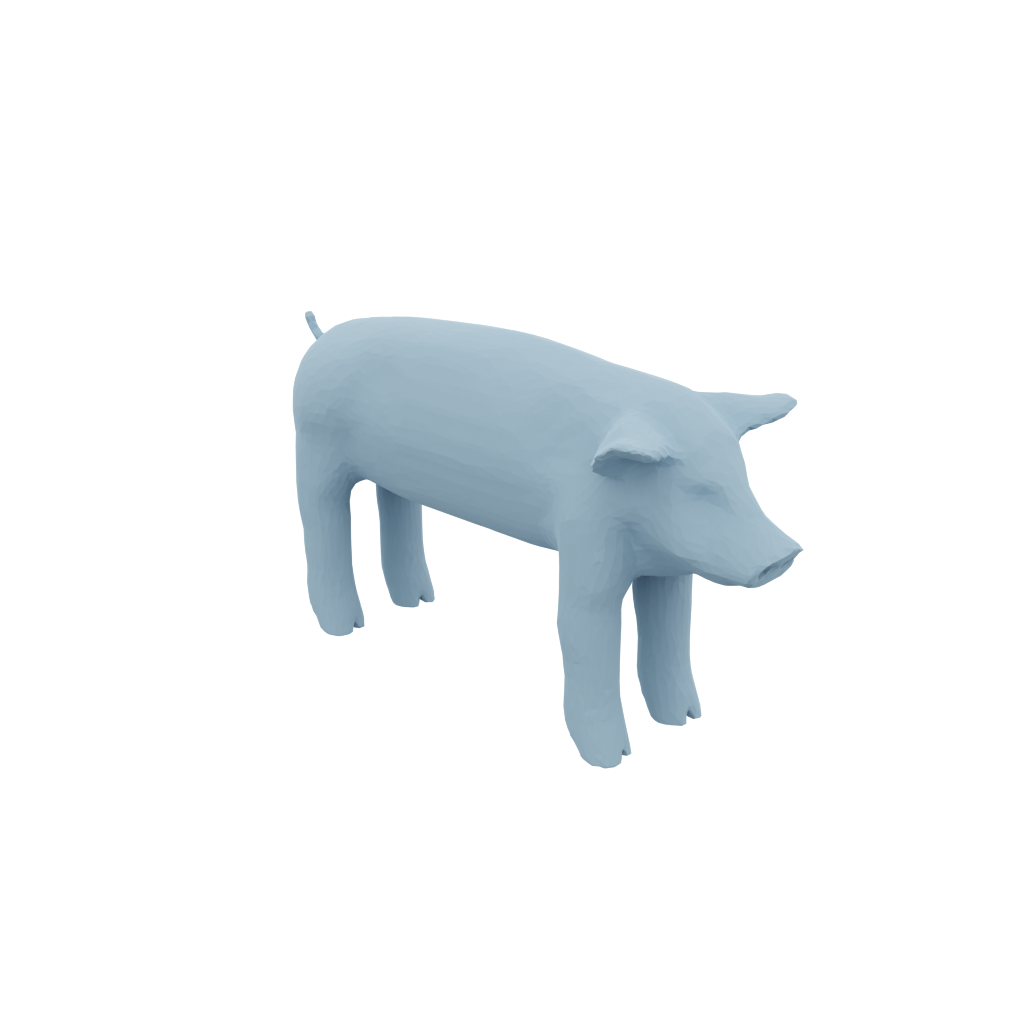}
    \end{subfigure}
    \vspace*{3mm} 

    \subfloat[Input Shape]{\hspace{.333\linewidth}}
      \subfloat[Generated Shape (ours)]{\hspace{.333\linewidth}}
        \subfloat[Generated Shape (ours)]{\hspace{.333\linewidth}}
\caption{\textbf{Samples of our results II.} This figure shows input models that were not used in the main paper, and some of the generated variants (both shown without and with texture to facilitate visual inspection) ShapeShifter outputs. Note that the last two examples (vase and pig) are an ablation test where we do not use colors among the per-voxel features in our approach.}
\label{fig:all_renderings}
\end{figure*}

We provide more results in Fig.~\ref{fig:ours_0}, \ref{fig:all_renderings0}, and~\ref{fig:all_renderings} to better illustrate the outputs of ShapeShifter on a variety of reference models. 
Note that we also show that \ourmethod can generate purely geometric variants from untextured meshes, see last result in Fig.~\ref{fig:all_renderings}.

\subsection{Comparison to SSG}
\label{sec:add_method}
\begin{figure}[!h]
\centering
 \begin{subfigure}{.32\linewidth}
        \centering
        \includegraphics[width=\linewidth]{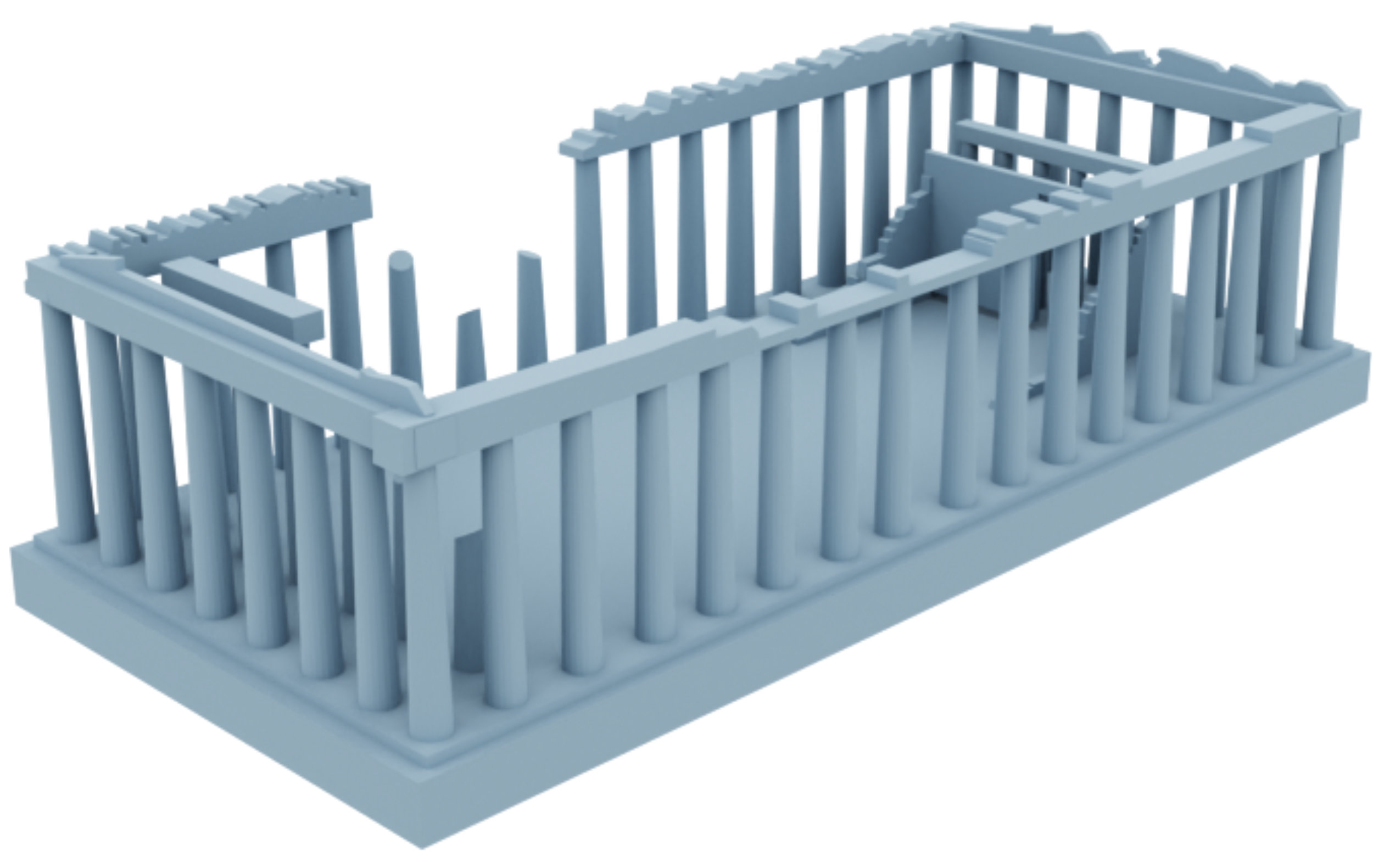}
    \end{subfigure}
    \begin{subfigure}{.32\linewidth}
        \centering
        \includegraphics[width=\linewidth]{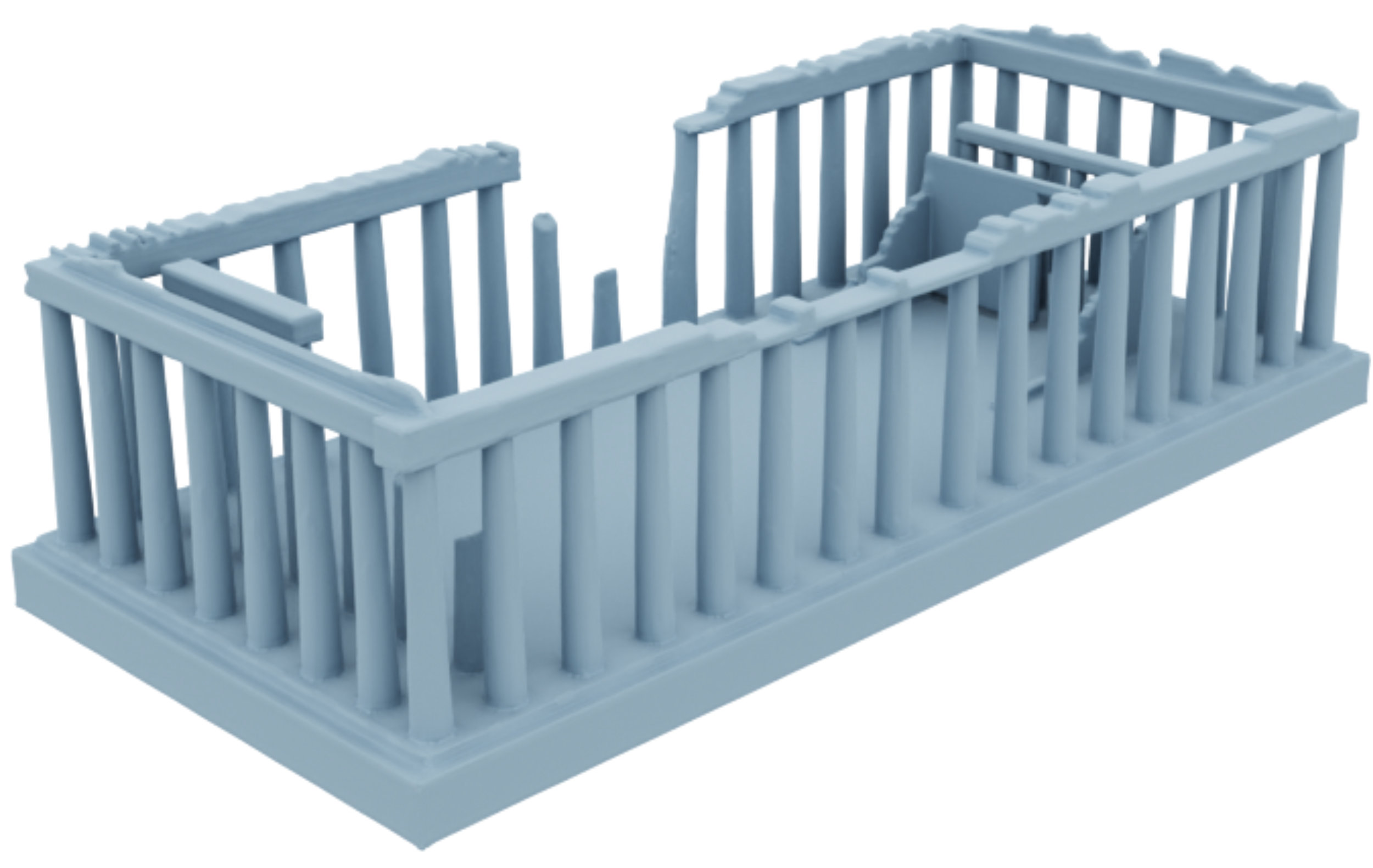}
    \end{subfigure}
    \begin{subfigure}{.32\linewidth}
        \centering
        \includegraphics[width=\linewidth]{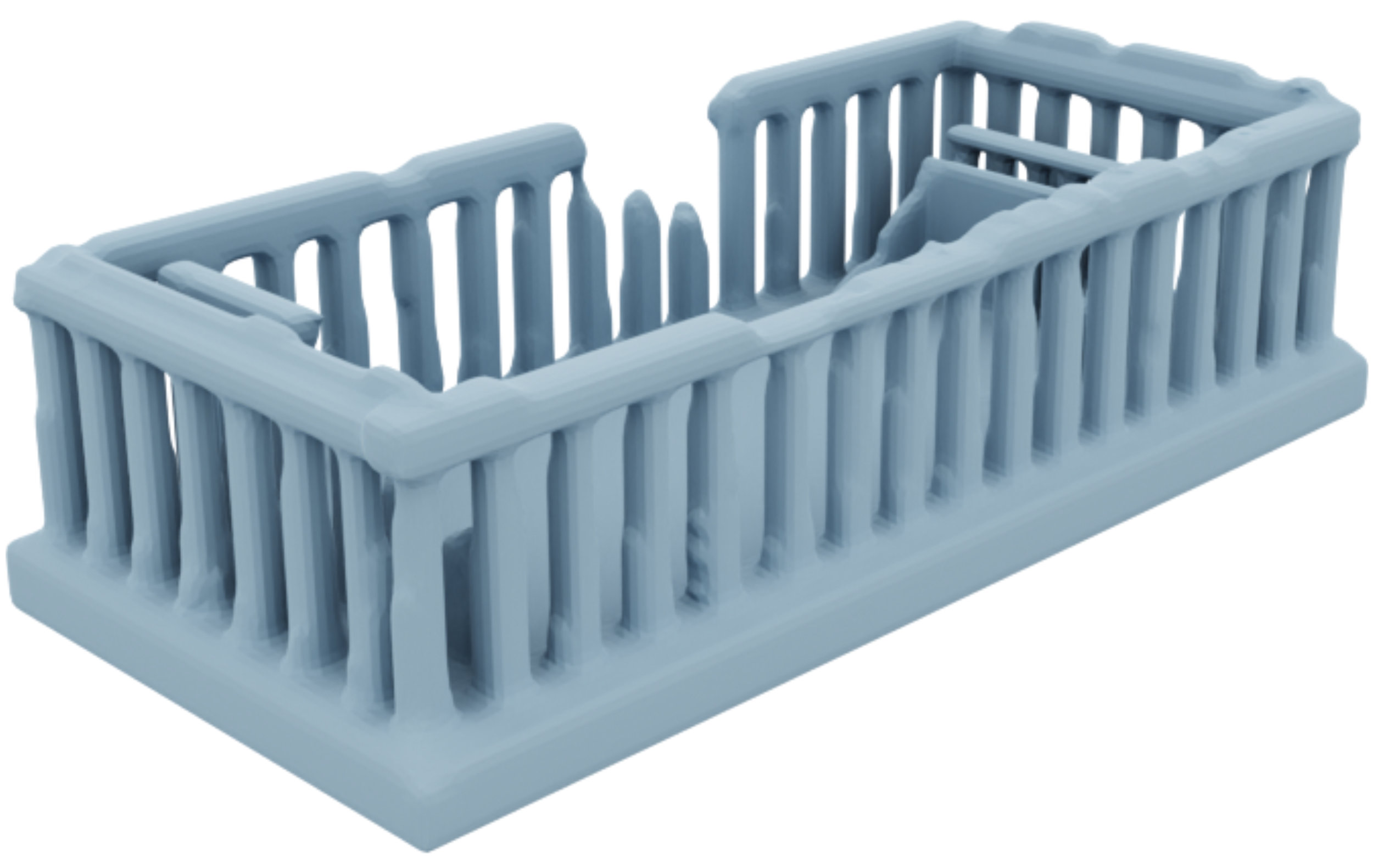}
    \end{subfigure}

    \begin{subfigure}{.32\linewidth}
        \centering
        \includegraphics[width=\linewidth]{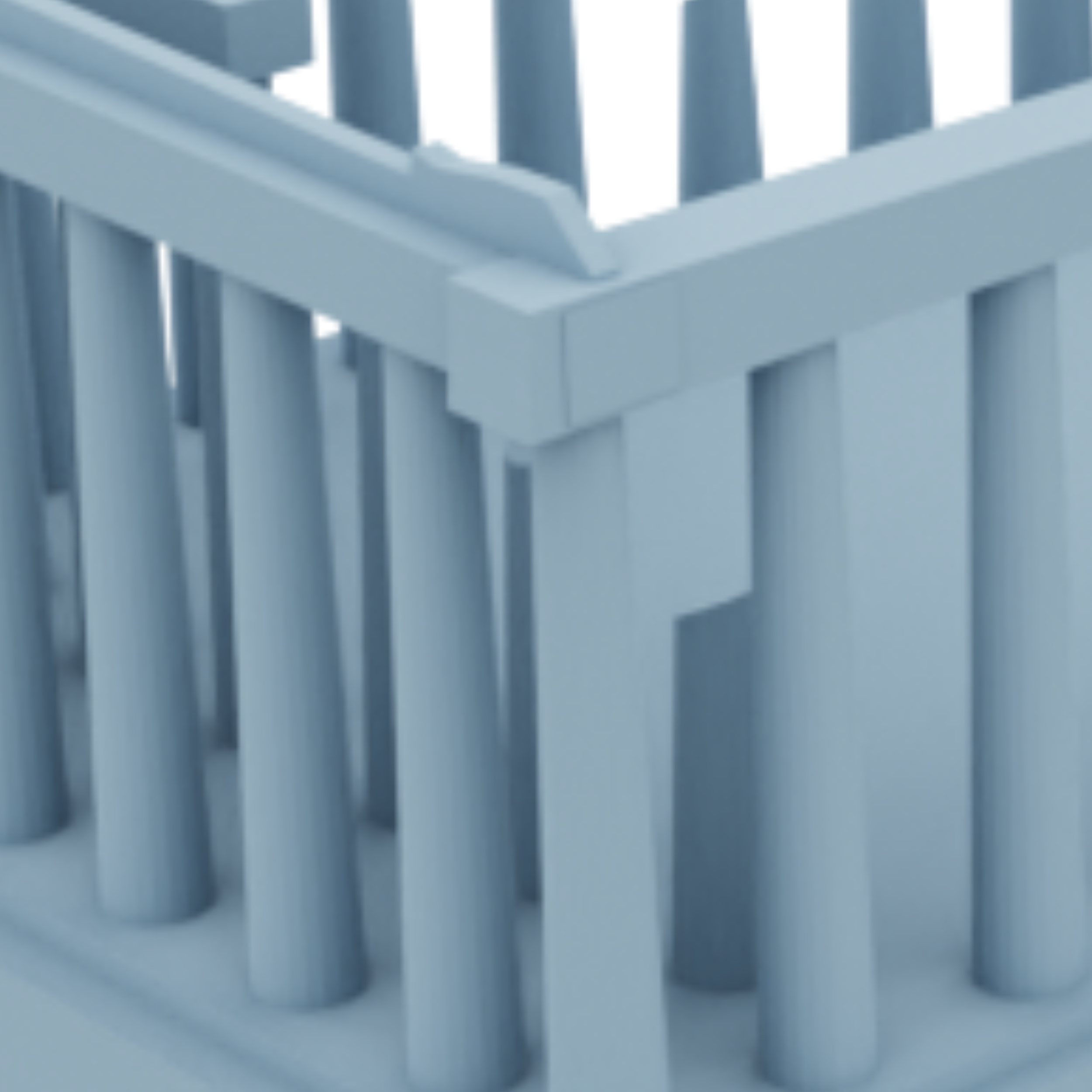}
    \end{subfigure}
     \begin{subfigure}{.32\linewidth}
        \centering
        \includegraphics[width=\linewidth]{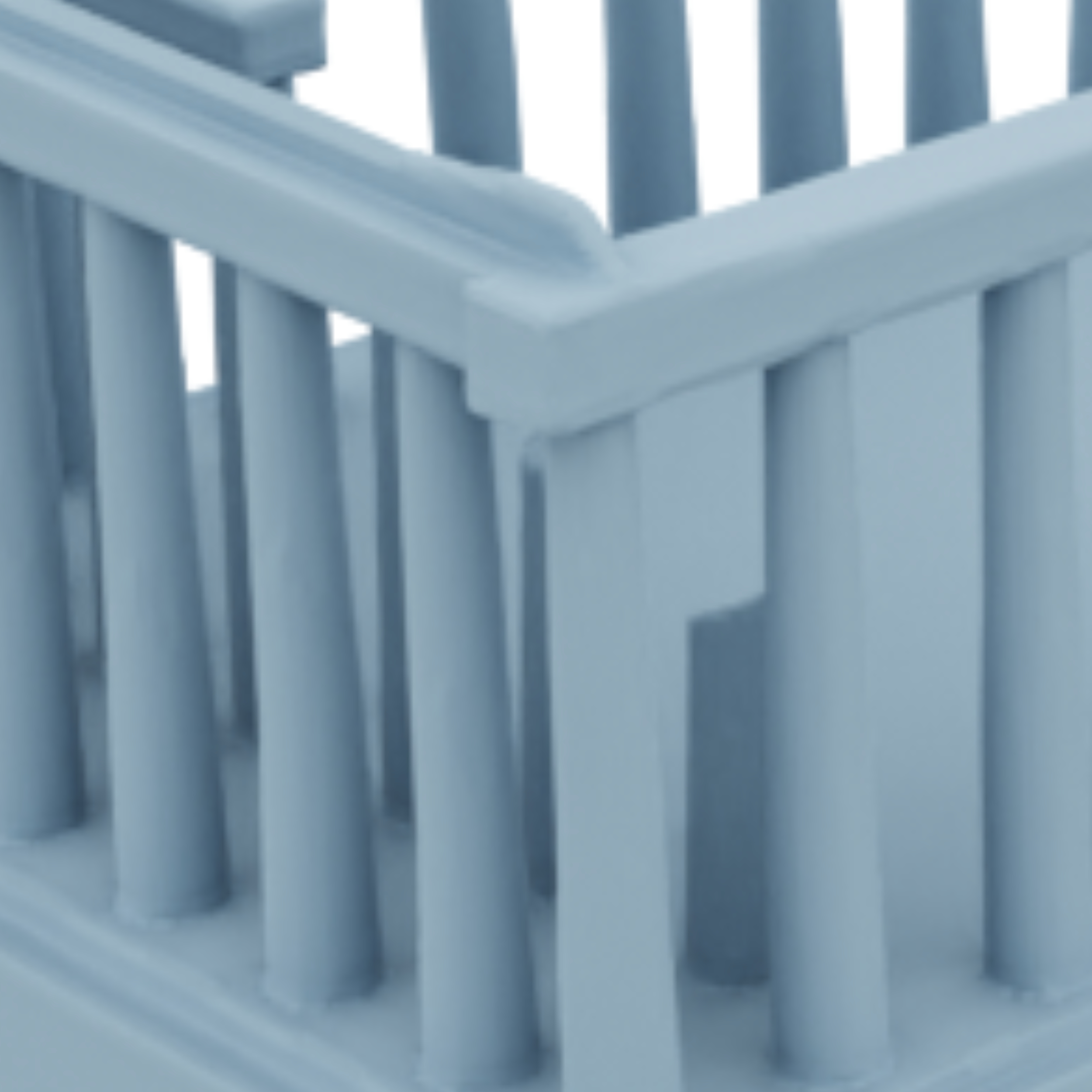}
    \end{subfigure}
    \begin{subfigure}{.32\linewidth}
        \centering
        \includegraphics[width=\linewidth]{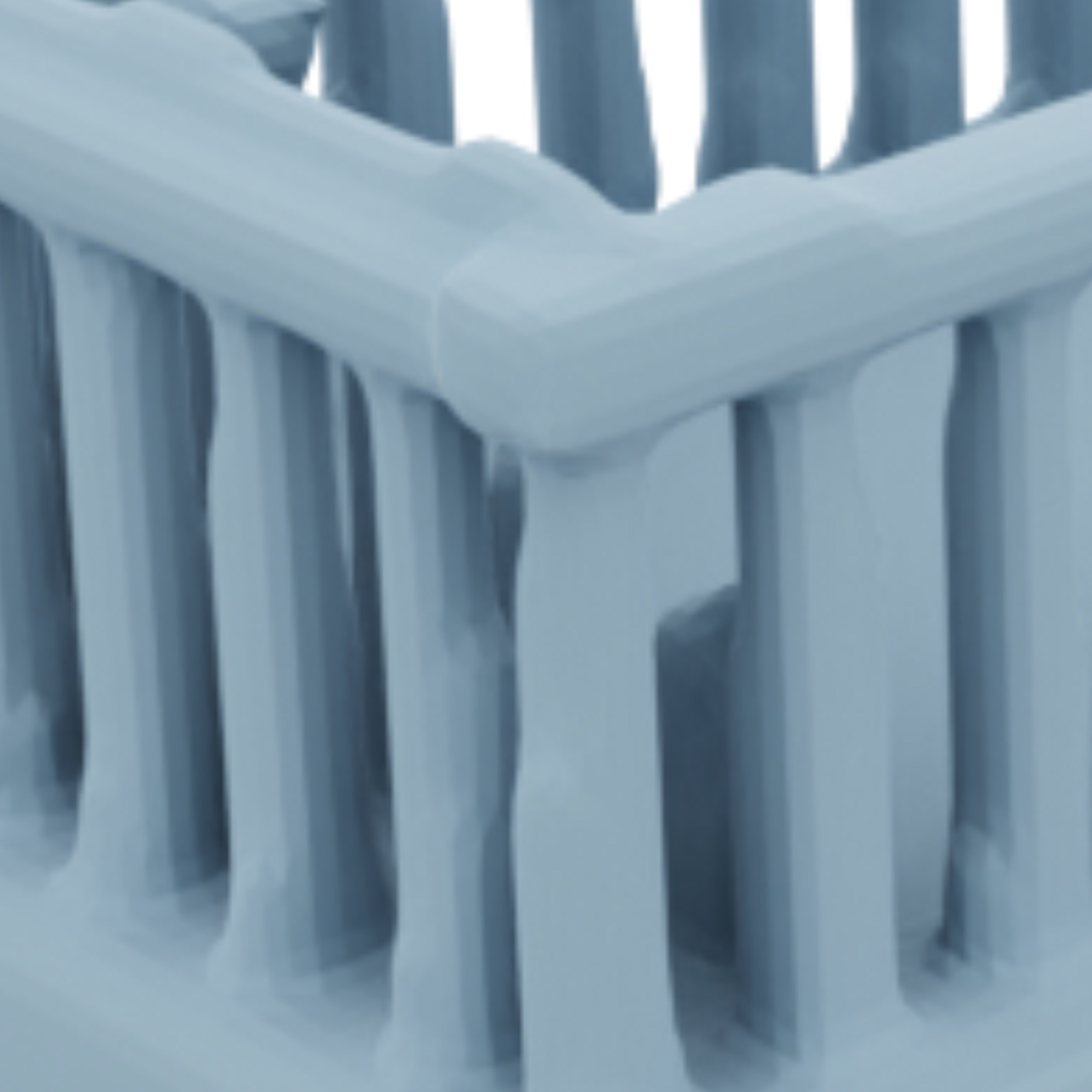}
    \end{subfigure}
    
 \begin{subfigure}{.32\linewidth}
        \centering
        \includegraphics[width=\linewidth]{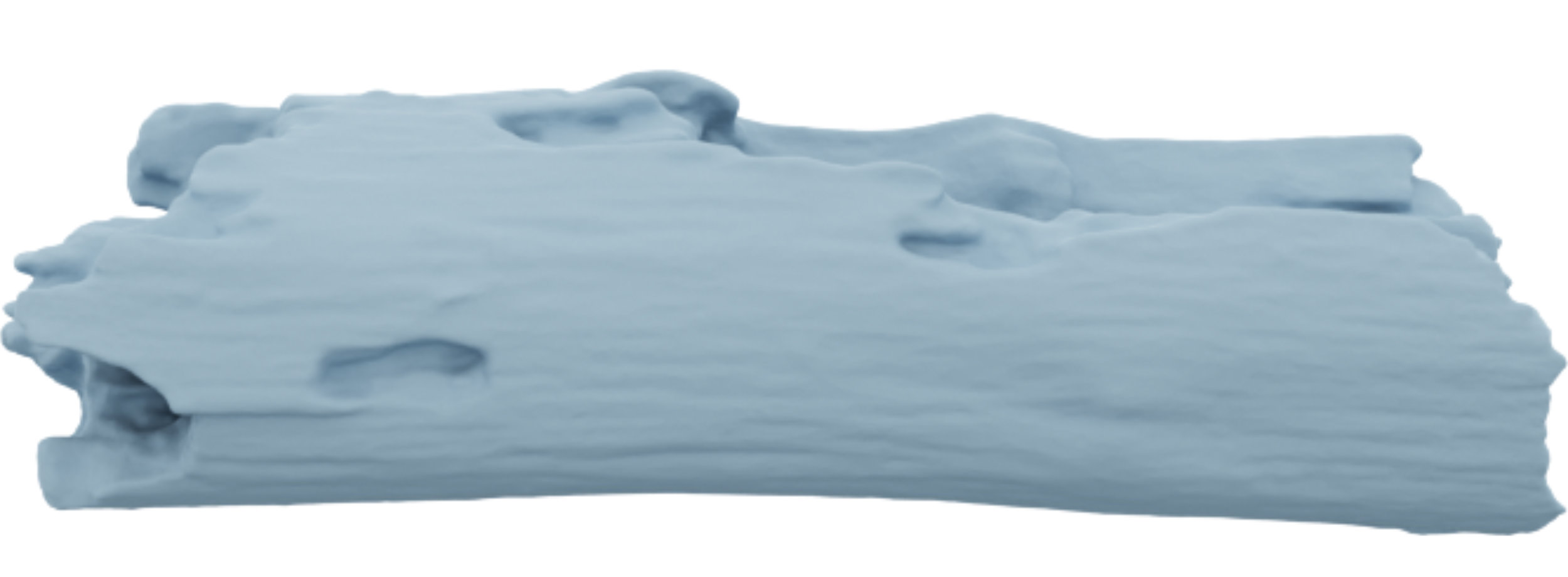}
    \end{subfigure}
    \begin{subfigure}{.32\linewidth}
        \centering
        \includegraphics[width=\linewidth]{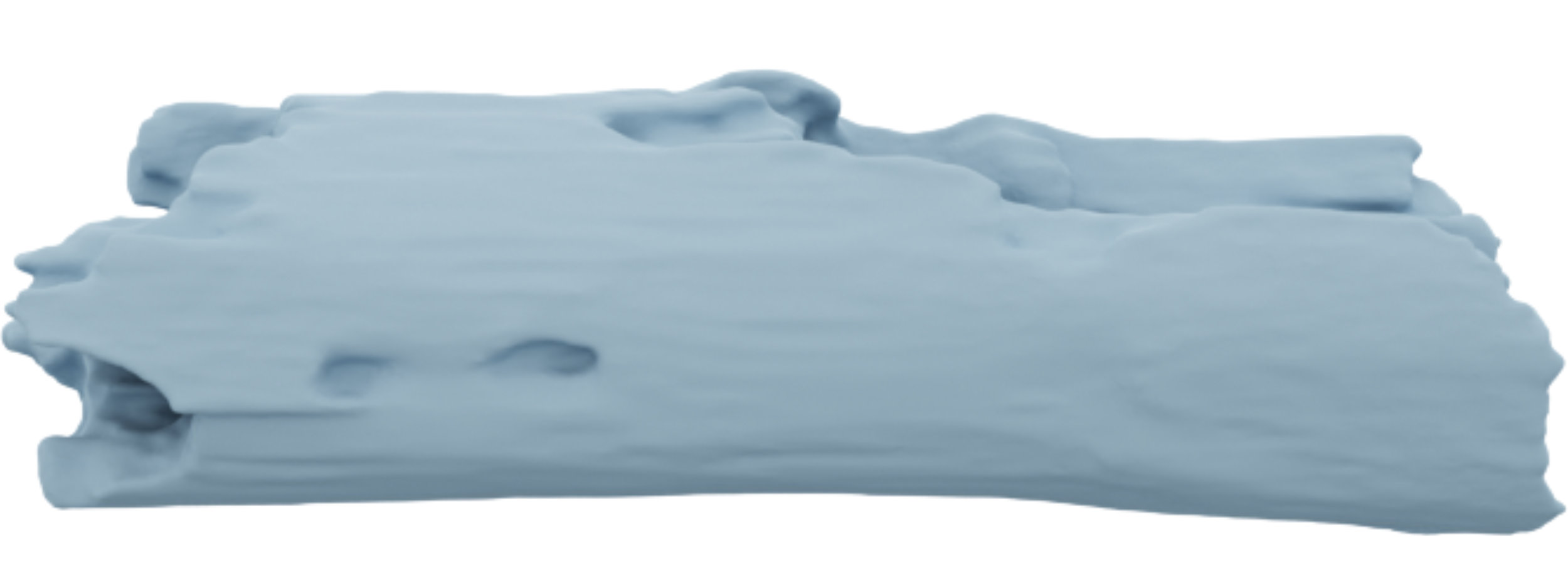}
    \end{subfigure}
    \begin{subfigure}{.32\linewidth}
        \centering
        \includegraphics[width=\linewidth]{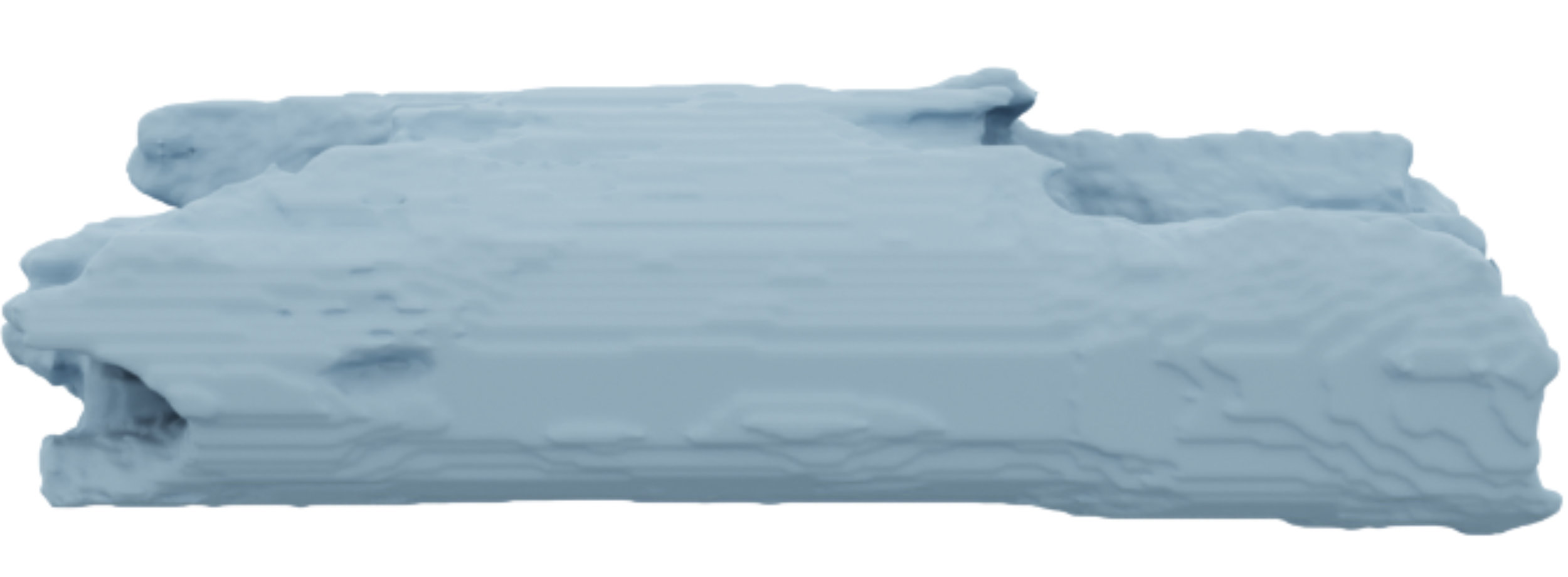}
    \end{subfigure}

\begin{subfigure}{.32\linewidth}
        \centering
        \includegraphics[width=\linewidth]{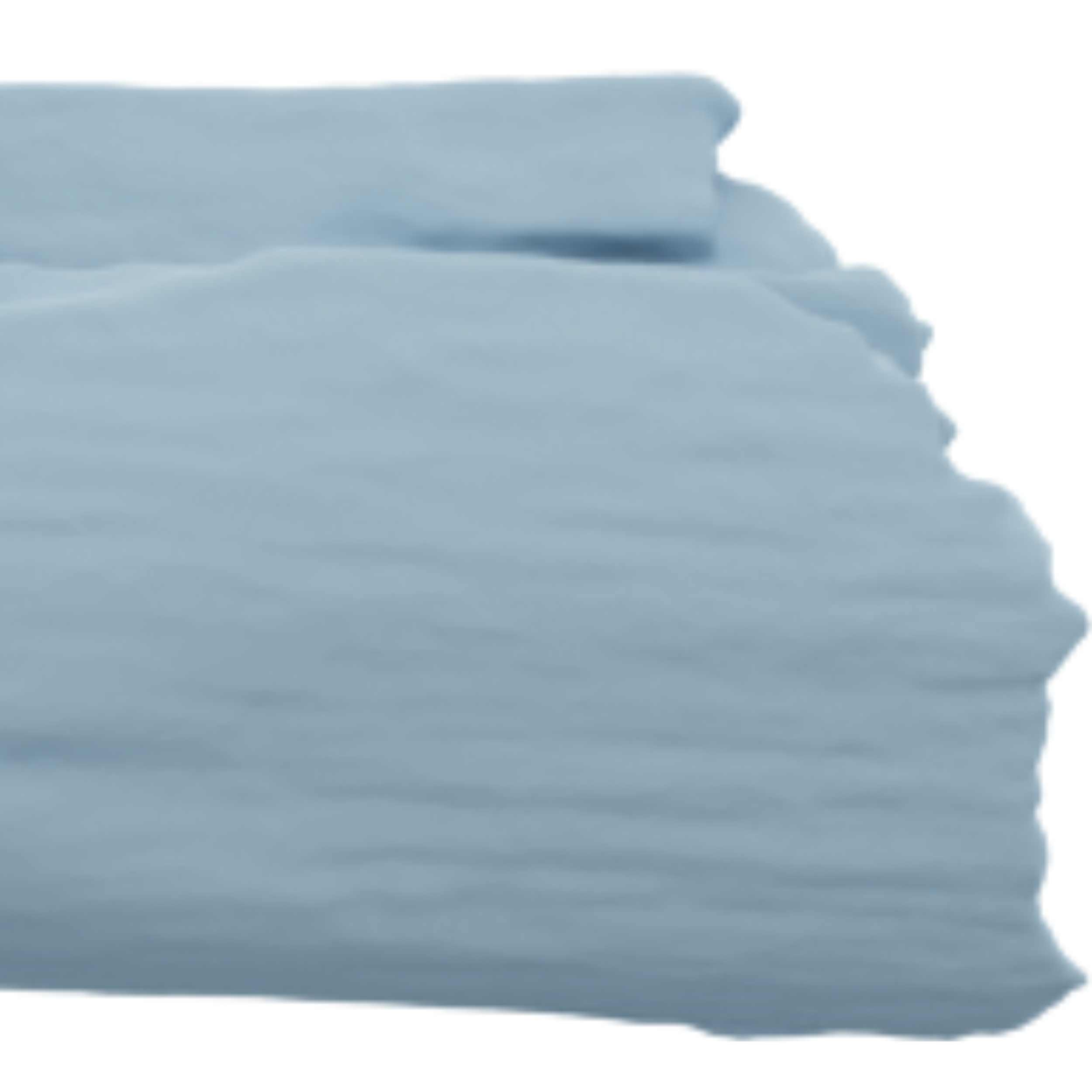}
        \caption{Input Geometry}
    \end{subfigure}
 \begin{subfigure}{.32\linewidth}
        \centering
        \includegraphics[width=\linewidth]{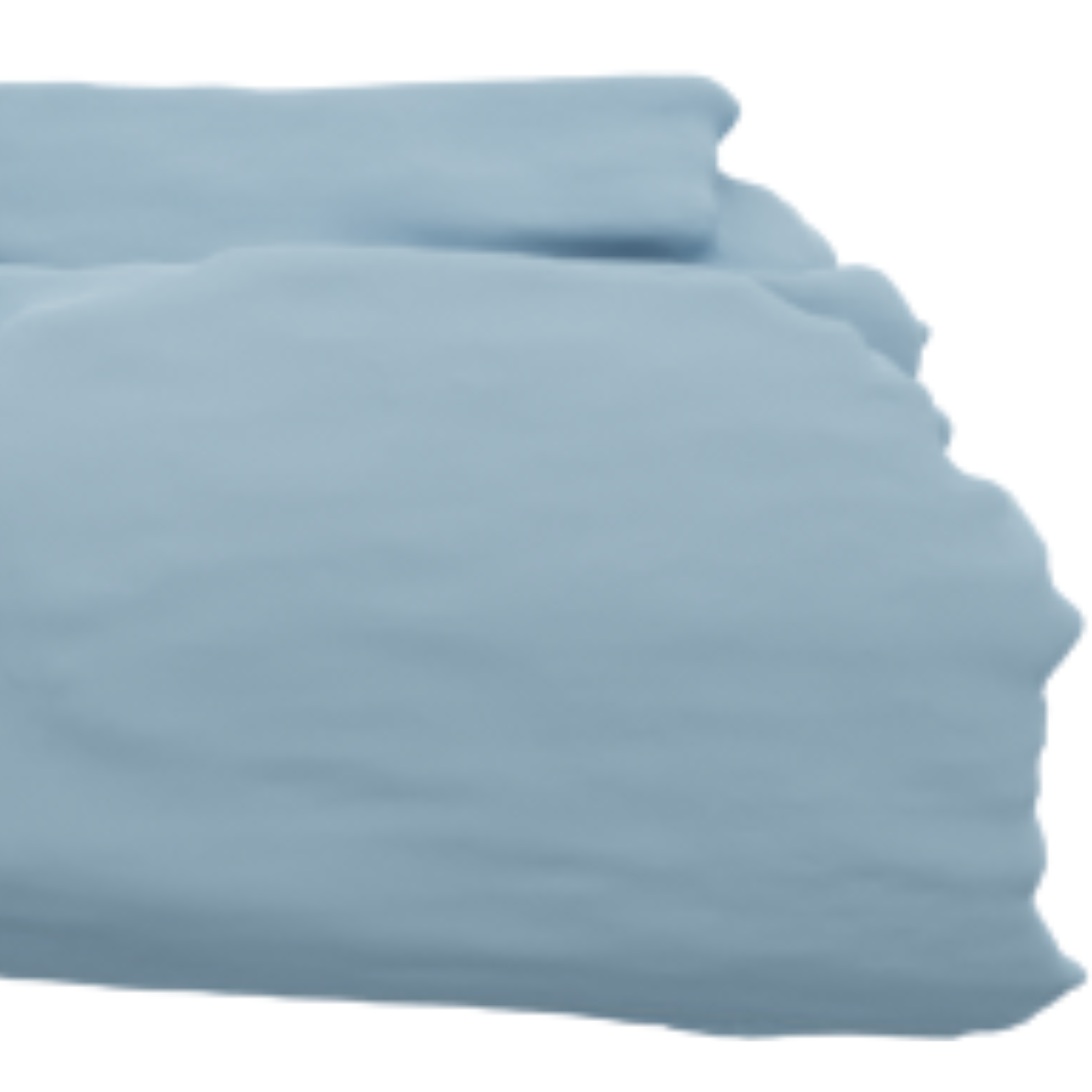}
        \caption{\ourmethod{}}
    \end{subfigure}
    \begin{subfigure}{.32\linewidth}
        \centering
        \includegraphics[width=\linewidth]{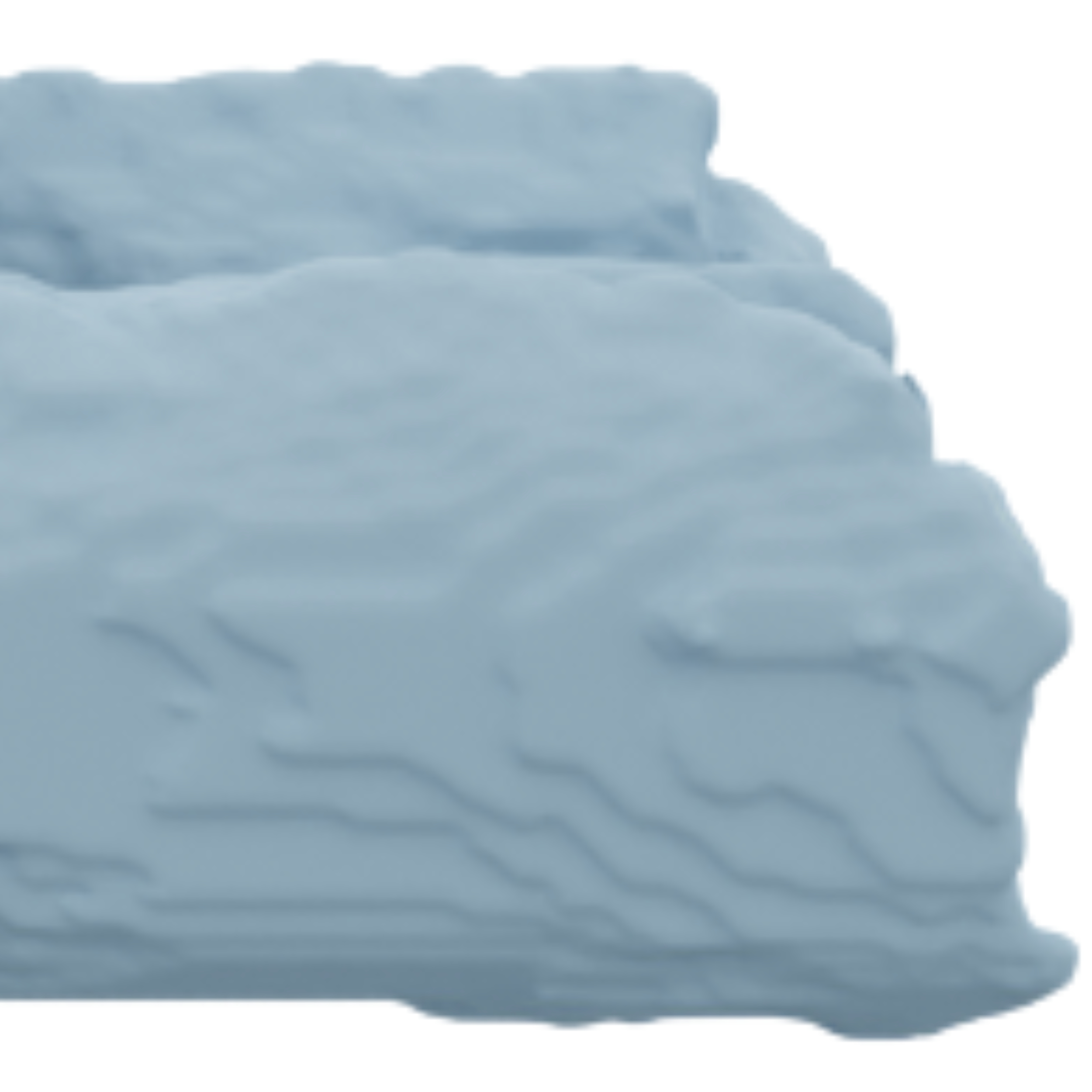}
        \caption{SSG}
    \end{subfigure}

    \caption{\textbf{Visual inspection of SSG~\cite{wu2022learning} results.} While SSG can generate 3D outputs with very short inference time, results are typically blobby or overly smooth, with spurious artifacts often visible due to its voxel-based generation process. In contrast, our method generates better sharp edges and subtle details.}
    \label{fig:ssg} 
\end{figure}

We provide additional comparisons with SSG~\cite{wu2022learning}, which is a 3D generalization of SinGAN~\cite{shaham2019singan} trained on multiscale triplane occupancy fields. Tab.~\ref{tab:ssfid2} shows quantitative evaluation on models for which SSG provides publicly available outputs, demonstrating the higher quality of \ourmethod. Furthermore, we demonstrate in Fig.~\ref{fig:ssg} that the typical results of this GAN-based method exhibit exaggerated smoothness like in all existing techniques, and often suffer from voxelized artifacts as well.

\begin{table}[!h]
    \centering
    \resizebox{1.\linewidth}{!}{\begin{tabular}{ l |c|  c  c  c  c }\toprule
Metric & Method & acropolis  & house & small-town & wood \\\midrule

\multirow{2}{*}{G-Qual. $\downarrow$} & SSG  &  2.81 & 0.91 & 1.71 & 0.07 \\
&\ourmethod{}  &  \textbf{0.01}   & \textbf{0.01}   &    \textbf{1.00}   &  \textbf{0.02}\\

\midrule
\multirow{2}{*}{G-Div. $\uparrow$} & SSG  &  \textbf{0.081}   &    \textbf{0.01}   &    0.19   &    \textbf{0.11}   \\
&\ourmethod{}  &  0.04   &    \textbf{0.01}    &   \textbf{0.60}   &     0.08 \\

\bottomrule

\end{tabular}

}
    \vspace*{-1mm}
    \caption{\textbf{Evaluating geometric quality and diversity using SSFID and pairwise IoU scores.} As we discussed in the main paper and in \cref{sec:add_metrics}, both metrics have their blindspots: SSFID tends to overlook geometric details, while pairwise IoU systematically rewards artifacts.} 
    \label{tab:ssfid2}
\end{table}

\subsection{Data-intensive vs. exemplar-based generation}

In this section, we discuss the value of exemplar-based 3D generation in light of the recent advancements in 3D generation models trained on millions of examples. The latter can be used to create highly diverse 3D assets and provide intuitive user controls through simple text and images. However, such models require immense computational resources for training and inference. Yet, as shown in \cref{fig:ours_0}, 
the state-of-the-art generator Rodin~\cite{zhang2024clay} (1.5B parameters) fails to create convincing geometric details comparable to those generated by our model.

Furthermore, the control provided by such models is limited, as the generation can only adhere to extremely coarse guidance. For example, in ~\cref{fig:ours_0}, we use the exemplar mesh as part of the inputs to Rodin for a conditioned generation. However, the output (right) completely loses the styles and details present in the exemplar mesh.



\section{Additional comments on metrics}
\label{sec:add_metrics}

While we use the two commonly-used metrics (geometric quality and diversity through SSFID and pairwise IoU scores) to evaluate our results and compare them to prior art, a few comments are in order. 

First, the validity of these two scores is debatable. While geometric quality is arguably fair but cannot really gauge the diversity of the results, the measure of diversity itself is quite delicate to analyze. In a sense, the diversity score rewards noise, not just real diversity. For instance, ten grids of random binary values would get a diversity of 0.66, while ten grids of axis-aligned planes that are not overlapping would have a score of 1.0 --- so a diversity score mixes different properties. This partial inadequacy of the score is the reason why we state in the main paper that geometric quality and geometric diversity should really be considered together to infer the success of an approach. Moreover, we also point out that the diversity scores should be clearly smaller for very structured models (like the acropolis model) than for free-form or organic shapes; our results have scores in line with this expected behavior, which seems more meaningful than systematic high scores which would point to noise artifacts instead of good results. 

Second, we wish to point out that our scores of Sin3DM~\cite{wu2024sindm} are \emph{different} from the ones they publish. The reason is that Sin3DM applies a pre-processing step to make the input meshes watertight. \emph{This initial step systematically inflates small details and thin surfaces} such as the roof of the house or the entablature of the acropolis, which negates many of the advantages of one-shot generative modeling: it degrades (at times severely) the input, losing the very reason why creating variants of a carefully-designed input model is highly sought after, i.e., the high-quality geometry of the exemplar. So we compared their results to the unprocessed input models, and did not re-train their neural network because we assumed that they made their best efforts to fit ground-truth shapes. So one should be aware that the low geometric quality scores we provide reflect \emph{both} the degradations of the pre-processing step and of their SDF-based generative approach --- again, to account for the real use of these generative approaches. 


\begin{table}[t]
    \centering
   \resizebox{1.\linewidth}{!}{\begin{tabular}{ l |c  c  c  c  c }\toprule
 Method & Level 0  & Level 1 & Level 2 & Level 3 & Level 4  \\\midrule
\ourmethod{}  &  0.49 & 0.17 & 0.18 & 0.26 &  0.78 \\
Sin3DM  &  - &5.18  & - & - &- \\


\bottomrule

\end{tabular}

}
    \vspace*{-3mm}
    \caption{\textbf{Inference timing for generating a single variation.} We report the inference time at each level for generating a single variation and compare it with Sin3DM, which has a grid resolution equivalent to our second level (level 1). Note that  in the main paper, we reported the inference time for 10 variations instead. DDIM sampling is used for both methods.\vspace*{-3mm}} 
    \label{tab:timings}
\end{table}

\section{Inference timings}
\label{sec:add_timings}

In the main paper, the inference times for \ourmethod and Sin3DM are reported for the generation of 10 variants. Here, we provide the inference timing for generating a single variant (i.e., using a batch size of 1 instead of 10) as shown in ~\cref{tab:timings}.

Our method generates a single variant in less than 2 seconds: approximately 0.5 seconds for the coarsest level, followed by less than 1.5 seconds in total for the four finer levels. In comparison, Sin3DM requires around 5 seconds, while Sin3DGen takes over 3 minutes, excluding the time needed for optimizing the input plenoxels and converting them to a mesh. Notably, our method produces the coarsest level in under half a second, which can be directly splatted using \cite{ravi2020pytorch3d} (see the video for live demonstrations). In contrast, Sin3DM\cite{wu2024sindm} takes 5.18 seconds to process an equivalent grid size ($32^3$).

\begin{figure}[h] \vspace*{-3mm}
\centering
    \begin{subfigure}{.49\linewidth}
        \centering
        \includegraphics[width=\linewidth]{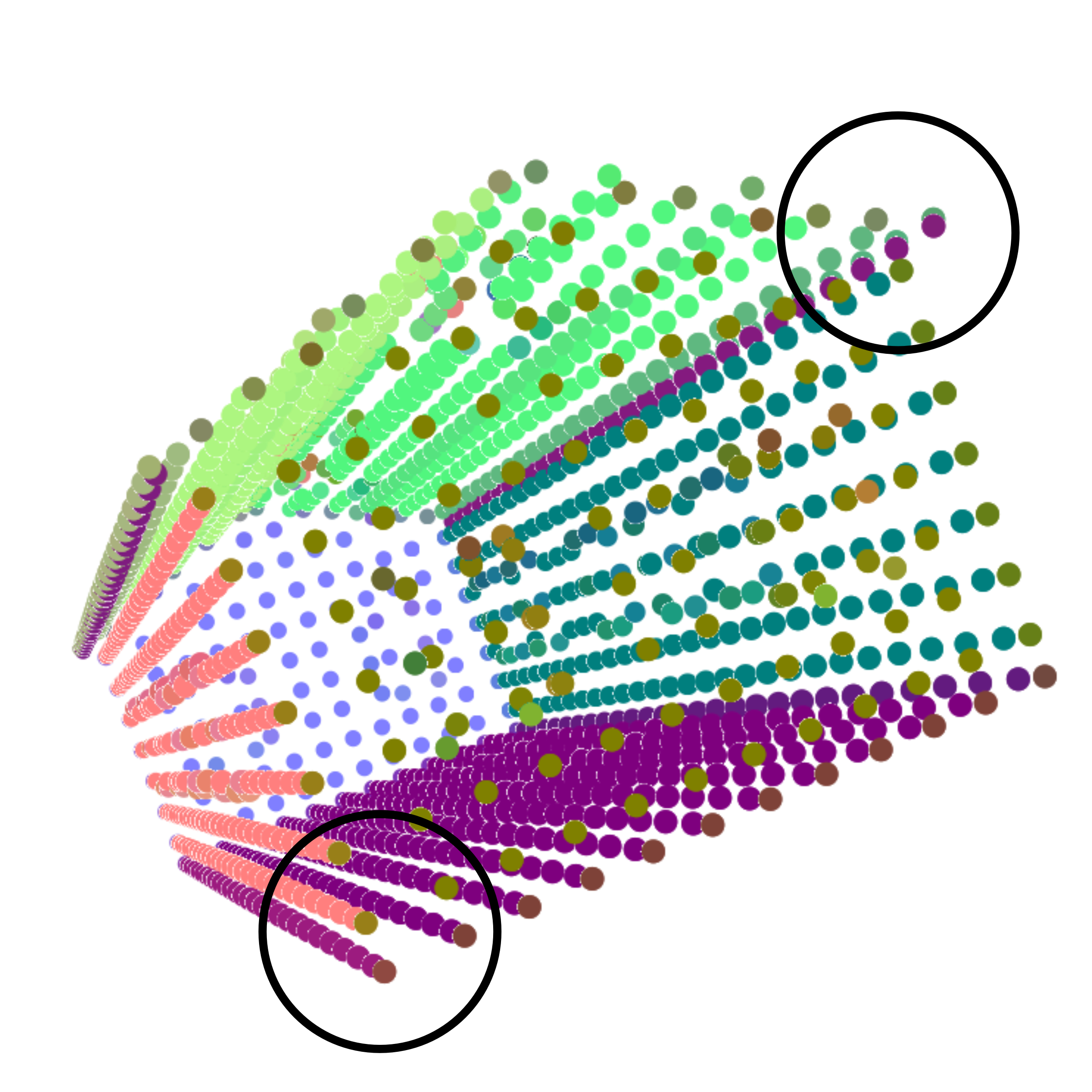}
        \caption{QEM averaging}
    \end{subfigure}
    \begin{subfigure}{.49\linewidth}
        \centering
        \includegraphics[width=\linewidth]{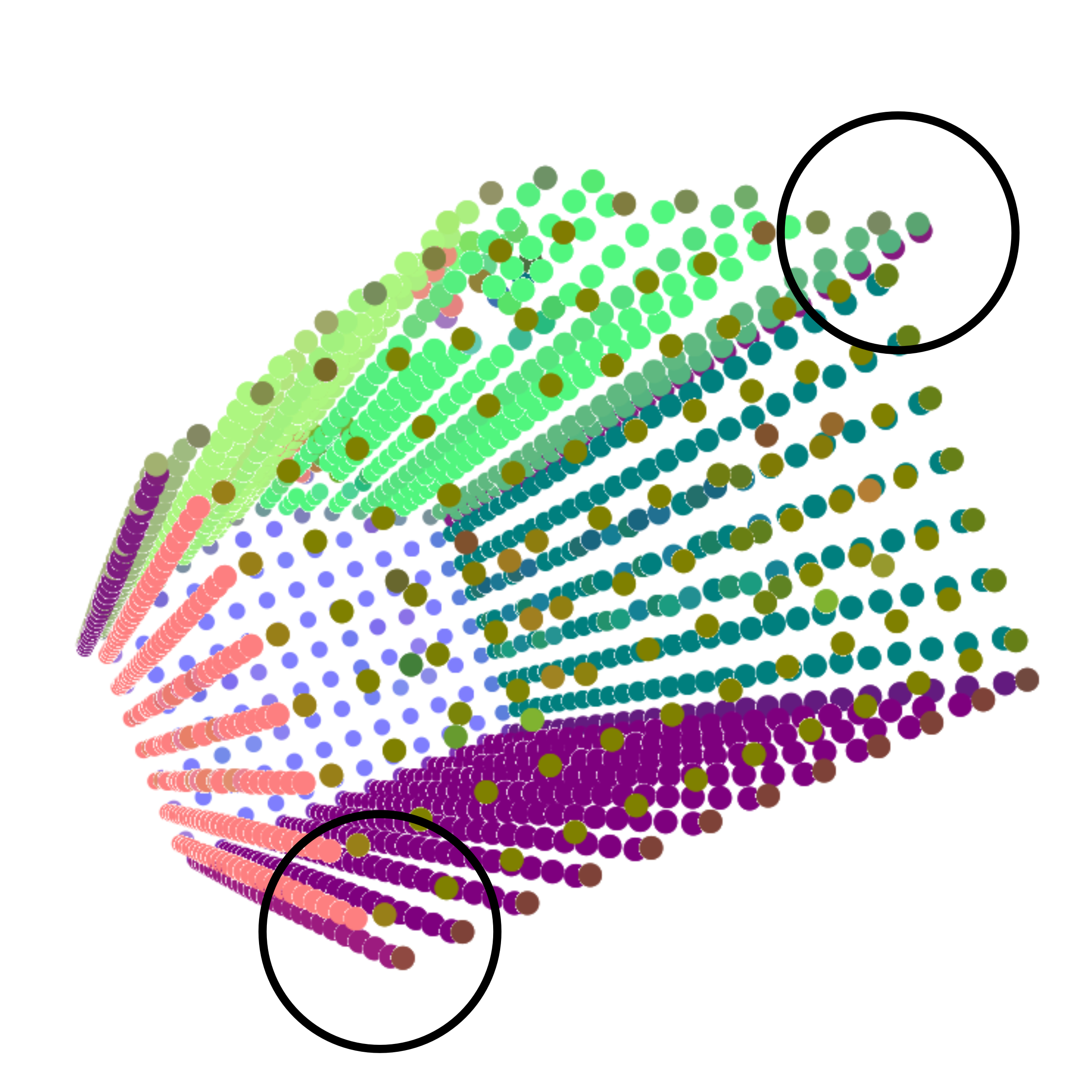}
         \caption{standard averaging}
    \end{subfigure}
    \caption{\textbf{QEM-averaging ablation.} While QEM-averaging (proposed in~\cite{maruani_ponq_2024}) keeps sharp features (like corners or spikes) in place helping our generative approach maintain these local details, a usual averaging would move the ``corner'' points inwards, increasing the probability of smoothing features out in generated variants.\vspace*{-1mm}}
    \label{fig:qem-ablation2}
\end{figure}

\section{QEM averaging}
\label{sec:qem}

Finally, we demonstrate why our use of QEM averaging during our fine-to-coarse analysis of the input models helps preserve sharp features of the ground truth. As Fig.~\ref{fig:qem-ablation2} demonstrates, standard scale-by-scale averaging of the points and normals from the finest sparse voxel grid all the way to the coarsest grid leads to drifts of the salient features: for instance, the bottom left corner of the house has migrated inwards, which may create rounding of the corner. Instead, applying the QEM averaging defined in the PoNQ method~\cite{maruani_ponq_2024} places the coarsest point on the corner, and of the intermediate points to remain right there as well --- resulting in outputs which will better preserve this geometric feature.

\end{document}